\documentclass[aos,preprint]{imsart}

\RequirePackage{amsthm,amsmath,amsfonts,amssymb}
\RequirePackage[numbers]{natbib}
\RequirePackage[colorlinks,citecolor=blue,urlcolor=blue]{hyperref}

\usepackage{longtable}
\usepackage{graphicx}
\usepackage{color}
\usepackage{mathrsfs}

\usepackage{natbib}
\usepackage[linesnumbered,ruled]{algorithm2e}

\startlocaldefs

\numberwithin{equation}{section}

\newcommand{\be}{\begin{equation}}
\newcommand{\ee}{\end{equation}}
\newcommand{\bes}{\begin{equation*}}
\newcommand{\ees}{\end{equation*}}
\newcommand{\beqn}{\begin{eqnarray}}
\newcommand{\eeqn}{\end{eqnarray}}
\newcommand{\beqns}{\begin{eqnarray*}}
\newcommand{\eeqns}{\end{eqnarray*}}
\newcommand{\lkr}{\left(}
\newcommand{\lkv}{\left[}
\newcommand{\rkv}{\right]}
\newcommand{\rkr}{\right)}
\newcommand{\lfi}{\left\{}
\newcommand{\rfi}{\right\}}

\newcommand{\fr}[1]{(\ref{#1})}

\newcommand{\del}{\delta}
\newcommand{\Del}{\Delta}

\newcommand{\eps}{\epsilon}

\newcommand{\ga}{\gamma}
\newcommand{\te}{\theta}
\newcommand{\om}{\omega}
\newcommand{\lam}{\lambda}
\newcommand{\Up}{\Upsilon}

\newcommand{\sig}{\sigma}

\newcommand{\Lam}{\Lambda}
\newcommand{\Om}{\Omega}
\newcommand{\Sig}{\Sigma}
\newcommand{\Te}{\Theta}

\newcommand{\EE}{\ensuremath{{\mathbb E}}}

\newcommand{\PP}{\ensuremath{{\mathbb P}}}

\newcommand{\RR}{{\mathbb R}}

\newcommand{\vect}{\mbox{vec}}

\newcommand{\card}{\mbox{card}}

\newcommand{\Var}{\mbox{Var}}
\newcommand{\Cov}{\mbox{Cov}}
\newcommand{\diag}{{\rm diag}}

\newcommand{\argmin}{\mbox{argmin}}
\newcommand{\argmax}{\mbox{argmax}}

\newcommand{\rank}{\mbox{rank}}
\newcommand{\Reg}{\mbox{Reg}}
\newcommand{\SVD}{\mbox{SVD}}
\newcommand{\sgn}{\mbox{sign}}
\newcommand{\iid}{\mbox{i.i.d.}}
\newcommand{\Uniform}{\mbox{Uniform}}
\newcommand{\Err}{\mbox{Err}}
\newcommand{\Ins}{\mbox{Ins}}
\newcommand{\ARI}{\mbox{ARI}}
\newcommand{\VI}{\mbox{VI}}
\newcommand{\Ent}{\mbox{Ent}}
\newcommand{\Inf}{\mbox{Inf}}

\newtheorem{thm}{Theorem}
\newtheorem{lem}{Lemma}
\newtheorem{cor}{Corollary}
\newtheorem{rem}{Remark}
\newtheorem{ex}{Example}
\newtheorem{prop}{Proposition}

\newcommand{\bA}{\mathbf{A}}

\newcommand{\bG}{\mathbf{G}}
\newcommand{\bH}{\mathbf{H}}

\newcommand{\bP}{\mathbf{P}}

\newcommand{\bU}{\mathbf{U}}

\newcommand{\bX}{\mathbf{X}}
\newcommand{\bY}{\mathbf{Y}}

\newcommand{\bTe}{\mbox{\mathversion{bold}$\Theta$}}
\newcommand{\bXi}{\mbox{\mathversion{bold}$\Xi$}}

\newcommand{\calB}{{\mathcal{B}}}

\newcommand{\calF}{{\mathcal{F}}}
\newcommand{\calG}{{\mathcal G}}
\newcommand{\calH}{{\mathcal H}}
\newcommand{\calI}{{\mathcal{I}}}

\newcommand{\calM}{{\mathcal M}}
\newcommand{\calN}{{\mathcal N}}
\newcommand{\calO}{{\mathcal O}}
\newcommand{\calP}{{\mathcal{P}}}
\newcommand{\calQ}{{\mathcal{Q}}}

\newcommand{\calV}{{\mathcal{V}}} 
\newcommand{\calU}{{\mathcal{U}}}
 
\newcommand{\cald}{{\mathfrak{D}}}

\newcommand{\tilcalF}{\widetilde{\calF}}

\newcommand{\lan}{\langle}
\newcommand{\ran}{\rangle}

\newcommand{\sumlL}{\sum_{l=1}^L}

\newcommand{\minnL}{(n \wedge L)}
\newcommand{\maxnL}{(n \vee L)}

\newcommand{\hD}{\widehat{D}}

\newcommand{\hX}{\widehat{X}}
\newcommand{\hY}{\widehat{Y}}
\newcommand{\hM}{\widehat{M}}

\newcommand{\hU}{\widehat{U}}
\newcommand{\hV}{\widehat{V}}
\newcommand{\hW}{\widehat{W}}
\newcommand{\hSig}{\widehat{\Sig}}
\newcommand{\hs}{\hat{s}}
\newcommand{\hd}{\hat{d}}
\newcommand{\hte}{\hat{\te}}
\newcommand{\hTe}{\widehat{\Te}}
\newcommand{\hS}{\widehat{S}}
\newcommand{\hPsi}{\widehat{\Psi}}

\newcommand{\barU}{\overline{U}}

\newcommand{\hr}{\hat{r}}

\newcommand{\hKM}{\widehat{\barK\, M}}

\newcommand{\barK}{\overline{K}}

\newcommand{\tilc}{\tilde{c}}
\newcommand{\tilA}{\widetilde{A}}
\newcommand{\tilC}{\widetilde{C}}

\newcommand{\tilR}{\widetilde{R}} 
\newcommand{\tilbG}{\widetilde{\bG}}
\newcommand{\tilbP}{\widetilde{\bP}}
\newcommand{\tilbA}{\widetilde{\bA}}
\newcommand{\tilbXi}{\widetilde{\bXi}}

\newcommand{\tilPhi}{\widetilde{\Phi}}

\newcommand{\tilK}{\widetilde{K}}
\newcommand{\tilX}{\widetilde{X}}
\newcommand{\tilU}{\widetilde{U}}
\newcommand{\tilD}{\widetilde{D}}
\newcommand{\tilO}{\widetilde{O}}
\newcommand{\tilP}{\widetilde{P}}

\newcommand{\tilV}{\widetilde{V}}
\newcommand{\tilW}{\widetilde{W}}
\newcommand{\tilB}{\widetilde{B}}
\newcommand{\tilb}{\widetilde{b}}

\newcommand{\tilXi}{\widetilde{\Xi}}
\newcommand{\tilsig}{\widetilde{\sig}}
\newcommand{\tilSig}{\widetilde{\Sig}}

\newcommand{\tilu}{\tilde{u}}
\newcommand{\tilv}{\tilde{v}}

\newcommand{\tilnu}{\tilde{\nu}}

\newcommand{\tDel}{\widetilde{\Del}}

\newcommand{\frB}{\mathfrak{B}}
\newcommand{\frb}{\mathfrak{b}}
\newcommand{\frZ}{\mathfrak{Z}}

 \newcommand{\scrT}{\mathscr{T}}

 \newcommand{\scrP}{\mathscr{P}}
 
 \newcommand{\scrX}{\mathscr{X}}

\newcommand{\minmM}{\displaystyle \min_{\minM}\ }
\newcommand{\maxmM}{\displaystyle \max_{\minM}\ }

 \newcommand{\maxL}{\displaystyle \max_{\linL}\ }

 \newcommand{\di}{\displaystyle}

\newcommand{\lowc}{\underline{c}}
\newcommand{\highc}{\bar{c}}
\newcommand{\lowC}{\underline{C}}
\newcommand{\highC}{\bar{C}}
\newcommand{\lowd}{\underline{d}}
\newcommand{\highd}{\bar{d}}
\newcommand{\lowvarpi}{\underline{\varpi}}
\newcommand{\lowa}{\underline{a}}
\newcommand{\higha}{\bar{a}}
\newcommand{\lowlam}{\underline{\lam}}

\newcommand{\rhon}{\rho_{n}}
 
\newcommand{\sinTe}{\sin\Te}

\long\def\ignore#1{}

\newcommand{\upl}{^{(l)}}
\newcommand{\uplm}{^{(l,m)}}
\newcommand{\upm}{^{(m)}}
\newcommand{\upmt}{^{(m,t)}}

\newcommand{\upt}{^{(t)}}
\newcommand{\upo}{^{(0)}}
\newcommand{\upto}{^{(t-1)}}
\newcommand{\upone}{^{(1)}}
\newcommand{\uptwo}{^{(2)}}

\newcommand{\linL}{l \in [L]}
\newcommand{\minM}{m \in [M]}

\newcommand{\tio}{{_{\times 1}}}
\newcommand{\tit}{{_{\times 2}}}
\newcommand{\tir}{{_{\times 3}}}

\newcommand{\delo}{\del_1}
\newcommand{\delt}{\del_2}
\newcommand{\delr}{\del_3}
\newcommand{\delu}{\del_u}
\newcommand{\delw}{\del_w}
\newcommand{\delk}{\del_k}

\newcommand{\nk}{n_k}

\newcommand{\twin}{{2, \infty}}

\newcommand{\Niter}{N_{\rm iter}}
\newcommand{\epstol}{\eps_{\rm tol}}

\newcommand{\delsnl}{\Del_s (n,L)}
\newcommand{\epsnl}{\eps (n,L)}
\newcommand{\epsunl}{\eps^{(U)} (n,L)}
\newcommand{\epswnl}{\eps^{(W)} (n,L)}
\newcommand{\lkd}{_{k,\delk}}

\newcommand{\ik}{i_k}
\newcommand{\io}{i_1}
\newcommand{\itw}{i_2}

\newcommand{\jo}{{j_1}}
\newcommand{\jt}{{j_2}}

\newcommand{\ipo}{i'_1}
\newcommand{\ipt}{i'_2}

\newcommand{\ijo}{i_{\jo}}
\newcommand{\ijt}{i_{\jt}}

\newcommand{\calvs}{\calV^{*}}

\newcommand{\icalI}{(i'_1,i'_2) \in \calI_{\io,\itw}}

\newcommand{\nus}{\nu^{*}}

\newcommand{\tinf}{_{2,\infty}}
\newcommand{\bGo}{\bG_{0}}
\newcommand{\Ctau}{C_{\tau}}
\newcommand{\Omtau}{\Om_{\tau}}

\newcommand{\timjo}{{_{\times \jo}}}
\newcommand{\timjt}{{_{\times \jt}}}
\newcommand{\sigo}{\sig_1}
\newcommand{\sigt}{\sig_2}
\newcommand{\sigr}{\sig_3}

\newcommand{\Ruk}{R^{(k)}}
\newcommand{\Pibot}{\Pi^{\bot}}
\newcommand{\hscrT}{\widehat{\scrT}}

\newcommand{\hUo}{\hU^{(0)}}
\newcommand{\tilUo}{\tilU^{(0)}}
\newcommand{\hWo}{\hW^{(0)}}
\newcommand{\tilWo}{\tilW^{(0)}}
\newcommand{\tileps}{\widetilde{\eps}}

\newcommand{\teps}{\widetilde{\eps}}
\newcommand{\tepso}{\teps_0}
\newcommand{\tepstinf}{\teps\tinf}
\newcommand{\tDelo}{\tDel_0}
\newcommand{\tDeltinf}{\tDel\tinf}

\newcommand{\tilXihu}{\tilXi_{\hU}}

\newcommand{\tilCtau}{\tilC_{\tau}}
\newcommand{\mdel}{m_{\del}}

\newcommand{\nAi}{n_{A,i}}
\newcommand{\nBj}{n_{B,j}}
\newcommand{\barARI}{\overline{\ARI}}

\endlocaldefs

\begin{document}

\begin{frontmatter}

\title{Perfect Clustering in Extremely Sparse Multilayer Networks
  with Groupwise Subspace Structure}

\runtitle{Perfect Clustering in Extremely Sparse Multilayer Network }

\begin{aug}

\author[A]{\fnms{Marianna}~\snm{Pensky}\ead[label=e1]{marianna.pensky@ucf.edu}}


\address[A]{Department of Mathematics,
University of Central Florida, Orlando, USA \printead[presep={,\ }]{e1}}

\end{aug}

\begin{abstract}
The paper studies a multilayer network on the same collection of nodes, where each layer follows the Generalized Random Dot Product Graph Model. Furthermore, all layers can be partitioned into groups, each group with its own unique latent subspace structure, but otherwise the matrices of connection probabilities can all be different. This setting is extremely general and includes most multilayer network models as special cases. In addition, we are interested in the situation where the layer networks may be extremely sparse and, therefore, the layer-by-layer analysis is unreliable. This problem can also be seen as a generalization of the classic setting in which one recovers a shared structure across an entire collection of matrices, extending it to the case where such structures are shared only within specific groups of matrices. 
The key task in this setting is to recover the groups of layers with unique subspace structures, since, after that, the information in those groups of layers can be pooled together by well-studied methods. Until now, clustering of layers in such networks was based on layer-by-layer analysis, which required the multilayer network to be sufficiently dense. Our paper is the first to provide a tensor-based approach to the problem that combines information from all layers in an unconventional manner. Our theoretical results, established under intuitive and non-restrictive assumptions, assert that the new technique achieves perfect clustering under sparsity conditions that, up to logarithmic factors, coincide with the computational lower bound derived for a much simpler model. 
The paper also delivers a nuanced message about the applicability of the new technique.
\end{abstract}

\begin{keyword}
\kwd{Multilayer Network}
\kwd{Generalized Dot Product Graph}
\kwd{Tensor}
\kwd{Clustering}
\end{keyword}

\end{frontmatter}



 
\section{Introduction}
\label{sec:introduction}

The paper studies a multilayer network on the same collection of nodes, where  all the edges between nodes are drawn within  layers, 
i.e., there are no edges connecting the nodes 
in different layers. We assume that  each layer follows the well-studied Generalized Random Dot Product Graph Model of \cite{GDPG}, 
which includes a variety of popular network models as its special cases: the Stochastic Block Model, the Degree Corrected Block Model,
the Popularity Adjusted Block Model \cite{RePEc:bla:jorssb:v:80:y:2018:i:2:p:365-386}, 
the Mixed Membership Model \cite{Airoldi:2008:MMS:1390681.1442798},
and the Degree Corrected Mixed Membership Model  \cite{jin2022mixedmembershipestimationsocial}.
In order to increase the flexibility of the model, we furthermore assume that  all layers can be partitioned into groups, 
 each group with its own  unique latent subspace structure, but otherwise the matrices of connection probabilities can be all different. 
This setting is extremely general and  includes the majority of popular multilayer network models as special cases.
There are many real-life networks that satisfy these 
constraints, such as, for instance, trade networks (nodes are countries, layers are commodities in which they trade) 
and brain networks (nodes are brain regions, layers are individuals).

Specifically, the paper considers  an $L$-layer network on the same set of 
$n$ vertices.  
The tensor of connection probabilities  $\bP \in [0,1]^{n \times n \times L}$
is formed by the layers  $P^{(l)}$, $l \in [L] = \{1,\cdots,L\}$,  that can be 
subdivided into clusters with unique subspace structure. The latter means that 
there exists a label function $s: [L] \to [M]$  such that  $P\upl$, $l \in [L]$, are given by
\be \label{eq:DIMPLE_GDPG}
 P\upl =  U\upm  Q\upl (U\upm)^T, \quad m = s(l),  \ \ \linL,  
\ee
where $Q\upl = (Q\upl)^T $ can be all different, and  $U\upm$, $\minM$,  are matrices  with orthonormal columns. 
Moreover, in order to improve adjustability of the model and to accommodate correlation networks, 
we allow elements of matrices $P\upl$ to   lie in the interval $[-1,1]$. The latter  is also 
partially motivated by the fact that non-negativity of elements of matrices $P\upl$ are hard to ensure.
In this setting, one observes the adjacency tensor $\bA \in \{0,1\}^{n \times n \times L}$ with layers $A\upl$ such that 
$A{\upl} (i,j)$  are independent from each other for $1 \leq i <  j \leq n$ and $\linL$, and
$A{\upl} (i,j) = A\upl (j,i)$. Here,  $A{\upl} (i,j)$ preserves the sign of $P{\upl} (i,j)$ while
$|A{\upl} (i,j)|$ follow the Bernoulli distribution 
\be  \label{eq:SGDPG_adjacen}
\PP(|A\upl(i,j)| = 1) = |P\upl(i,j)|, \quad
  \sgn(A\upl(i,j)) = \sgn(P\upl(i,j)). 
\ee
It is easy to see that the latter reduces to the traditional  binary network when all elements $P{\upl} (i,j)$
of tensor $\bP$ are nonnegative, so all results of the paper remain valid in this common case. 
In what follows, we denote the number of layers of type $m$ by $L_m$, and the ambient dimension of a layer of type $m$ by 
$K_m$, so that $Q\upl  \in [0,1]^{K_m \times K_m}$, $\minM$, in \fr{eq:DIMPLE_GDPG}. 
 This setting  was named the  DIverse MultiPLEx  Signed Generalized Random Dot 
Product Graph ({\bf DIMPLE-SGRDPG}) model in \cite{pensky2024signed}. 
As we show in Section~\ref{sec:related_work}, many multilayer network designs are 
generalized by the model above. In particular, if all layers of the network are embedded in the 
same subspace ($M=1$), DIMPLE-SGRDPG model \fr{eq:DIMPLE_GDPG} reduces to the COmmon Subspace Independent Edge
({\bf COSIE}) model studied in \cite{JMLR:v18:17-448}  and \cite{arxiv.2007.10455}.

In order to motivate diverse subspace structure, consider    a common particular case, 
where layers of the network follow the Stochastic Block Models ({\bf SBM}). 
For the SBM-equipped network, the matrix of probabilities of connection in layer $l$ can be expressed as 
\be \label{eq:SBM}
 P\upl =  Z\upm  B\upl (Z\upm)^T, \quad m = s(l), \ \linL,\  \minM,. 
\ee 
Here  $Z\upm \in \{0,1\}^{n \times K_m}$  is the clustering matrix in the layer of type $m = s(l)$  such that 
$Z\upm(i,k) = 1$ iff node  $i$ in the layer of type $m$ belongs to the community $k$,
and $B\upl$ is a signed symmetric $(K_m \times K_m)$ matrix of block connection probabilities, $\linL$.
Then, the COSIE model corresponds to the case where all layers have a common community structure 
while the new DIMPLE-SGRDPG model allows community structures to vary from one group of layers to another. However,  
for both models,  the block connectivity  matrices can be all different.

Although in the model above one can be interested in a variety of problems, such as 
recovery of the layers' label function $s$, estimation of the ambient  subspaces' bases  $U\upm$, 
or estimation of the loading matrices $Q\upl$, in this paper we shall focus only on the first task. 
Indeed, it is easy to see that model \fr{eq:DIMPLE_GDPG} constitutes a scrambled concatenation of $M$
independent COSIE models. Therefore, as soon as the groups of layers are identified,
the rest of the tasks can be accomplished by techniques developed for the COSIE model  by a variety of authors 
(see, e.g.,  \cite{JMLR:v22:19-558},  \cite{arxiv.2007.10455},  \cite{Xie_IEEE2024} 
and \cite{MinhTang_arxiv2022}  among others).


\subsection{Existing   results and our contributions} 
\label{sec:related_work}

\subsubsection{Previous results for the DIMPLE-SGRDPG model and our contributions}   
The DIMPLE-SGRDPG model was   studied in \cite{pensky2024signed},
where the authors  suggested a  strongly consistent algorithm for   
between-layer clustering on the basis of layer-by-layer analysis.
This result   was, however, derived under the condition that the layers of the 
network are not too sparse. In particular, if $\rhon$ is the sparsity of tensor $\bP$ 
(formally defined later in Section~\ref{sec:assump}), then it was assumed that 
$n \rhon /\log n \to \infty$ as $n \to \infty$, regardless of the number of layers $L$. 
This is a rather common assumption, even  in the simpler  multiplex network models 
with  persistent ambient subspaces, i.e., $M=1$ (see, e.g., 
\cite{JMLR:v22:19-558},  
\cite{arxiv.2007.10455},  
\cite{MinhTang_arxiv2022},  
among others). {\bf Our main contribution} is the demonstration that, while this condition is required for layer-by-layer analysis, 
it is not necessary for accurate  clustering of layers in the DIMPLE-SGRDPG network. 
Specifically,  we achieve perfect clustering  when $\rhon\, n\,  \minnL$ and $\rhon \, n \sqrt{L}$ grow faster than some specific power of $\log n$. 
As  \cite{Lei_Zhang_AOS2024_Comp_Lower_Bounds} shows, those constraints, up to  logarithmic factors, 
coincide with the computational lower bounds for a much simpler model than ours.
To the best of our knowledge, the only similar results have been established  in \cite{TWIST-AOS2079} and its sequel  
\cite{lyu2023optimalclusteringdiscretemixtures} for a much simpler, $k$-means-based 
model, MMLSBM \fr{eq:MMLSBM}, and only in the case where $L = O(n)$. In this paper, we allow $L$ 
to grow as any positive power of $n$. Strongly consistent clustering is proved  by applying 
new upper bounds for the errors in the two-to-infinity norm, derived in \cite{pensky2024daviskahan}.

\begin{figure*} [t!] 
\centering
\[\includegraphics[width=8.0cm, height=3.8cm]{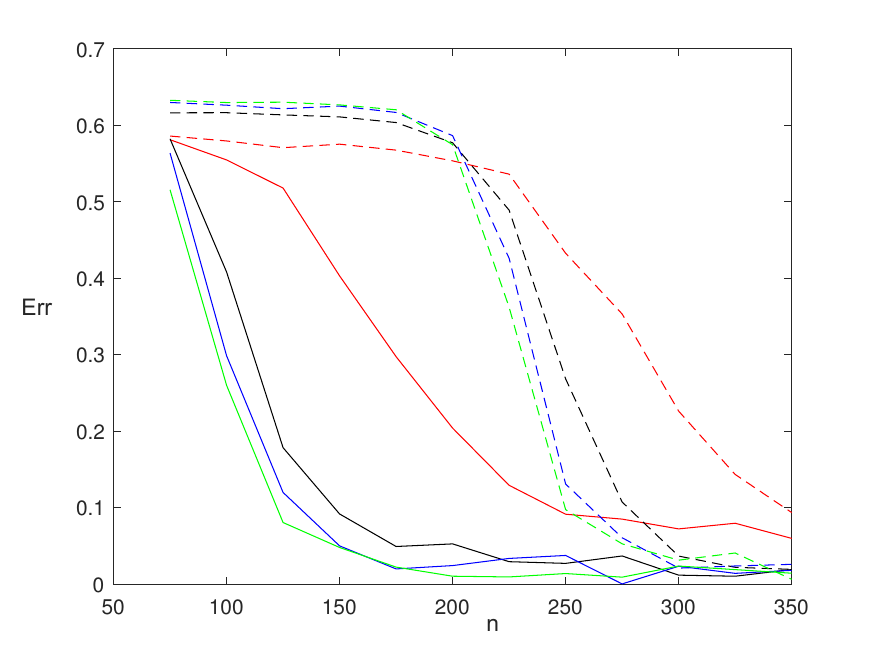}   
\includegraphics[width=8.0cm, height=3.8cm]{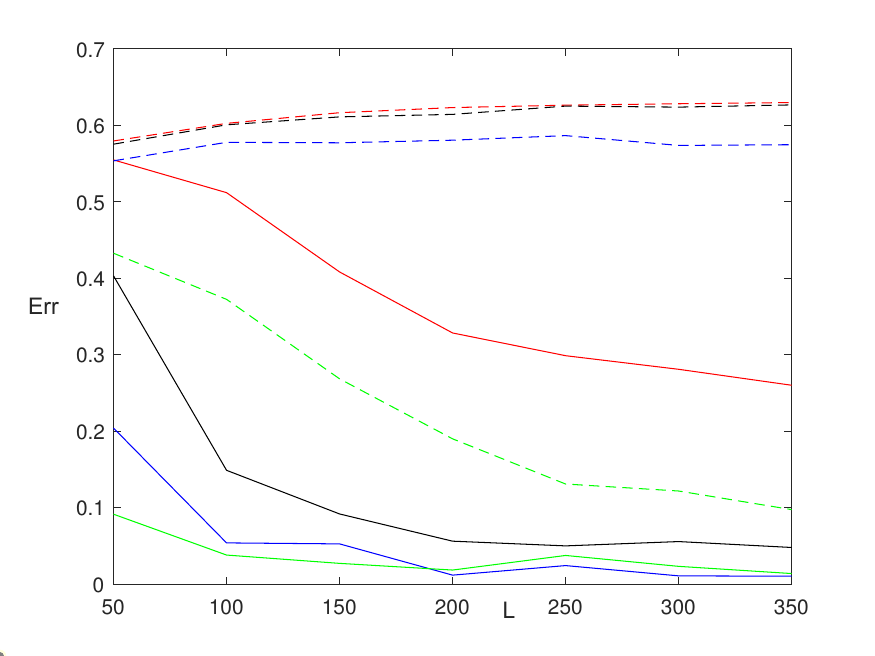}   \]%
\caption{\footnotesize{The between-layer clustering errors of the tensor-based  (solid lines) and the layer-per-layer analysis based 
technique (dash lines). 
Matrices $X\upm$  are generated by the truncated normal distribution  with $\sig =1$, $\mu =0$,   $M=3$ and $K_m=K=3$.
The entries of $B\upl$ are generated  as uniform random numbers  between   $c=-0.02$ and
$d = 0.02$. 
Left panel: varying $n$,   $L=50$ (red),  $L=150$ (black),   $L=250$ (blue), $L=350$ (green).
Right panel: varying $L$,  $n=100$ (red),  $n=150$ (black),   $n=200$ (blue), $n=250$ (green).
Errors are averaged over 100 simulation runs.  
}}
\label{fig:showcase}
\end{figure*}

Figure~\ref{fig:showcase} presents a computational comparison between   
our new tensor-based technique and the layer-by-layer algorithm of 
\cite{pensky2024signed} for a very sparse network. The left panel demonstrates that, until $n \, \rhon$ reaches the required threshold,
the latter technique is as good as pure guessing while the former delivers a progressively more accurate clustering.
In addition, as the right panel shows, the new tensor-based methodology 
takes   advantage of the growing number of layers $L$ while the layer-by-layer method cannot. 


\vspace{-5pt}

\subsubsection{Relation to other multilayer network models with groupwise subspace structure and our contributions}    
The DIMPLE-SGRDPG model generalizes two types of multilayer network models: the COSIE 
model, reviewed in the previous section,   and the Mixed MultiLayer Stochastic Block Model ({\bf MMLSBM}) 
\cite{fan2021alma}, \cite{TWIST-AOS2079}.
In the COSIE model,  all layers of the network are embedded in the same subspace, which corresponds to $M=1$ in 
\fr{eq:DIMPLE_GDPG} (see, e.g., \cite{JMLR:v22:19-558}, \cite{arxiv.2007.10455}, \cite{Xie_IEEE2024} and \cite{MinhTang_arxiv2022}). 
While the COSIE model does not enforce any layer clustering, in 
 the setting of the MMLSBM,   the  layers can be partitioned into $M$ groups,
with each group of layers  equipped with its own unique  SBM, so that 
\be \label{eq:MMLSBM}
 P\upl =  Z\upm  B\upm (Z\upm)^T, \quad m = s(l), \ \linL,\  \minM. 
\ee 
The MMLSBM is a particular case of   \fr{eq:SBM}, the reduction of \fr{eq:DIMPLE_GDPG} to the Stochastic Block Model,
where $B\upl \equiv B\upm$ depend  on $m = s(l)$ only, $\linL$.
It is easy to see that MMLSBM in \fr{eq:MMLSBM} imposes the $k$-means clustering on the layer networks.
For this scenario, \cite{fan2021alma}  and  \cite{TWIST-AOS2079}  showed that  
clustering can be successful under conditions that $\sqrt{L} n \rhon \geq C (\log n)^a$ for a specific constant $a$.

The MMLSBM  setting  is, however, much simpler than our model.  
Indeed, MMLSBM easily reduces to a $k$-means formulation for the network layers, which,
in turn, leads to its convenient representation via the Tucker decomposition with a low-dimensional core tensor. 
The partition of layers in this model can then be accomplished by straightforward clustering of 
the rows of the matrix of singular vectors associated with the layer dimension, as in \cite{TWIST-AOS2079},
or by Alternating Minimization Algorithm which is based on the $k$-means structure of the probability tensor, as in \cite{fan2021alma}. 
In addition, the layers of the MMLSBM are equipped with   SBMs rather than  SGRDPG-based networks.
Therefore, as \cite{pensky2021clustering} shows, none of the techniques designed for the MMLSBM works 
for the new, more flexible DIMPLE-SGRDPG model.

The three-way tensor $\bP$ in our paper reduces to a partial multilinear structure (see, e.g., \cite{JMLR:AZhang21}).
By carefully decomposing the tensor, we were able to recover the matrix of singular vectors $W$ 
associated with the layer dimension, which, however, is not directly related to the clustering assignment of the layers.
{\bf Our contribution} is   the development of a new methodology for extracting 
information from $W$, which makes clustering of layers into groups possible and leads to fast and scalable 
computational algorithms that  can be applied even for very large values of $n$ and $L$. As a result, our theoretical analysis   
is significantly more difficult than that in \cite{TWIST-AOS2079}. 

\vspace{-5pt}

\subsubsection{Relation to low-rank tensor  models and our contributions}   
In the DIMPLE-SGRDPG model and its particular cases, the COSIE model and the MMLSBM,
observations are generated on the basis of the probability tensor $\bP$.
In particular, in the case of the MMLSBM, the probability tensor 
$\bP$ has a simple Tucker decomposition, with the subspace structures associated with the problems of interest.

In the last  few years, many authors have studied tensor estimation and subspace recovery 
when the underlying tensor has a  Tucker decomposition with a low-rank core tensor
or a low-rank  Canonical Polyadic Decomposition (CPD). In the former case, the method of choice is the Higher Order 
Orthogonal Iteration (HOOI) algorithm studied by \cite{agterberg2022_HOOI},
 \cite{agterberg2024_Heteroscedastic},
 \cite{lyu2023optimalclusteringdiscretemixtures},
 \cite{LyuXia_2023},
 \cite{Xia_Zhang_AOS2022} and 
 \cite{Xia_IEEE2018},
to name a few. The CPD is usually computed using the Alternating Least Squares (ALS)
algorithm (see, e.g., \cite{Kolda09tensordecompositions}). 
The convergence of the ALS  algorithm  for the CPD of a noisy tensor 
has been recently proved in \cite{klopp2025_Tensor_CPD}.
Specifically, in the case of the MMLSBM,  \cite{TWIST-AOS2079} 
used a Tucker decomposition with a low-dimensional core tensor, and then recovered 
the subspaces and the core tensor using   HOOI.
The partition of the layers into groups was subsequently accomplished by straightforward clustering 
of the rows of the singular vector matrix associated with the layer dimension.
While we utilize the HOOI technique,
we apply it in a much less restricted environment than in, e.g., 
 \cite{agterberg2024_Heteroscedastic},  \cite{Xia_Zhang_AOS2022} and 
 \cite{Xia_IEEE2018}, where unitary factor matrices are associated with the problem under consideration.

{\bf Our contribution}  is to show how the Tucker decomposition with the HOOI algorithm can be 
used for a clustering task that is only loosely related to the tensor structure and to deliver 
a practical and computationally efficient methodology for doing this.
As such, our paper is only tangentially related to the publications that   establish  
fundamental statistical and computational limits for multilayer models such as
\cite{gong2026fundamentallimits2026}, \cite{Lei_Zhang_AOS2024_Comp_Lower_Bounds} and
\cite{Subhabr_Sen_Limits_Comm_Detect_AOS2025}.  
The common feature of  our paper and \cite{gong2026fundamentallimits2026}, \cite{Lei_Zhang_AOS2024_Comp_Lower_Bounds} and
\cite{Subhabr_Sen_Limits_Comm_Detect_AOS2025} is that   the network structure is described within a Bayesian framework. 
The main difference however is that, unlike the present paper,  the goal of those papers  is to derive sharp 
thresholds for community detection. As such, the main objective of the authors is the construction of statistical 
and computational lower bounds under the settings postulated in those papers.
This objective motivates very restrictive generating mechanisms: 
all these papers consider SBMs with two communities; \cite{Lei_Zhang_AOS2024_Comp_Lower_Bounds} assumes persistent communities 
throughout the layers of the network while \cite{Subhabr_Sen_Limits_Comm_Detect_AOS2025} studies a setting  
where communities in each of the layers deviate from the ``initial structure'' with some probability $\rho < 1/2$.
As a result, in both \cite{Lei_Zhang_AOS2024_Comp_Lower_Bounds}   and \cite{Subhabr_Sen_Limits_Comm_Detect_AOS2025} 
there is no group structure  among the layers.
\cite{gong2026fundamentallimits2026} allows for two types of layers but assumes that the number of layers $L$ is finite.

On the contrary, our paper does not pursue the derivation of lower bounds
in relatively artificial settings. We provide   a meaningful generalization of 
\cite{TWIST-AOS2079}, allowing for a significantly more flexible technique of   network 
generation mechanism than those considered in  \cite{gong2026fundamentallimits2026}, \cite{TWIST-AOS2079},
\cite{Lei_Zhang_AOS2024_Comp_Lower_Bounds} and
\cite{Subhabr_Sen_Limits_Comm_Detect_AOS2025}. 
The main contribution  of our paper is the development of a methodology for extracting the layer 
memberships in this more sophisticated network under the signal strength  assumptions 
similar to those in \cite{TWIST-AOS2079}. As a result, we are able to use the statistical and the computational lower bounds 
of \cite{TWIST-AOS2079}. Our main contribution is to construct a scalable computational algorithm and 
to prove that it attains those lower bounds up to logarithmic factors.

\vspace{-5pt}

\subsubsection{Relation to estimation of shared and unique subspaces over multiple data matrices and our contributions}    
The task of clustering layers in the model \fr{eq:DIMPLE_GDPG} can be viewed as 
a problem of discovering shared subspaces on the basis of noisy data matrices which has recently 
attracted a great deal of attention (see, e.g.,
\cite{baharav2025stackedsvdsvdstacked}, 
\cite{arroyo_2024_jointindividualstructure}, 
\cite{li2025_heterojive}, 
\cite{ma_ma_2024optimal_est_shared}, 
\cite{Gaynanova_Spectral_Multiview_Biom2026}, 
\cite{cong_ma_2025_AJIVE}  
 and \cite{MinhTang_arxiv2022}, 
 to name a few).
However, in the majority of those papers,  the authors  are interested only in discovering the fully shared 
subspace  \cite{baharav2025stackedsvdsvdstacked}, \cite{ma_ma_2024optimal_est_shared}, treating
the non-shared portions  as nuisance structures (even if they have some communalities). 
In the case where  unshared components are deemed to be of no interest, the analysis is 
usually based on a version of the AJIVE algorithm
 \cite{li2025_heterojive}, \cite{ma_ma_2024optimal_est_shared}, \cite{cong_ma_2025_AJIVE}
and is carried out by layer-by-layer spectral decompositions which impose stronger requirements on the 
signal strength. 
Alternatively, the methods in which partially shared subspaces are of interest such as 
\cite{arroyo_2024_jointindividualstructure}  and \cite{Gaynanova_Spectral_Multiview_Biom2026}  
are designed for small finite values of $L$ and do not provide scalable computational techniques for large values of $L$.

Our paper can be viewed as complementary to those efforts. In our model, the subspace structure  
is not shared by all matrices but rather by  groups of matrices.   {\bf Our contribution}  is the
derivation of the sufficient conditions for  perfect clustering of layer matrices when shared 
subspaces have a groupwise structure. It is a generalization of  the papers that assume that all 
layer matrices belong to the same subspace (see, e.g., \cite{lei2021biasadjusted} and    \cite{Xie_IEEE2024}).
In addition, our paper is the first to apply a tensor-based technique to estimation of shared and unique subspaces 
over multiple data matrices. While we tackle the case of symmetric Bernoulli matrices, 
 it is fairly trivial to extend our techniques and theoretical results to   situations 
involving sub-Gaussian or sub-exponential noise
and non-symmetric layer matrices (see Remark~\ref{rem:other_errors}).



\subsection{Notations}  
\label{subsec:notation}

We denote $a_n = O(b_n)$ if $a_n \leq C b_n$;  $a_n = \om(b_n)$ if $a_n \geq c b_n$;
$a_n \propto b_n$ if $c b_n \leq a_n \leq C b_n$,  where $0<c\leq C <\infty$ are absolute constants independent of $n$.
Also, $a_n = o(b_n)$  and $a_n = \Om(b_n)$ if, respectively, $a_n/b_n \to 0$ and  $a_n/b_n \to \infty$
as $n \to \infty$.  Denote $\min(a,b)  = a \wedge b$, $\max(a,b) = a \vee b$.
We use $C$ as a generic absolute constant independent of $n$, $L$, $M$ and $K$.

For any vector $v \in \RR^p$, denote  its $\ell_2$, $\ell_1$, $\ell_0$ and $\ell_\infty$ norms 
by $\|   v\|$, $\|   v\|_1$,  $\|   v\|_0$ and $\|  v\|_\infty$, respectively. 
Denote by $1_m$  the $m$-dimensional column vector with all components equal to one. 
For any matrix $A$,  denote its spectral and Frobenius norms by, respectively,  $\|  A \|$ and $\|  A \|_F$. 
Denote the eigenvalues and singular values of $A$ by $\lam(A)$ and $\sig(A)$, respectively.
Let $\SVD_r(A)$ be $r$ left leading eigenvectors of $A$.
The column $j$ and the row $i$ of a matrix $A$ are denoted by $A(:, j)$ and $A(i, :)$, respectively.
Let $\vect(A)$ be the vector obtained from matrix $A$ by sequentially stacking its columns. 
Denote by $A \otimes B$ the Kronecker product of matrices $A$ and $B$. 
Denote the diagonal of a matrix $A$ by $\diag(A)$. Also, denote the $K$-dimensional 
diagonal matrix with $a_1,\ldots,a_K$ on the diagonal by $\diag(a_1,\ldots,a_K)$.
A matrix $X \in \{0,1\}^{n_1 \times n_2}$ is a clustering matrix  if it is binary and has exactly one 1 per row.  
Denote 
$\calO_{n,K} = \left \{A \in \RR^{n \times K} : A^T A = I_K \right \}$,   $\calO_n=\calO_{n,n}$.

For any tensor $\bX\in\RR^{n_1\times n_2\times n_3}$, denote its mode-$k$ matricization by
$\calM_k(\bX)$. 
For any tensor $\bX\in\RR^{n_1\times n_2\times n_3}$ and a matrix $A\in\RR^{m\times n_1}$,
their mode-1  product $\bX\times_1\bA$ is a tensor in $\RR^{m\times n_2\times n_3}$ defined by
$\di [\bX\times_1 A](j,i_2,i_3)=\sum_{i_1=1}^{n_1}  \bX(i_1,i_2,i_3)   A(j,i_1), \  j \in [M]$.
The scalar product of tensors $\bX, \bY \in\RR^{n_1\times n_2\times n_3}$ is
$\di \lan \bX, \bY \ran = \sum_{i_1=1}^{n_1}  \sum_{i_2=1}^{n_2}  \sum_{i_3=1}^{n_3}  \bX(i_1, i_2, i_3) \bY(i_1, i_2, i_3).$
%
The Frobenius norm $\|\bX\|_F $ and the largest singular value $\sigma_1(\bX)$ of
a tensor $\bX\in\RR^{n_1\times n_2\times n_3}$ are defined as
$\di \|\bX\|_F   =\sqrt{\sum \ \lkv \bX(i_1,i_2,i_3)\rkv^2},\  
\sigma_1(\bX) = \|\bX\|    =\max_{\|u_i\|=1}   (\bX _{\times 1} {u_1} _{\times 2} {u_2} \tir u_3).$
%
In what follows, we shall use standard tensor identities (see, e.g., \cite{Kolda09tensordecompositions}).
Also, we denote an absolute constants independent of $n, K, L$ and $M$, which can take different values at different instances, by $C$
and $\Ctau$, where $\Ctau$ may depend on $\tau$.


\section{Tensor approach to the DIMPLE-SGRDPG model}
\label{sec:tens_approach}


\subsection{The DIMPLE-SGRDPG network formulation}
\label{sec:net_form}

In order for successful estimation and clustering in \fr{eq:DIMPLE_GDPG} to be possible,
one needs to ensure that   the subspace matrices $U\upm$ of the groups of layers are  not too similar.
While one can postulate  the necessary  assumptions similarly to, e.g., \cite{TWIST-AOS2079}, and then make  
the reader wonder how easy it is to satisfy them, we propose, in what follows, a simple, non-restrictive generative mechanism
for forming a network that guarantees successful inference.

For  obtaining SGRDPG networks in the layers, we follow the standard protocols of  \cite{JMLR:v22:19-558}.
For each layer $l \in [L]$, we generate its group membership as 
\be \label{eq:group_memb}
  s(l) \sim {\rm Multinomial}(\vec{\pi}), \quad 
  \vec{\pi}  = (\pi_1, ..., \pi_M) \in [0,1]^M, \ \ 
\pi_1 + ... + \pi_M = 1.  
\ee
In order to create the   probability tensor $\bP$, we generate $M$ matrices 
$X\upm \in [-1,1]^{n \times K_m}$, $\minM$, with i.i.d. rows:
\be  \label{eq:Xm_gen}
X\upm(i,:) \sim \iid \, f_m, \ \  \EE(X\upm(i,:)) = \mu\upm, \ \   
  \Cov(X\upm(i,:)) = \Sig\upm, \  \  i \in [n],  
\ee
where $f_m$ is a probability distribution on  a subset  of $[-1,1]^{K_m}$.
Subsequently, we consider  a collection of symmetric matrices $B\upl \in \RR^{K_m \times K_m}$, 
 $m = s(l)$, $\linL$, and   form the layer probability matrices $P\upl$ as 
\be \label{eq:DIMPnew_GDPG}
 P\upl =  X\upm  B\upl (X\upm)^T, \quad    B\upl = (B\upl)^T, \quad  m = s(l), \ l \in [L],\ m \in [M].
\ee
Since matrices $P\upl$ are usually sparse, we introduce  a sparsity factor $\rhon$ 
and impose sparsity via matrices $B\upl$, $\linL$.
Specifically, we write
\be \label{eq:spars_factor}
B^{(l)} = \rho_n B^{(l)}_0, \quad P^{(l)} = \rho_n P^{(l)}_0, \quad 0 < \rhon <1, \quad  \linL.
\ee
It is easy to see that \fr{eq:DIMPnew_GDPG} is equivalent to the model \fr{eq:DIMPLE_GDPG}.
Indeed, if $X\upm = U_X\upm D_X\upm  O_X\upm$ is the SVD of $X\upm$, then \fr{eq:DIMPnew_GDPG}
can be rewritten as  \fr{eq:DIMPLE_GDPG} with 
\be \label{eq:UupmQupl}
 U\upm = U_X\upm, \quad
Q\upl = D_X\upm  O_X\upm B\upl (O_X\upm)^T D_X\upm, 
\ \  m = s(l), \ \linL. 
\ee 
Observe also, that since $X\upm$ and  $B\upl$ are  defined up to a scaling constant, 
in what follows, we assume that    $X\upm(i,:) \in  \calB (K_m)$, and model sparsity via matrices $B\upl$, $\linL$. 
In addition,   we assume that matrices $\Sig\upm$ in  \fr{eq:Xm_gen} are non-singular. 
In the cases where the latter is not true, like in, e.g., the SBM, one can still apply our methodology, albeit with some minor modifications 
(see Remark~\ref{rem:SBM_MMM}).
There are many ways of generating rows $X\upm(i,:)  \in \calB(K_m)$ of matrices $X\upm$, 
$i \in [n]$, $\minM$, satisfying conditions \fr{eq:Xm_gen}.
Below are some examples, where we suppress  the index $i$:  
  
\vspace{4pt}
 
{\it Case 1:\ }   $f_m$ is a truncated normal distribution: $\xi \sim f_m$ is obtained as 
$\xi = \eta/\|\eta\|$  where $\eta \sim \calN(0, \Sig)$, the normal distribution 
with the mean 0 and covariance $\Sig$.

{\it Case 2:\ }  Generate  a vector $\eta \in [0,1]^{K+1}$ as $\eta \sim {\rm Multinomial} (\vec{\varpi})$, where
$\vec{\varpi}= (\varpi_1, ..., \varpi_{K+1})$, $\varpi_k > 0$, $\varpi_1 + ...+ \varpi_{K+1} = 1$.
Set $\xi$ to be equal to the first $K$ components of $\eta$.

{\it Case 3:\ }   Generate  a vector $\eta \in [0,1]^{K+1}$ as $\eta \sim {\rm Dirichlet} (\vec{\varpi})$,  where
$\vec{\varpi} = (\varpi_1, ..., \varpi_{K+1})$, $0\leq \varpi_k \leq 1$, $\varpi_1 + ...+ \varpi_{K+1} = 1$.
Set $\xi$ to be equal to the first $K$ components of $\eta$.
 
\vspace{4pt} 
 
Note that Cases~2 and 3 bring to mind, respectively,  the Stochastic Block Model (SBM) 
and the Mixed Membership Model (MMM), where the last column of the community assignment 
matrix $X\upm$ is missing. Indeed, in the case of the SBM or the MMM,
the rows of the matrix $X\upm$ sum to one, which means that the covariance matrix   $\Sig\upm$ is singular.
However,   our methodology can handle networks generated under 
the SBM or the MMM,  even though the matrices $\Sig\upm$ 
are singular in those cases (see Remark~\ref{rem:SBM_MMM}).

Although matrices $X\upm$ and, hence, $U_X\upm$ are different for different values of $m$, they are too 
similar for ensuring a reliable separation of layers of the network, due to the  common mean $\mu\upm$
of rows of $X\upm$, which implies that $\EE(X\upm) = 1_n  \mu\upm$, $\minM$.
For this reason, one needs to ``de-bias'' the layers by removing this common mean  and replacing $X\upm$ with 
$\tilX\upm = X\upm - 1_n  \bar{X}\upm$, where $\bar{X}\upm$ is the sample mean of the rows of $X\upm$.
By introducing a projection operator $\Pi= n^{-1} 1_n 1_n^T$,   rewrite $\tilX\upm$ as
\be    \label{eq:tX}
\tilX\upm =   (I_n - \Pi)\,   X\upm = \Pi^{\bot} X\upm, \quad \Pi = n^{-1} 1_n 1_n^T.  
\ee 
It is easy to see that for $\minM$,\   $\linL$ 
\be \label{eq:tilbp1}
\tilP\upl = \Pi^{\bot} P\upl  \Pi^{\bot} = \tilX\upm   B\upl (\tilX\upm)^T, \ \  s(l)=m.
\ee 
Introduce tensors $\tilbP$ and $\tilbA$ with slices $\tilP\upl = \Pi^{\bot} P\upl  \Pi^{\bot}$ and 
$\tilA\upl = \Pi^{\bot} A\upl  \Pi^{\bot}$, respectively, $\linL$, so that 
\be  \label{eq:tilbPA} 
\tilbP  = \bP \tio \Pi^{\bot} \tit \Pi^{\bot}, \quad 
\tilbA  = \bA \tio \Pi^{\bot} \tit \Pi^{\bot}.
\ee
Consider the SVDs of $\tilX\upm$ and $\tilP\upl$:
\be \label{eq:X_P_svds}
\tilX\upm = \tilU\upm \tilD\upm  (\tilO\upm)^T, \ \   \tilP\upl = \tilU_P\upl D_P\upl  (\tilU_P\upl)^T, \quad 
\tilO\upm \in \calO_{K_m}.
\ee
Below we show that $\tilX\upm$ and $\tilU\upm$ are related via the low-dimensional sample covariance matrix
of the rows of $X\upm$ (which coincides with the sample covariance matrix
of the rows of $\tilX\upm$). 
Indeed, due to the relation $\tilP\upl = \tilU\upm \tilB\upl \tilU\upm$, 
where $\tilB\upl = \tilD\upm (\tilO\upm)^T   B\upl  \tilO\upm \tilD\upm$,
one can rewrite the SVD of  $\tilP\upl$ using the SVD of $\tilB\upl$:
\be    \label{eq:tilbp2} 
\tilB\upl = \tilU_B\upl \tilD_B\upl (\tilU_B\upl)^T,  \quad    
\tilP\upl = \tilU\upm  \tilU_B\upl \tilD_B\upl  (\tilU\upm  \tilU_B\upl)^T, \quad  \tilU_B\upl \in \calO_{K_m}.
\ee  
%
%
Consider  the sample covariance matrices $\hSig\upm$ with $\EE(\hSig\upm) = \Sig\upm$. If $n$ is large enough,
matrices  $\hSig\upm$ are non-singular, and it follows from \fr{eq:X_P_svds} that for  $\tilO\upm \in \calO_{K_m}$
\be \label{eq:hSig_def}
\hSig\upm =  (n-1)^{-1}\, (\tilX\upm)^T \tilX\upm 
=  (n-1)^{-1}\, \tilO\upm\, (\tilD\upm)^2  (\tilO\upm)^T.
\ee 
Let $\lkr  \hSig\upm \rkr^{1/2}  = (n-1)^{-1/2}\, \tilD\upm  (\tilO\upm)^T$ and 
$\lkr  \hSig\upm \rkr^{-1/2} = (n-1)^{1/2}\, \tilO\upm (\tilD\upm)^{-1}$. Then, 
whenever  $s(l)=m$,
\be \label{eq:tilUX}
\tilU\upm = (n-1)^{-1/2}\, \tilX\upm\, \lkr  \hSig\upm \rkr^{-1/2}, \quad
\tilB\upl = (n-1)\, \sqrt{ \hSig\upm} B_0\upl \lkr \sqrt{ \hSig\upm} \, \rkr^{T}.
\ee


\subsection{Assumptions}
\label{sec:assump}

Observe  that, since $X\upm$ and  $B\upl$ are  defined up to a scaling constant, 
in what follows, we assume that    $X\upm(i,:) \in  \calB (K_m)$, where $\calB (h)$  is the $h$-dimensional unit ball, 
and model sparsity via matrices $B\upl$, $\linL$. 
We impose the following assumptions: 
 
\vspace{4mm} 

\noindent
{\bf A1.}  Vector $\vec{\pi}  = (\pi_1, ..., \pi_M) $ is  such that  
$\lowc_{\pi}\, M^{-1}  \leq \pi_m \leq \highc_{\pi}\, M^{-1}$, $\minM$.   
 
\vspace{4mm}

\noindent
{\bf A2.} Distributions $f_m$ with means $\mu\upm$ and covariance matrices $\Sig\upm$, $\minM$, 
are supported on the $K_m$-dimensional unit balls $\calB (K_m)$. 
Here, $\displaystyle {K =  \max_m K_m}$ and $\displaystyle {K \leq C_K  \min_m K_m}$ for some  constant  $C_K$.
 
\vspace{4mm}

\noindent
{\bf A3.}  Covariance matrices $\Sig\upm$  satisfy 
\be \label{eq:lam_Sig} 
0 < \lowc  \leq \lam_{\min} (\Sig\upm)  \leq \lam_{\max} (\Sig\upm) \leq  \highc < \infty, \ \  \minM .
\ee

\noindent
{\bf A4.} For some positive constants $\underline C$ and $\bar C$, $0 < \underline C \le \bar C < \infty$ , one has
\be \label{eq:Bol}
B^{(l)} = \rho_n B^{(l)}_0 \quad \mbox{with} \quad 
0 < \lowC \leq \sig_{\min}(B^{(l)}_0) \leq \sig_{\max}(B^{(l)}_0) \leq \highC < \infty
\ee 
where, for some constant $\tilC>0$
\be \label{eq:rhon_main_cond}
\rhon^2\, n^2 \, L \geq \tilC \, \log n
\ee

\noindent
{\bf A5. } Matrices   $\Phi\upm \in \RR^{L_m \times K_m^2}$ with rows $[\vect(B_0\upl)]^T$, $s(l) =m$,
are of the respective ranks $r_m$, where $1 \leq r_m \leq K_m (K_m +1)/2$, and, for some constant $\lowC_{\Phi}>0$, one has 
\be \label{eq:low_sig_Psiupm}
\min_m \, \sig_{r_m}^2 (\Phi\upm) \geq \lowC_{\Phi} \, M^{-1}\,L.
\ee

\noindent
{\bf A6.}  The number of layers $L$ is such that, for some constants $\tilc$ and 
 $\tau_0$, one has
\be \label{eq:nLtau}
\tilc\, (\log n)^2\,  \leq L \leq n^{\tau_0}, \quad \tilc >0, \ \  \tau_0 < \infty .
\ee

\noindent 
Note that Assumption~{\bf A1} ensures that the sizes $L_m$ of groups of layers with the common ambient subspace structures are balanced.
Specifically, Lemma~5
in the Suppelementary Material  asserts that if $L \geq 2 \lowc_\pi^{-2}  M^2  \log(2 \, M\, n^{\tau})$,
then, with probability at least $1 - n^{-\tau}$, simultaneously for all $\minM$, one has 
\be \label{eq:hLm}
 \lowc_{\pi}\, L/(2\, M)  \leq  L_m \leq  3\, \highc_{\pi}\ L/(2\, M),
\ee
where $L_m$ is the number of layers of type $m$. Hence, \fr{eq:hLm} is true as long as $L$ grows at least logarithmically 
with respect to $n$.
This is a reasonable condition since,  for a finite $L$, analyzing layers together does not lead to 
an asymptotically significant gain in accuracy.


Assumptions  {\bf A1}-{\bf A4}, which  are very common, are also easy to ensure. 
In particular, Assumption  {\bf A1} as well as the conditions on 
$K_m$  are always satisfied if $M$ is a constant independent of $L$ and $n$.  
Assumptions {\bf A2} and  {\bf A3} are non-restrictive and 
are true for the majority  of probability distributions.
Some versions of Assumption {\bf A4} appear in the vast majority of papers dealing with the GRDPG and its particular cases.
However, the lower bound \fr{eq:rhon_main_cond}  for the sparsity parameter is much lower than that in, e.g. 
\cite{pensky2021clustering} or   \cite{MinhTang_arxiv2022},  
and is motivated by the computational lower bound  derived in \cite{Lei_Zhang_AOS2024_Comp_Lower_Bounds}.
We should point out here  that the  constants in \fr{eq:lam_Sig}  and \fr{eq:Bol}. 
can be modified to depend on $m$. However, this will make the notation  too cumbersome. For this reason, we 
impose uniform bounds in Assumptions~{\bf A3}  and {\bf A4}. Again, if  $M$  and $K_m$, $\minM$, are constants 
independent of $L$ and $n$, those restrictions are easily justifiable.

The goal of Assumption~{\bf A5} is to ensure that the matrix $W$ of  bases vectors of the subspace,
associated with the layer dimension  has a low two-to-infinity norm and, on the other hand, that
the scalar products of its rows  belonging to the same group are above the noise level. Since we would like
to allow a variety of scenarios, ranging from matrices $B_0\upl$ being the same for all layers in the same 
group to matrices being generated independently at random for every layer, we enforce the required 
properties of the matrix $W$ by regulating the variability of the matrices  $B_0\upl$.
Specifically, we assume that  the matrices $\Phi\upm$  with rows  formed by the vectorized versions of the
matrices $B_0\upl$  in group $m \in [M]$ are well conditioned, and prove that this translates into the 
desired properties of the matrix $W$.
While this Assumption~{\bf A5} is the most unusual among the six, it is not hard to ensure its validity, due to the fact
that $r_m$ can take any value between 1 and $K_m (K_m +1)/2$. 
Assumption~{\bf A5}   will be discussed at length in Section~\ref{sec:assump_A5}.
Specifically, Assumption~{\bf A5} can be satisfied by fixing the same $B_0\upl$ for all layers with $s(l) = m$, 
or randomly sampling matrices  $B_0\upl$.

The lower bound for $L$ in Assumption~{\bf A6} means that, in this paper, 
we are interested in the case where  $L$ is fairly large but grows at most polynomially with $n$.
The upper bound in \fr{eq:nLtau} is hardly restrictive and is also relatively common. Indeed,    
\cite{TWIST-AOS2079} assumed  that $L \leq n$, so, in their paper, 
\eqref{eq:nLtau}  holds with $\tau_0=1$.


\section{Probability tensor and its   properties}
\label{sec:tensor_components}


\subsection{Structure of the probability tensor}
\label{sec:tensor_structure}

Observe that matrices $B\upl$ in \fr{eq:DIMPnew_GDPG} can be all different, 
and the same is true for $\tilB\upl$ in \fr{eq:tilbp2}. Therefore, 
  tensor $\tilbP$ with layers $\tilP\upl$ in \fr{eq:tilbp2} seems to fall into the category of   partial multi-linear low rank tensor
models, which guarantee small error rates only for the low rank modes (see Theorem~4 of Luo {\it et al.} \cite{JMLR:AZhang21}). 
However, this is not true. Below, we show that $\tilbP$ has a low-rank Tucker decomposition.

\begin{figure*}[t!]  
\centering   
\[\includegraphics[width=438pt]
{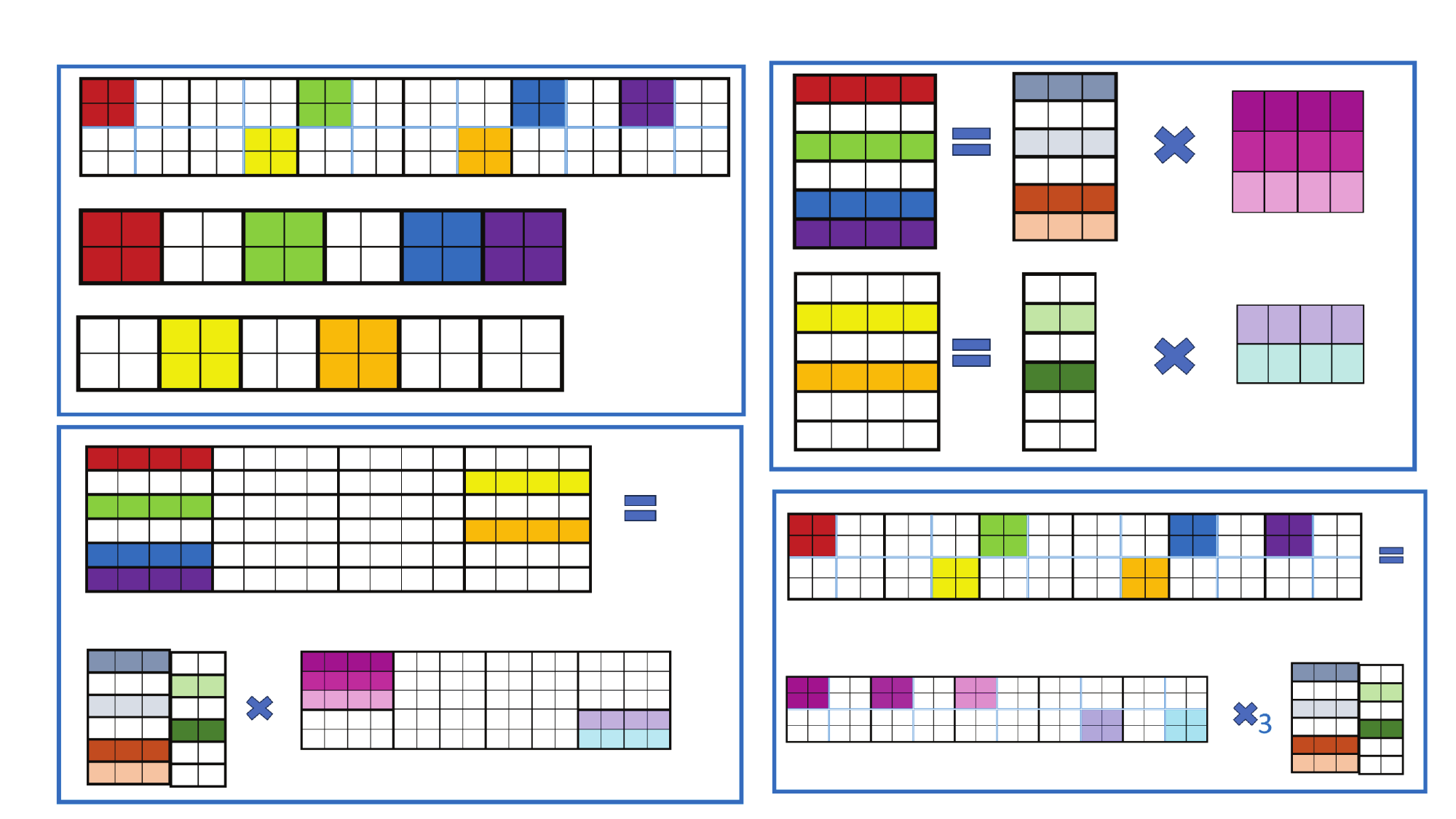}  \]
\caption{{\bf Tensor structure} with $M=2$, $L=6$, $L_1=4$, $L_2 =2$, $K_1 = K_2 = \barK = 2$, 
$r_1 = 3$, $r_2=2$, $r=5$.  
{\bf Tensors are  exhibited by stacking layers  side by side.}
White color means zero value but other colors  are used  for objects' identification only.
{\bf Identical colors do not mean identical element values}. 
{\bf Top left panel:}   tensors $\bG$ (top row), $\tilbG^{(1)}$ (middle row) and 
$\tilbG^{(2)}$ (bottom row) (tensors' layers stacked side by side). 
{\bf Top right panel:} decompositions  $\calM_3 (\tilbG\upm) = W\upm (H\upm)^T$ with $m=1$  (top row)
and $m=2$ (bottom row). 
{\bf Bottom left panel:} decompositions  $\calM_3 (\bG) = W   H^T$.
{\bf Bottom right panel:} decompositions $\bG =  \bH \tir W$ (tensors' layers stacked side by side).
}
\label{fig:tensor}
\end{figure*}

It follows from \fr{eq:DIMPnew_GDPG} that,  by concatenating matrices $\tilU\upm$, $\minM$, 
one can construct a common basis for the left and the right ambient subspaces of the matrices $\tilP\upl$. Denote
\be \label{eq:barU_K} 
\barU = [\tilU^{(1)}|...|\tilU^{(M)}] \in \RR^{n \times M \barK}, \quad \barK = M^{-1} \, (K_1 + ...+ K_M). 
\ee
Then, matrices $\tilP\upl$ can be written as $\tilP\upl = \barU G\upl \barU^T$, where $G\upl \in \RR^{M \barK \times M \barK}$ 
are  block-diagonal matrices $G\upl = \diag \lkr (G\upl)_{1,1}, \ldots, (G\upl)_{M,M} \rkr$  with diagonal blocks 
\be \label{eq:Guplmm1}
(G\upl)_{m,m} = \tilB\upl \in \RR^{K_m \times K_m} \  \mbox{if}\quad s(l)=m, \  (G\upl)_{m,m} = 0 \  \mbox{otherwise}, \  \linL. 
\ee 
Introduce a tensor $\bG$ with slices $G\upl$. Then
\be \label{eq:tensP}
\tilbP = \bG\tio \barU \tit\barU,  \quad  \bG \in \RR^{M \barK \times M \barK \times L}, 
\quad \bG(:,:,l) = G\upl, \ \ \linL.
\ee
Note that it follows from \fr{eq:tensP} that mode 3 matricization of tensor $\tilbP$ is of the form 
\bes
\calM_3 (\tilbP) = \calM_3 (\bG) (\barU \otimes \barU)^T, \quad \calM_3 (\tilbP)  \in \RR^{L \times n^2}, \quad
\calM_3 (\bG) \in \RR^{L \times (M \barK)^2}.
\ees
Let  $r = \rank(\calM_3 (\bG))$ and let $W \in \calO_{L,r}$ be 
the matrix of $r$ left eigenvectors of $\calM_3 (\bG)$. Then, 
for   $H \in \RR^{(M \barK)^2 \times r}$ one has
\be \label{eq:M3G}
\calM_3 (\bG) = W H^T, \quad H  = W^T\, \calM_3 (\bG), 
\quad r = \rank(\calM_3 (\bG)) \leq \min[L, (M \barK)^2].
\ee
Let $\tilbG\upm \in \RR^{K_m \times K_m \times L}$ be the sub-tensor of $\bG$ with 
the slices $(G\upl)_{m,m}$.  Then, one has $\tilbG\upm(:,:,l)\equiv 0$ if $s(l) \neq m$ (see the top left panel of Figure~\ref{fig:tensor}). 
Moreover,  using $W\upm = \SVD_{r_m} (\calM_3 (\tilbG\upm)) \in \calO_{L,r_m}$ and $H\upm \in \RR^{K_m^2 \times r_m}$, matrix  
$\calM_3 (\tilbG\upm) \in \RR^{L \times K_m^2}$ can be presented as 
\be \label{eq:WH_present}
 \calM_3 (\tilbG\upm) = W\upm (H\upm)^T,  
\quad r_m = \rank(\calM_3 (\tilbG\upm)),
\ee 
and matrices $W\upm$ are such that $W\upm(l,:)=0$ if $s(l) \neq m$ 
(see the top right panel of Figure~\ref{fig:tensor}).  
Combining them into matrix $W$, obtain $\calM_3 (\bG) = W H^T$ (see the bottom left panel of Figure~\ref{fig:tensor}). 

Now, consider tensors $\bH\upm \in \RR^{K_m \times K_m \times r_m}$, $\minM$, and 
$\bH  \in \RR^{ M \barK \times  M \barK  \times r}$
which are, respectively,  obtained by packing matrices $H\upm \in \RR^{K_m^2 \times r_m}$, $\minM$, 
and matrix $H$ into the tensors (see the bottom right panel of Figure~\ref{fig:tensor}). 
Then, formula \fr{eq:WH_present} leads to 
\be \label{eq:tens_pres}
\calM_3(\bG) = W \, \calM_3(\bH), \quad
\bG =  \bH \tir W,  \quad 
\tilbG\upm =  \bH\upm  \tir W\upm, 
\ee
where $W \in \calO_{L,r}$ and $W\upm \in \calO_{L,r_m}$.
Let   $\barU = U \Lam_U O_U^T$ be the SVD of  $\barU$ in \fr{eq:barU_K}.
It turns out that \fr{eq:M3G}--\fr{eq:tens_pres} lead  to a low-rank Tucker decomposition of tensor $\tilbP$.
which is given by the following statement.

\begin{lem}  \label{lem:Tucker_for_tilbP}      
Tensor $\tilbP$ in \fr{eq:tensP} is of the form  $\tilbP = \bH \tio \barU \tit \barU \tir W$ and 
has a low-rank Tucker decomposition with a core tensor $\bTe$
\be \label{eq:bP_total}
\tilbP = \bTe \tio U \tit U \tir W,
\quad U \in \calO_{n, M \barK},\  W \in \calO_{L,r},\ \bTe \in \RR^{M \barK \times M \barK \times r}
\ee
where 
\be \label{eq:bTe}
\bTe = \bH \tio (\Lam_U O_U^T) \tit (\Lam_U O_U^T) \in \RR^{M \barK \times M \barK \times r}, \quad
\bH \in \RR^{M \barK \times M \barK \times r}, \ \  O_U \in \calO_{M \barK}.
\ee 
Here,  
$r_m \leq K_m(K_m+1)/2, \quad \di \sum_m  r_m = r $
and  matrix $W$ is formed by scrambled  concatenation of matrices $W\upm  \in \RR^{L \times r_m}$,   where 
\be \label{eq:W_str}
W = [W^{(1)}| ...| W^{(M)}], \quad (W^{(m_1)})^T (W^{(m_2)}) = 0, \quad \rm{if}\  m_1 \neq m_2.
\ee
\end{lem}

Lemma~\ref{lem:Tucker_for_tilbP} asserts that  $\tilbP$ is the low rank tensor with the core tensor $\bTe$.
However, we emphasize that matrix $W$ in \fr{eq:bP_total} is not directly related to 
clustering of layers.


\subsection{Properties of layer   matrices  U  and  W}
\label{sec:matrU_W}

Note that the success of clustering the tensor layers relies on 
discovering dissimilarities between rows of the matrix $W$ for different groups of layers
and also on the accuracy of estimating these matrices based on the adjacency tensor $\bA$.
For this reason, in this section we establish the properties of the matrices $U$ and $W$.

\vspace{-4pt}

\begin{lem}  \label{lem:matr_U_LamU}      
Let Assumptions {\bf A1} - {\bf A3} hold  and $n$ be large enough. 
Let  $\tilP\upl = \tilU_P\upl \tilD\upl (\tilU_P\upl)^T$ be the SVDs of $\tilP\upl$, $\linL$, and 
 $\barU = U \Lam_U O_U^T$ be the SVD of $\barU$.
Assume that $M^4\, n^{-1} \log n \to 0$ as $n \to \infty$.
Then, if $n$ is large enough,  simultaneously for all $l_1, l_2 \in [L]$, with probability at least $1 - c\, n^{-\tau}$,
one has  $\rank(\barU) = \barK\, M$ and 
\beqn
& \|  (\tilU_P^{(l_1)})^T \tilU_P^{(l_2)} \| =1 \hspace{22mm}   {\rm if} \   s(l_1) = s(l_2); \nonumber\\ 
& \|(\tilU_P^{(l_1)})^T \tilU_P^{(l_2)} \| \leq C    n^{-1/2} \sqrt{\log n}\ \ \ 
  {\rm if} \   s(l_1) \neq s(l_2). \label{eq:matrU}
\eeqn
\be \label{eq:LamU}
1 -   C\,  M^2\, n^{-1/2} \sqrt{\log n} \leq \lam^2_{\min}(\Lam_U)
\leq \lam^2_{\max}(\Lam_U)  \leq 1 +    C\, M^2\,  n^{-1/2} \sqrt{\log n},
\ee 
\be \label{eq:for_del1}
\|U\|_\twin \leq C\, n^{-1/2}\,  M^{1/2}.  
\ee 
\end{lem}


\begin{lem} \label{lem:matr_W_new}  
Let Assumptions~{\bf A1} -- {\bf A6} hold.  Then, 
\be \label{eq:W2inf}  
\lan W(l_1,:),  W(l_2,:)\ran = 0 \quad \mbox{if}\quad s(l_1) \ne s(l_2), \quad 
 \|W\|_{2, \infty} \leq \highC\, \lowC_\Phi^{-1/2}\, \sqrt{M\, K}/\sqrt{L}.  
\ee
Let one of the conditions (a) or (b) holds:  
(a)\  $r_m = \tilK_m$ where $\tilK_m \leq K_m (K_m +1)/2$ is the size of the 
common support of matrices $B_0\upl$ with $s(l) = m$;   
(b)\  values of $B_0\upl$ with $s(l)=m$  come from a   dictionary $\lfi\calB_{0,m,t}\rfi_{t=1}^{t_m}$  
of a constant size $t_m \leq t_0$  with $t_m \leq K_m (K_m +1)/2)$, $\minM$, 
and the number $L_{t,m}$ of matrices  of type $t \in [t_m]$, $s(l)=m$, is such that 
\be \label{eq:Ltm}
\lowC_{t,0}\,  M^{-1}\,  L \leq  L_{t,m} \leq \highC_{t,0}\, M^{-1}\,  L.
\ee
Then, for $C$ that depends on $t_0$,  $\lowC_{t,0}$ and the constants in Assumptions~{\bf A1} -- {\bf A6} only,
\be \label{eq:sc_prod_W_same}
|\lan W(l_1,:),  W(l_2,:)\ran| \geq C \,  M\,L^{-1}  \quad {\rm if} \quad s(l_1) = s(l_2).
\ee 
\end{lem}

Lemma~\ref{lem:matr_W_new} implies that one can obtain a clustering assignment of layers by 
examining the matrix of scalar products $\lan W(l_1,:),  W(l_2,:)\ran$ for $l_1, l_2 \in [L]$. 
Therefore, while  it is impossible to partition the layers of the network by clustering the rows of the matrix $W$, 
Lemma~\ref{lem:matr_W_new} paves the way for successful clustering.



\subsection{Validity of Assumption {\bf A5}}
\label{sec:assump_A5}

As it was mentioned in Section~\ref{sec:assump}, Assumptions~{\bf A1}--{\bf A4} 
are quite common, while Assumption~{\bf A5} is not.  However, the validity of {\bf A5} is easy to ensure by   
considering the case where the matrices $B_0\upl$ are  randomly sampled from some distribution. 
Specifically, we introduce the following assumption.
\\

\noindent
{\bf A5*. } 
For $\minM$, vectorized versions $b_0\upl \in [0,1]^{K_m(K_m+1)/2}$ of the 
upper diagonal portions of matrices $B_0\upl$ with $s(l)=m$, are generated independently 
according to the pdf/pmf $q_m$,  supported on a set of $\tilK_m$-dimensional vectors, where $\tilK_m \leq K_m(K_m+1)/2$. 
Denote  $\EE(b_0\upl) = \mu_0\upm$, $\Psi\upm = \Cov(b_0\upl)$, the mean and the 
covariance matrix of nonzero portions of vectors $b_0\upl$ with $s(l)=m$.
Assume that 
\be \label{eq:assump_A5}
0 < \lowd_0 \leq \|b_0\upl\|  \leq \highd_0.
\ee 

\medskip

Unlike Assumption~{\bf A5}, Assumption~{\bf A5*} explicitly regulates   variability of the matrices  $B_0\upl$, $\linL$,
and it is very non-restrictive. 
It turns out that   Assumption~{\bf A5*} guarantees the validity of Assumption~{\bf A5} with high probability.
Specifically, the following statement is true.

\begin{lem} \label{lem:matr_Phim}  
Let Assumptions~{\bf A1} -- {\bf A4}, {\bf A5*} and {\bf A6} hold.
If $r_m=1$ and $\lowd_0>0$,  or  
\be \label{eq:eig_Psi_upm}
r_m \geq 2 \quad {\rm and} \quad \min_m \sig_{r_m}(\Psi\upm) > \lowlam_0   >0, 
\ee 
then,   for any $\tau>0$ and some constants $c$ and $C$ that depend on $\tau$ and 
the constants in Assumptions~~{\bf A1} -- {\bf A4}, {\bf A5*} and {\bf A6} only,   one has 
\be \label{eq:sig_hPsiupm}
\PP \lfi \min_m \, \sig_{r_m}^2 (\Phi\upm) \geq C \, M^{-1}\,L  \rfi    \geq 1 - c\, (L/M)^{-\tau}.
\ee 
%
\end{lem}

\noindent
Below, we consider several examples where  Assumption~{\bf A5*} and conditions of Lemma~\ref{lem:matr_W_new}  hold. 
The details of these justifications are provided in 
Section~8.4 of the Supplementary Material.

\medskip

\begin{ex}  \label{example1}
{\rm Let $q_m = \del(\tilb_0\upm)$ be a mass at a point $\tilb_0\upm$ where $0 < \lowd_0 \leq \|\tilb_0\upm\|  \leq \highd_0$.
In this case, for any $\minM$, matrix $\tilPhi\upm$ has $L_m$ identical rows $\tilb_0\upm$, and $r_m =1$.
This setting corresponds to the situation where each group of layers has a 
unique matrix $B\upl \equiv B\upm$, where $m = s(l)$. If the layers are equipped with SBMs, 
the setting reduces to the MMLSBM of \cite{fan2021alma} 
and  \cite{TWIST-AOS2079}. 
}
\end{ex}


\begin{ex}  \label{example2}
{\rm Consider the case where, for $s(l)=m$, the values of   $B_0\upl$ are sampled from a finite dictionary $\frB_0\upmt$, $t \in [t_m]$,
at random with probabilities $\varpi_t\upm$, and matrices  $\frB_0\upmt$  satisfy Assumption~{\bf A4} for $t \in [t_m]$, $\minM$.
Then,  in Assumption~{\bf A5*}, one has 
\bes 
q_m = \sum_{t=1}^{t_m} \varpi_t\upm \, \del(\frb_0\upmt), \quad \sum_{t=1}^{t_m} \varpi_t\upm =1, 
\ees
Let $\frb_0\upmt$ be vectorized versions of the upper triangular portions of matrices $\frB_0\upmt$, $t \in [t_m]$, with
common supports of size $\tilK_m \in [K_m, K_m(K_m+1)/2]$, $\minM$. Consider matrices  
$\calQ\upm$ with rows $\calQ\upm(t,:) = \frb_0\upmt$, $t \in [t_m]$. 
Assume that in this setting, for $\minM$ and some absolute constants $t_0$, $\lowvarpi_0$  and $\tilsig_0$,
one has $t\upm \leq \min(t_0, \tilK_m)$, $\varpi_t\upm \geq \lowvarpi_0$ and $\sig_{t_m} (\calQ\upm) \geq \tilsig_0$.
Then, Assumption~{\bf A5} is satisfied with high probability and Lemma~\ref{lem:matr_W_new}  holds. 
}
\end{ex}


\begin{ex}  \label{example3}
{\rm Consider the case where, for $s(l)=m$, the entries of  the upper portion of  $B_0\upl$ are generated independently as follows:
\bes
B_0\upl(k_1, k_2) \sim \lfi
\begin{array}{ll}
\Uniform(\lowa,\higha), & \mbox{if}\ k_1 = k_2,\\
\Uniform(\lowa,\higha) \times \om, & \mbox{if}\ k_1 < k_2.\\
\end{array} \right.
\ees
Here $\Uniform(\lowa,\higha)$ is uniform distribution on the interval $[\lowa,\higha]$, $\om \in \RR$, and
and values of $\lowa, \higha$  and $\om$ are such that the elements of $P\upl$ lie between -1 and 1.
Since all rows of matrices $\tilPhi\upm$ are different, one has $r_m = \tilK_m$, 
where $\tilK_m = K_m(K_m+1)/2$ if $\om >0$ and  $\tilK_m = K_m$ if $\om =0$,
the latter case corresponding to the situation, where all matrices $B_0\upl$ are  diagonal.
Then, Assumption~{\bf A5*} and Lemma~\ref{lem:matr_W_new}  hold.
}
\end{ex}

\medskip


\begin{rem} \label{rem:SBM_MMM}  
{\bf  The  Stochastic Block Model and the Mixed Membership Model.}
{\rm
Observe that in the case of the  multiplex network,  where layers are equipped with the SBM or the MMM,   
one has $X\upm = Z\upm$, where  $Z\upm$ is a membership matrix.
Therefore, each row of matrix $Z\upm$ sums to one. The latter leads to  all
covariance matrices $\Sig\upm$ being  singular. Nevertheless, 
Algorithm~\ref{alg:between}  can still be 
successfully applied  with $K\upl$ replaced by $K\upl -1$, and $K_m$ replaced by $K_m-1$,
as long as  $\min K_m \geq 2$. Repeating the steps in Section~3.4 of \cite{pensky2024signed},
one can transform the theoretical results in Section~\ref{sec:theory} to this case.
}
\end{rem}


\section{Estimation and clustering techniques }
\label{sec:est_clust}


\subsection{Clustering of network layers}
\label{sec:layer_clust}

Unfortunately, the low-rank tensor representation \fr{eq:bP_total} does not provide 
an easy solution to the layer clustering problem, which is the key task in the analysis of the model. 
Recall that, in the case of the MMLSBM in  \cite{fan2021alma}  and  \cite{TWIST-AOS2079},
for each  group of layers,  matrix $W$ has identical distinct rows, and, therefore, the 
group assignment of layers could be carried out by a version of $k$-means clustering
of the rows of the matrix $W$.
This, however,  is no  longer true in the case of our model. Indeed, since 
matrix $W$ is defined only up to an orthogonal rotation, its estimated version will not abide 
by the ``zero-nonzero'' patterns presented in Figure~\ref{fig:tensor}. Moreover, the rows of $W$,  
that correspond to the same group of layers,  are all different and can be further away from each other 
than from the rows, corresponding to another group of layers.  
For this reason, to separate  the layers into groups,
one cannot apply  $k$-means clustering  to  the rows of $W$: a different approach is necessary.  

%
\begin{algorithm} [t] 
\caption{\ The between-layer clustering}
\label{alg:between}
\begin{flushleft} 
{\bf Input:} Estimated basis matrix $\hW \in \calO_{L, r}$; number of groups of layers $M$;  
threshold  $T$; parameter $\eps$ \\
{\bf Output:} Estimated clustering function $\hs: [L] \to [M]$ for groups of layers\\
{\bf Steps:}\\
{\bf 1:} Form matrix $\hY$ with elements given by \fr{eq:Y_hY} and matrix $\scrX$ with elements 
 $\scrX(l_1, l_2) = I\lkr |\hY(l_1, l_2)| > T \rkr$, $l_1, l_2 \in [L]$.\\ 
{\bf 2:} Cluster  $L$ rows of  the matrix $\SVD_M (\scrX) \in \calO_{L,M}$  into $M$ clusters using   
$(1+\eps)$-approximate $k$-means clustering. Obtain 
estimated clustering function $\hs$. \\   
%
\end{flushleft} 
\end{algorithm}
%


In this paper, we   use the relationship between the rows $W(l,:)$ of matrix $W$, $\linL$.
Let $\hW$ be an estimator of matrix $W$. 
Define matrices $Y, \hY \in \RR^{L \times L}$ with elements 
\be \label{eq:Y_hY}
Y(l_1, l_2) = \lan W(l_1,:),  W(l_2,:)\ran, \quad
\hY(l_1, l_2) = \lan \hW(l_1,:),  \hW(l_2,:)\ran. 
\ee
Then, Lemma~\ref{lem:matr_W_new}  implies that, for $l_1, l_2 \in [L]$, one has
\be \label{eq:Y_properties}
Y(l_1, l_2) = 0\ \mbox{if} \ \  s(l_1) \neq s(l_2); \quad \  
|Y(l_1, l_2)| \geq C \,  M\,L^{-1} \ \mbox{if} \ \  s(l_1)  = s(l_2).
\ee
Form a matrix $\scrX$ with elements $\scrX(l_1, l_2) = I(|\hY(l_1, l_2)| > T)$,
where  a threshold $T= T(n,L)$ is set up above the noise level.
Specifically, we choose $T=T(n,L)$ as 
\be \label{eq:T-value}
T(n,L) =  M\, L^{-1}\,  R(n,L),
\ee 
where $R(n,L)$ is defined in \fr{eq:RnL}.
If the estimator $\hW$ of $W$ is accurate enough,  matrix $\hY$ will have  properties similar to matrix $Y$, so that, 
$\scrX(l_1, l_2) =1$ if $s(l_1)  = s(l_2)$ and $\scrX(l_1, l_2) =0$ if $s(l_1) \neq s(l_2)$.
In this case, matrix $\scrX$ will coincide with the membership matrix with high probability. 
After that, to partition the $L$ layers into $M$ groups, one can  apply $k$-means clustering 
to the rows of  the matrix $\SVD_M(\scrX)$ of the $M$ leading singular vectors of the matrix $\scrX$.
The procedure  above is summarized as  Algorithm~\ref{alg:between}.

Unfortunately, an estimator of $W$ obtained by the 
straightforward SVD of the matrix $\calM_3 (\bA)$, has low accuracy. 
In order to improve   clustering  precision, we use  the Higher-Order Orthogonal Iteration Algorithm ({\bf HOOI}) 
to estimate matrices $U$ and $W$ in \fr{eq:bP_total}.
The HOOI algorithm is known to significantly improve   low-rank tensor recovery  (see, e.g., \cite{Xia_IEEE2018}).


\subsection{Estimation of  matrices  U  and  W  by HOOI algorithm}
\label{sec:HOOI}

Application of  Algorithm~\ref{alg:between} relies on securing an accurate  estimator $\hW$ of $W$.
While it follows from \fr{eq:bP_total}  that the matrices $U$ and $W$ can be estimated as the left singular vector matrices of a
proxy $\tilbA$ of the tensor $\tilbP$, the modern literature on tensor recovery (see, e.g., \cite{TWIST-AOS2079} and \cite{Xia_IEEE2018})   
implies that these estimators will be suboptimal if the signal strength, defined in \fr{eq:sig_strength} is not high enough,
and that their estimation precision improves if the HOOI  algorithm is applied. 
Specifically, in this paper, we use the regularized  HOOI algorithm suggested in \cite{ke2019_hypergraph} and \cite{TWIST-AOS2079}.
We present it here for the reader's convenience.

For any $U \in \RR^{n \times r}$, we define  a regularization operator  
\be \label{eq:reg_oper}
\Reg_{\del}(U) = \SVD_r(U_*) \quad \mbox{with} \quad 
U_*(i,:) = \frac{U(i,:)}{\|U(i,:)\|} \min(\del, \|U(i,:)\|), \quad i \in [n]. 
\ee
Denote
\be \label{eq:delu_delw}
\delu = (M\, K)^{1/2}\, n^{-1/2}\, \log n, \quad
\delw = (M\, K)^{1/2}\, L^{-1/2}\, \log n.
\ee 
If  initial estimators   $\hU^{(0)}$ and $\hW^{(0)}$ of $U$ and $W$ are available, then, 
$\hU$ and $\hW$ can be obtained by Algorithm~\ref{alg:HOOI}.

%
\begin{algorithm} [t] 
\caption{\ Regularized Orthogonal Power Iterations}
\label{alg:HOOI}
\begin{flushleft} 
{\bf Input:} Adjacency tensor $\bA \in \{0,1\}^{n \times n \times L}$; number of groups of layers $M$;\\  
average   ambient dimension $\barK$ of   groups of layers defined in \fr{eq:barU_K}; rank $r$ of $\calM_3(\bG))$; \\
regularization parameters $\delu$ and $\delw$;   initial estimators $\hU^{(0)}$ and $\hW^{(0)}$;
with $\|\hU^{(0)}\|\tinf \leq \sqrt{2}\, \delu$, $\|\hW^{(0)}\|\tinf \leq \sqrt{2}\, \delw$;   
maximum number of iterations $\Niter$; tolerance $\epstol$.
\\
{\bf Output:} Estimators $\hU$ and $\hW$  \\
{\bf Steps:}\\
{\bf 1:}\ Construct tensor $\tilbA$ with layers $\tilA\upl = \Pi^\bot A\upl\, \Pi^\bot$, $\linL$.\\
{\bf 2:}\ Set $t=0$, $\tilU^{(0)}= \hUo$, $\tilWo = \hWo$. \\  
%
{\bf 3:} {\bf While} $t < \Niter$ and $\eps > \epstol$\ \  {\bf do}:\\
\hspace*{6mm} {\bf 3a:} Set $t = t+1$\\
\hspace*{6mm} {\bf 3b:} Set $\tilU^{(t)}$ to be the top left $(M \barK)$ singular vectors of 
$\calM_1(\tilbA \tit [\hU^{(t-1)}]^T  \tir [\hW^{(t-1)}]^T)$.\\
\hspace*{6mm} {\bf 3c:} Set $\tilW^{(t)}$ to be the top left $r$ singular vectors of 
$\calM_3(\tilbA \tio [\hU^{(t-1)}]^T  \tit [\hU^{(t-1)})^T])$.\\
\hspace*{6mm} {\bf 3d:} Set $\hU^{(t)} = \Reg_{\delu} (\tilU^{(t)})$ and 
$\hW^{(t)} = \Reg_{\delw} (\tilW^{(t)})$.\\
\hspace*{6mm} {\bf 3e:} Set  $\eps = \|\tilU^{(t)} - \tilU^{(t-1)}\| + 
\|\tilW^{(t)} - \tilW^{(t-1)}\|$ \\
\hspace*{5mm} {\bf end While}\\
{\bf 4:}   Set $\hU = \hU^{(t)}$ and $\hW = \hW^{(t)}$
\end{flushleft} 
\end{algorithm}
%


In order to apply Algorithm~\ref{alg:HOOI}, one needs to construct $\hUo$ and $\hWo$, which 
are close enough to $U$ and $W$ and are such that $\|\hU^{(0)}\|\tinf \leq \sqrt{2}\, \delu$, 
$\|\hW^{(0)}\|\tinf \leq \sqrt{2}\, \delw$.
For this purpose, we first construct  $\tilUo$. It is easy to see that   
\be \label{eq:scrT}
U = \SVD_{(M\barK)} (\Pibot\,\scrT\, \Pibot) \quad \mbox{with} \quad
  \scrT = \sum_{\linL} \, \lkr  P\upl  \rkr^2. 
\ee
Consider a hollowing operator $\calH$ such that, for any square matrix $X$, one has $\calH(X) = X - \diag(X)$.
Construct an estimator $\hscrT$ of $\scrT$ of the form
\be \label{eq:hscrT}
\hscrT = \sum_{\linL} \,   \calH \lkr A\upl \rkr^2.
\ee 
Set $\tilUo = \SVD_{M\barK} (\Pibot\, \scrT \Pibot)$ and   $\hUo = \Reg_{\delu} (\tilUo)$. 
Subsequently, in the spirit of Algorithm~\ref{alg:HOOI}, obtain  $\tilWo$ as the top left $r$ singular vectors of 
$\calM_3(\tilbA \tio [\hUo]^T  \tit [\hUo)^T])$. This procedure is summarized as Algorithm~\ref{alg:initial}.


\begin{rem} \label{rem:unknown_K_M_r}  
{\bf  Estimation of the unknown parameter values for Algorithms~\ref{alg:between}  and \ref{alg:HOOI}}.
{\rm
Application of Algorithm~\ref{alg:between} requires the knowledge of 
parameters $M$, $M\, \barK$ and $r$. Since the latter two parameters coincide, respectively, with 
ranks of matrices  $\calM_1(\tilbA \tit [\hU^{(t-1)}]^T  \tir [\hW^{(t-1)}]^T)$ and 
$\calM_3(\tilbA \tio [\hU^{(t-1)}]^T  \tit [\hU^{(t-1)})^T])$,
they can be recovered by standard techniques such as, e.g., ScreeNOT of  \cite{donoho_screenot},
or the ``elbow''  method of \cite{Ghodsi_elbow_2006}. 

In order to estimate $M$ using the estimated matrix $\hW$, one can create a matrix $\hPsi$ with elements 
$\hPsi(l_1, l_2) = |\hY(l_1, l_2)|$,  and build a set of $J$ thresholds  $T_j$, $j \in [J]$, 
from quantiles of the off-diagonal entries of $\hPsi$,  for example 60\%, 65\%, ..., 95\%.  
For each of the thresholds $T_j$ one obtains a binary matrix $\scrX_j$  with elements 
 $\scrX_j(l_1, l_2) = I\lkr  \hPsi(l_1, l_2)  > T_j \rkr$. Compute eigenvalues $\lam_1^{(j)} \geq \lam_2^{(j)} \geq ... \geq  \lam_L^{(j)}$ 
of  $\scrX_j$, $j \in [J]$, and use them to construct an estimator $\hM_j$ of $M$  for each $J$ values of the threshold.
For example, one can use the value delivering the largest eigengap $\hM_j = \di \argmax_k (\lam_k^{(j)} - \lam_{k+1}^{(j)})$
or the index of the smallest eigenvalue that is larger than a fixed fraction $\alpha \in (0,1)$ of the largest eigenvalue.
This procedure gives a ``sample'' of $J$ possible values of $\hM_j$, and $\hM$ can evaluated as a sample mode or a sample median. 

While in the present paper we did not carry out a simulation study with unknown parameter values,
we explored the sensitivity of the technique to  the misspecification of the parameters $\barK$, $r$ and $M$.
The results of this investigation can be found in Section~\ref{sec:sensitivity} of the Supplementary Material. 
}
\end{rem}


 



\section{Theoretical analysis}
\label{sec:theory}


\subsection{Error measures and signal strength}
\label{sec:errors_sig_strength}

In this section, we study the  between-layer clustering error rates of the Algorithm~\ref{alg:between}. 
%
Since clustering is unique only up to a permutation of cluster labels, 
denote the set of $K$-dimensional permutation functions of $[K]$ by $\aleph(K)$, 
and the set of $(K \times K)$ permutation matrices by 
$\mathfrak{F} (K)$. 
The misclassification error rate of the between-layer clustering is then given by
\be \label{eq:err_betw_def}
R_{BL} = (2\,L)^{-1}\ \min_{\scrP \in \mathfrak{F}  (M)}\  \|\hS  - S \, \scrP\|^2_F,
\ee
where $\hS$  and $S$ are clustering matrices corresponding to clustering functions $\hs$ and $s$.
In order to assess the accuracy of estimating a matrix $U \in \calO_{n,K}$ by $\hU  \in \calO_{n,K}$, 
we use the $\sinTe$ distances.
Suppose that  the singular values of $U^T \hU$ are $\sig_1 \geq \sig_2 \geq ... \geq \sig_K>0$.  
Then, 
\be \label{eq:sinTheta}  
   \| \sin \Te(U, \hU) \| =  \lkr 1 - \sig_{K}^2 (U^T \hU) \rkr^{1/2},  \quad 
   \| \sin \Te(U, \hU)  \|_F = \lkr K - \|U^T \hU)\|^2_F\rkr^{1/2}. 
\ee 
We also define
\be \label{eq:sinT2inf}
D(U,\hU; 2,\infty) = \min_{W \in \calO_K} \|\hU \, W - U \|_{2,\infty}.
\ee
Observe (see, e.g., \cite{10.1214/17-AOS1541})  that
$D(U,\hU; 2, \infty) \leq \sqrt{2}\, \| \sin \Te(U, \hU) \|$ and also 
$\| \sin \Te(U, \hU) \| \leq \|\hU \hU^T - U U^T\| \leq \sqrt{2}\, \| \sin \Te(U, \hU) \|.$
%

%
\begin{algorithm} [t!] 
\caption{\ Initial estimation of $U$ and $W$}
\label{alg:initial}
\begin{flushleft} 
{\bf Input:} Adjacency tensor $\bA \in \{0,1\}^{n \times n \times L}$; number of groups of layers $M$;\\  
average   ambient dimension $\barK$ of   groups of layers defined in \fr{eq:barU_K}; rank $r$ of $\calM_3(\bG)$; \\
regularization parameters $\delu$ and $\delw$.
\\
{\bf Output:} Estimators $\hUo$ and $\hWo$  such that $\|\hU^{(0)}\|\tinf \leq \sqrt{2}\, \delu$, 
$\|\hW^{(0)}\|\tinf \leq \sqrt{2}\, \delw$
\\
{\bf Steps:}\\
{\bf 1:}\  Construct an estimator $\hscrT$ of $\scrT$ using formula \fr{eq:hscrT}.  \\
{\bf 2:}\ Set $\tilUo = \SVD_{M\barK} (\Pibot\, \scrT\, \Pibot)$ and  $\hUo = \Reg_{\delu} (\tilUo)$.   \\  
{\bf 3:}\ Set  $\tilWo = \SVD_r (\calM_3(\tilbA \tio [\hUo]^T  \tit [\hUo)^T])$ and 
$\hWo = \Reg_{\delw} (\tilWo)$.   \\ 
\end{flushleft} 
\end{algorithm}
%

Due to application of HOOI algorithm, the accuracy of estimation of $U$ and $W$ by $\hU$ and $\hW$   depends on the strength of the signal contained in the 
core tensor $\bTe$, which, in our case, is  defined as the minimum of the lowest singular values of 
$\calM_1 (\bTe)$ and $\calM_3 (\bTe)$:
\be \label{eq:sig_strength}
\sig_{\min}(\bTe) := \min \lkv \sig_{\min}(\calM_1 (\bTe)), \sig_{\min}(\calM_3 (\bTe))\rkv
\ee
Since DIMPLE-SGRDPG is a much more complex model, derivations of the signal strength are much more involved than in \cite{TWIST-AOS2079}.
It follows from \fr{eq:bTe} and \fr{eq:LamU} that 
$\sig_{\min}(\bTe) \geq C  \min \lfi \sig_{\min}(\calM_1 (\bH)), \sig_{\min}(\calM_3 (\bH))\rfi$,
where tensor $\bH$ is defined in \fr{eq:tens_pres}.
Due to \fr{eq:tens_pres}, one has 
$\calM_1 (\bG) = \calM_1 (\bH)(W^T \otimes I_{M \barK})$
and 
$\calM_3 (\bG) = W \calM_3 (\bH)$. Hence, 
\be \label{eq:sig_stren_G}
\sig_{\min}(\bTe) \geq C  \min \lfi \sig_{\min}(\calM_1 (\bG)), \sig_{\min}(\calM_3 (\bG))\rfi.
\ee
Then, the following statement is true.

\begin{lem} \label{lem:sig_strength}  
Let the probability  tensor $\bP$ be generated by the random procedure described in Section~\ref{sec:net_form}.
Let    Assumptions~{\bf A1} -- {\bf A6} hold.
Then, for any $\tau>0$ and some constants $c$ and $C$ that depend on $\tau$ and 
the constants in Assumptions~{\bf A1} -- {\bf A6} only, one has
\begin{align} 
& \PP \lfi \sig_{\min}^2(\calM_1 (\bTe))   \geq C \,\rhon^2\, n^2\, L\rfi \geq 1 - c\, n^{-\tau}, \label{eq:sig_M1}\\ 
& \PP \lfi \sig^2_{\min}(\calM_3 (\bTe)) \geq C\,  M^{-1} \rhon^2 n^2  L \rfi \geq 1 - c\, n^{-\tau}. \label{eq:sig_M3}
\end{align}
\end{lem}


\subsection{HOOI algorithm performance and accuracy of the between-layer clustering} 
\label{sec:HOOI_Perform_Accuracy}

In this section, we evaluate performance of Algorithm~\ref{alg:HOOI}. 
Note that we cannot simply repeat the calculus of \cite{TWIST-AOS2079} or \cite{Xia_IEEE2018}, since \cite{Xia_IEEE2018} assumes Gaussian errors while \cite{TWIST-AOS2079}  carries out derivations under a much simpler model and only for $L \leq n$. 

Recall that, due to \fr{eq:for_del1}  and \fr{eq:W2inf}, respectively, 
one has $\|U\|_\twin \leq C\, n^{-1/2}\, (K\, M)^{1/2}$ and 
$\|W\|_{2, \infty} \leq c\, L^{-1/2}\,  (K\, M)^{1/2}$. 
Hence,  if $n$ is large enough, $\|U\|\tinf \leq \delu$ and 
 $\|W\|\tinf \leq \delw$, so, for the true matrices $U$ and  $W$, one has 
$\Reg_{\delu} (U) = U$ and $\Reg_{\delw} (W) = W$.
Then, the  following statement holds.

\begin{thm}\label{thm:HOOI_accuracy} 
Let $\hU^{(t)}$ and  $\hW^{(t)}$ be step $t$ 
estimators,  obtained by Algorithm~\ref{alg:HOOI} with initial estimators $\hU\upo \in \calO_{n,(\barK M)}$ and 
$\hW\upo \calO_{L,r}$, and  regularization parameters $\delu$ and $\delw$ defined in \fr{eq:delu_delw}. 
Denote
\be \label{eq:delsnL}
\delsnl =  \Ctau\, K\, M^{3/2}\, (\log n)^3\,  \log\log n  \, \lkr \rhon\, n\, \min(n,L) \rkr^{-1/2},
\ee
\be \label{eq:epsnl}
\epsnl = \Ctau\,  M\, \sqrt{K} \, (\log n)^{3/2} \, \lkr \rhon\, n\, \min(n,L) \rkr^{-1/2}, 
\ee
\be \label{eq:errt}
\Err(t) = \max \lkr  \| \sin \Te(U, \hU^{(t)}) \|, \| \sin \Te(W, \hW^{(t)}) \| \rkr,
\ee
where $\|\sin\Te (U,V)||$   is defined in \fr{eq:sinTheta}.
If $\delsnl <1$ in \fr{eq:delsnL} and $\Err(0) < 1/4$, then  
\be  \label{eq:err_rel}
\Err(t) \leq \delsnl \Err(t-1) + \epsnl, \quad t =   1, 2,  ...
\ee
and $\hU\upt$ and $\hW\upt$ are such that 
$\|\hU^{(t)}\|\tinf \leq \sqrt{2}\,  \delu$, $\|\hW^{(t)}\|\tinf \leq \sqrt{2}\, \delw$.  
Moreover, if  $\delsnl <1/2$, then $\Err(t) \leq 3 \, \epsnl$ for $t \geq t_{\max} = -\log_2(2 \epsnl)$.
\end{thm}

 
\medskip

It is easy to check that 
$\epsnl \leq  C \, \delsnl\, (\log n)^{-3/2}\, (\log \log n)^{-1}\, (K\, M)^{-1/2}$ 
(since $\Ctau$ may take different values in \fr{eq:delsnL} and \fr{eq:epsnl}).
Therefore, if    $n$ is large enough, then   $\delsnl <1/2$ leads to small values of $\epsnl$.

Validity of  Theorem~\ref{thm:HOOI_accuracy}   relies on  construction of $\hUo$ and $\hWo$ such that 
$\|\hU^{(0)}\|\tinf \leq \sqrt{2}\,  \delu$, $\|\hW^{(0)}\|\tinf \leq \sqrt{2}\, \delw$ and  
$\Err(0) < 1/4$.  The following statement ensures that $\hUo$ and $\hWo$ satisfy the conditions above.


\begin{thm}\label{thm:initial}  
Let Assumptions {\bf A1}--{\bf A6} hold, and let $\hUo$ and  $\hWo$ be estimators of $U$ and $W$, 
obtained by Algorithm~\ref{alg:initial}.
Then, with probability at least $1 - c\, n^{-\tau}$, one has 
$\|\sinTe(U, \hUo)\|   \leq \Ctau\, \epsunl$ and $\|\sinTe(W, \hWo)\|  \leq \Ctau\, \epswnl$, where
\begin{align}
&   \epsunl = M\, \log n \, \lkv (\rhon\, n \sqrt{L})^{-1} + (\rhon\, n\, L)^{-1/2} + (n\, \log n)^{-1} \rkv,
\label{eq:epsunl}\\
&  \epswnl  = K\,   M^{3/2} \,  (\log n)^{5/2} \lkv \frac{\log \log n\, \sqrt{\log n}}{\sqrt{\rhon \, n\, \minnL}}\,
 \epsunl + \frac{1}{n\, \sqrt{\rhon}} \rkv.   \label{eq:epswnl}
\end{align}
\end{thm}

\begin{cor}  \label{cor:initial}  
Let conditions of Theorem~\ref{thm:initial} hold.  If $\delsnl <1/2$ in \fr{eq:delsnL}  and, in addition,
\be \label{eq:breveps_cond}
M\, \log n \, \lkr \rhon\, n\, \sqrt{L} \rkr^{-1} \to 0
\quad \mbox{as} \quad n,L \to \infty,
\ee
then, for $n$ large enough, one has 
$\|\hU^{(0)}\|\tinf \leq \sqrt{2}\,  \delu$, $\|\hW^{(0)}\|\tinf \leq \sqrt{2}\, \delw$ and  
$\Err(0) \leq 1/4$, where $\Err(t)$ is defined in \fr{eq:errt}.
Thus, for $\epsnl$ defined in \fr{eq:epsnl}, one has 
$\Err(t) \leq 3 \, \epsnl$ for $t \geq t_{\max} = -\log_2(2 \epsnl)$.
\end{cor}

Now, we can assess the accuracy of clustering in Algorithm~\ref{alg:between}. 
The latter is given by the following statement.

\begin{thm}\label{thm:clust_between}  
Let Assumptions~{\bf A1}--{\bf A6} and \fr{eq:breveps_cond} hold. Let  $\delsnl <1/2$ in \fr{eq:delsnL}
and $T(n,L)$ be defined in \fr{eq:T-value}.  
If assumptions of Lemma~\ref{lem:matr_W_new}  hold and  $R(n,L) \to 0$ as $n,L \to \infty$, 
where 
\be \label{eq:RnL}
R(n,L) = \frac{(KM)^{3/2}\, (\log n)^4 \, \log \log n}{\sqrt{\rhon\, n\, \minnL}}
+ \frac{K\, M\, (\log n)^{3/2}\, (\log \log n)^{1/2}}{\sqrt{n}}, 
\ee
then clustering by Algorithm~\ref{alg:between} is perfect with high probability, so that, 
up to a permutation of clusters' labels one has
$\PP(\hs = s) \geq 1 - c\, n^{-\tau}$.
\end{thm}

\medskip

Below, we elaborate on the accuracy of Algorithms~\ref{alg:HOOI}~and~\ref{alg:initial}.
We shall discuss the clustering precision of Algorithm~\ref{alg:between} later in Section~\ref{sec:discussion}.
Recall that Theorems~\ref{thm:HOOI_accuracy} and \ref{thm:initial}, respectively,  produce upper bounds for $\Err(t)$ 
when $t = t_{\max}$ and $t=0$. Since we do not know the respective values of the constants $\Ctau$, we postulate that 
employing the HOOI Algorithm~\ref{alg:HOOI} is advantageous if $\Err(t_{\max}) =o\lkr \Err(0)\rkr$.
Otherwise,  one should apply Algorithm~\ref{alg:between}  to $\hW = \hWo$, obtained by applying Algorithm~\ref{alg:initial}. 
In order to get an insight of when the application of Algorithm~\ref{alg:HOOI} improves  accuracy,
we assume that $K$ and $M$ are constants, that $\delsnl <1/2$ in \fr{eq:delsnL}, and that 
$R(n,L) \to 0$ as $n,L \to \infty$, where $R(n,L)$ is defined in \fr{eq:RnL}.
Since   $(n \rhon)^{-1} \log n \to 0$ guarantees perfect  clustering   in 
\cite{pensky2024signed}, here we focus on the case where $n \rhon \leq C \, \log n$. 
The summary is given by the following statement.
\\

\begin{cor}  \label{cor:comparison}  
Let conditions of Theorem~\ref{thm:HOOI_accuracy} and Corollary~\ref{cor:initial} hold and, in addition,
$K$ and $M$ be fixed constant independent of $n$ and $L$ and  $n \rhon \leq C \, \log n$.
Then, HOOI algorithm   improves the accuracy of the initial estimation unless 
\be \label{eq:unless}
C \, (\log n)^{-1/2} \leq n \rhon \leq C\, \log n  \quad \mbox{and} \quad  L\, (\log n)^2 = O(n),\quad  n,L \to \infty.
\ee
%
\end{cor}

Corollary~\ref{cor:comparison} confirms that, when $n \rhon \leq C\, \log n$, the HOOI algorithm improves estimation accuracy 
outside a very small subset of parameter values. However, we believe that this set of parameter values may be even smaller 
than that suggested by Corollary~\ref{cor:comparison}, since the logarithmic factors in
Theorems~\ref{thm:HOOI_accuracy}~and~\ref{thm:initial} are possibly not exact.


\begin{rem} \label{rem:other_errors}
{\bf   Other error distributions.}
{\rm
We should also comment on the fact that our paper studies the difficult case of 
the signed Bernoulli distribution, for which the error rates and the signal strength are intertwined.
However, the methodology developed in this paper can be extended to clustering any tensor whose groups 
of layers are embedded into distinct subspaces and whose loading matrices are all different. 
We believe that this methodology will work for sub-Gaussian and sub-exponential errors, 
which are not related to the underlying tensor, as long as the signal strength is sufficiently high.   
The  only parts of the paper that will change significantly 
are the error assessment in Section~8.1 of the Supplementary Material 
and the threshold $T$ in  Algorithm~\ref{alg:between}.  
The algorithms and the theoretical results can also be extended to the case where the layer matrices are not symmetric. 
In this case, it is necessary to add another iteration in Step~3 of Algorithm~\ref{alg:HOOI}.
}
\end{rem}






\section{Simulations study}
\label{sec:simul}
\setcounter{equation}{0}

%
\begin{algorithm} [t] 
\caption{\ The between-layer clustering (Pensky \cite{pensky2024signed})}
\label{alg:between_old}
\begin{flushleft} 
{\bf Input:} Adjacency tensor $\bA \in \{0,1,-1\}^{n \times n \times L}$; number of groups of layers $M$;  
ambient dimension $K^{(l)}$ of each layer $\linL$;  parameter $\eps$ \\
{\bf Output:} Estimated clustering function $\hs: [L] \to [M]$ for groups of layers\\
{\bf Steps:}\\
{\bf 1:} Find matrices $\hU_A\upl = \SVD_{K\upl} (\tilA\upl)$ 
of $K\upl$ left leading singular vectors  of matrices $\tilA\upl = \Pi^{\bot} A\upl  \Pi^{\bot}$, $\linL$ \\
{\bf 2:} Form  vectors $\hte\upl =  \vect(\lkr \hU_A\upl (\hU_A\upl)^T \rkr$, $\linL$ \\
{\bf 3:} Form matrix $\hTe$ with elements $\hTe(l_1, l_2) =  \lan \hte^{(l_1)},  \hte^{(l_2)} \ran$, $l_1, l_2 \in [L]$\\
%
%
{\bf 4:} Obtain matrix $\hV = \SVD_M (\hTe) \in \calO_{L,M}$  of $M$  leading singular vectors of matrix 
$\hTe$ \\
{\bf 5:}  Cluster  $L$ rows of  $\hV$ into $M$ clusters using   $(1+\eps)$-approximate $K$-means clustering. Obtain 
estimated clustering function $\hs$   
\end{flushleft} 
\end{algorithm}
%


In this section we carry out a simulation comparison of the new between-layer clustering Algorithm~\ref{alg:between}
with the between-layer clustering algorithm of \cite{pensky2024signed}, which, for convenience 
we present here as Algorithm~\ref{alg:between_old}. 
Since both papers study the same generative setting, this is a fair comparison. 
Later, we extend this comparison to a real-data setting in Section~\ref{sec:real_data}.

In order to obtain the DIMPLE-SGRDPG network, we follow the procedure described in  Section~\ref{sec:net_form}.
Specifically, we fix  the number of groups of layers, $M$,
and the ambient dimension of each group of layers, $K_m$. We generate the group membership of  each layer $\linL$, using 
\fr{eq:group_memb}. Subsequently, we  generate $M$ matrices $X\upm \in [-1,1]^{K_m}$ with i.i.d. rows using \fr{eq:Xm_gen}, 
where $f_m$ is the truncated normal distribution (Case~1) or the Dirichlet distribution  (Case~2) described in Section~\ref{sec:net_form}.
We generate the upper-triangular entries of the symmetric matrices $B\upl$, $\linL$, according to \fr{eq:DIMPnew_GDPG} 
as uniform random numbers  between $c$ and $d$, and form the
 connection probability matrices $P\upl$, $\linL$, using \fr{eq:DIMPnew_GDPG}.  
After that, the layer adjacency matrices $A\upl$ are  obtained  according to \fr{eq:SGDPG_adjacen}.


\begin{figure*} [t!] 
\centering
\[\includegraphics[width=8.0cm, height=3.8cm]{Graphs/Gauss_Bet_cd002_alln_L50red150black250blue350green.pdf}   
\includegraphics[width=8.0cm, height=3.8cm]{Graphs/Gauss_Bet_cd002_allL_n100red150black200blue250green.pdf}   \]%
\[\includegraphics[width=8.0cm, height=3.8cm]{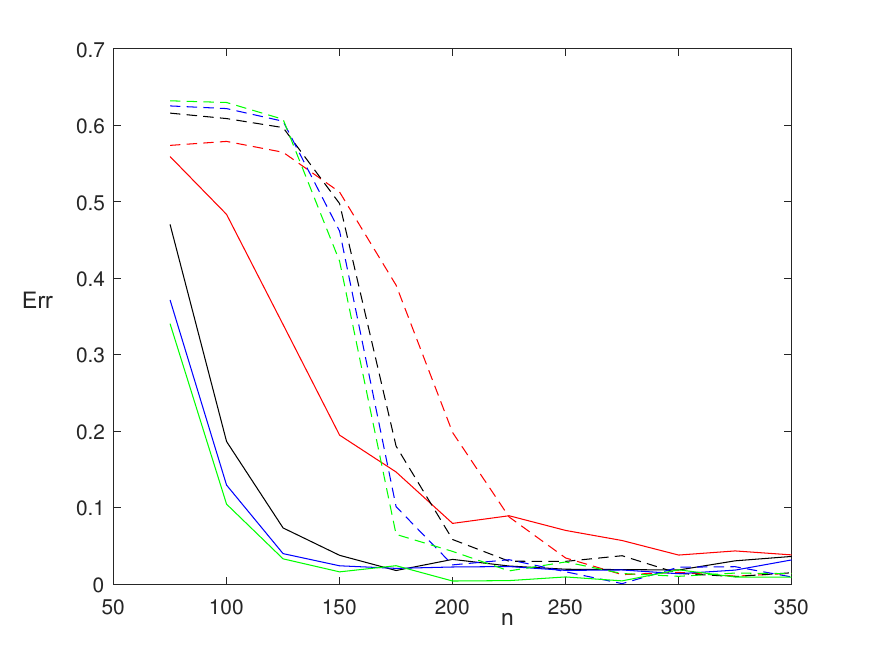}   
\includegraphics[width=8.0cm, height=3.8cm]{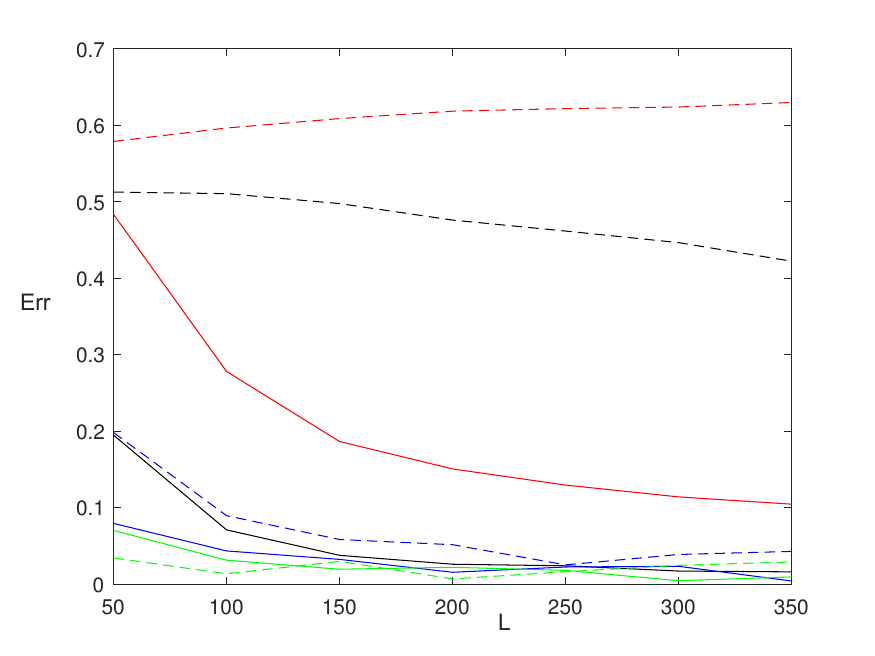}   \]%
\[\includegraphics[width=8.0cm, height=3.8cm]{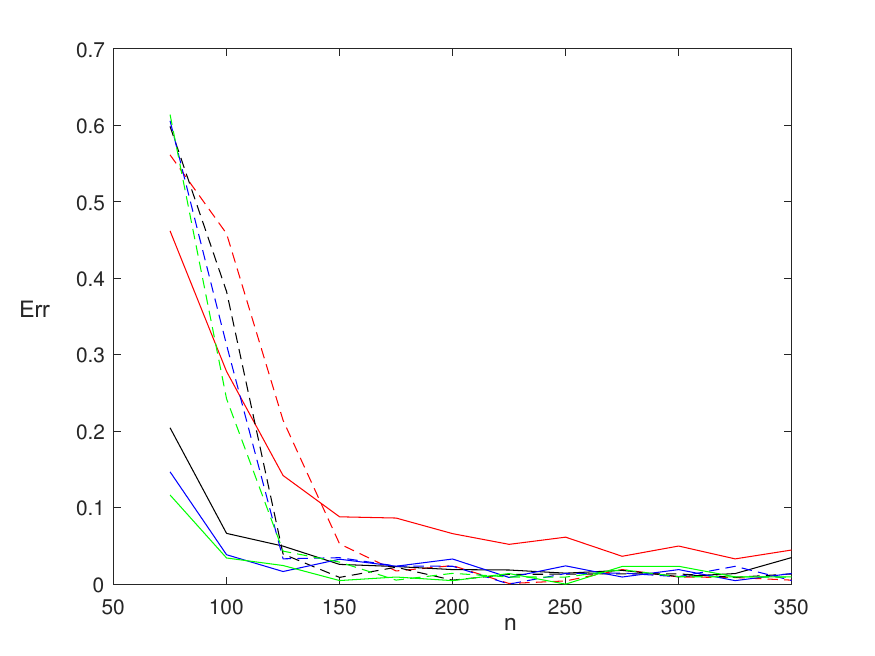}   
\includegraphics[width=8.0cm, height=3.8cm]{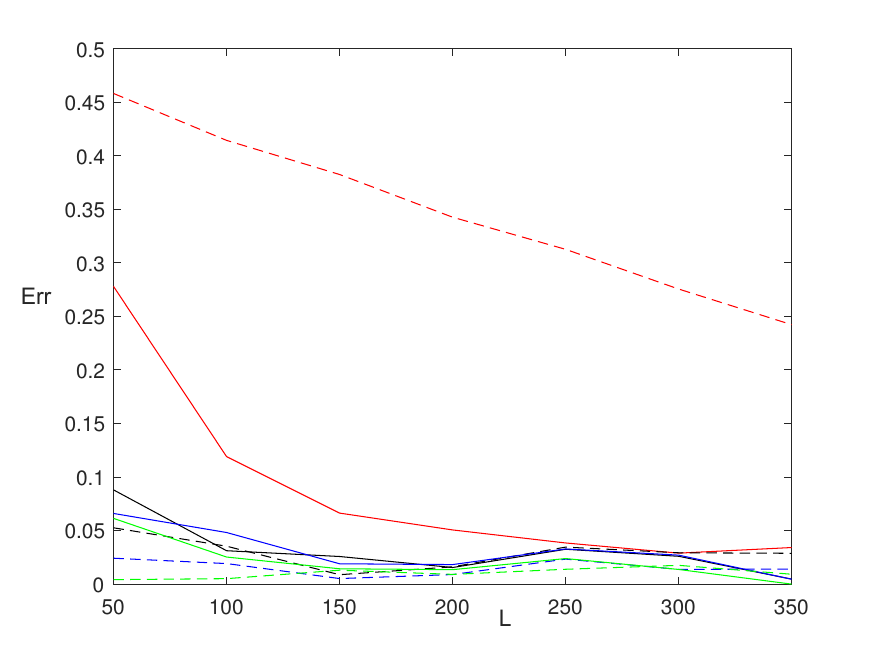}   \]%
\caption{\footnotesize{The between-layer clustering errors $R_{BL}$  in \fr{eq:err_betw_def} of Algorithms~\ref{alg:between}
 (solid lines) and  \ref{alg:between_old} (dash lines). 
Matrices $X\upm$  are generated by the truncated normal distribution  with $\sig =1$, $\mu =0$,   $M=3$ and $K_m=K=3$.
The entries of $B\upl$ are generated  as uniform random numbers  between   $c=-0.02$ and
$d = 0.02$ (top),  $c=-0.03$ and $d = 0.03$  (middle) and  $c=-0.05$ and $d = 0.05$  (bottom). 
%
Left panels: varying $n$;   $L=50$ (red),  $L=150$ (black),   $L=250$ (blue), $L=350$ (green).
Right panels: varying $L$;  $n=100$ (red),  $n=150$ (black),   $n=200$ (blue), $n=250$ (green).
Errors are averaged over 100 simulation runs.  
}}
\label{fig:normal}
\end{figure*}


\begin{figure*} [t!] 
\centering
\[\includegraphics[width=8.0cm, height=3.8cm]{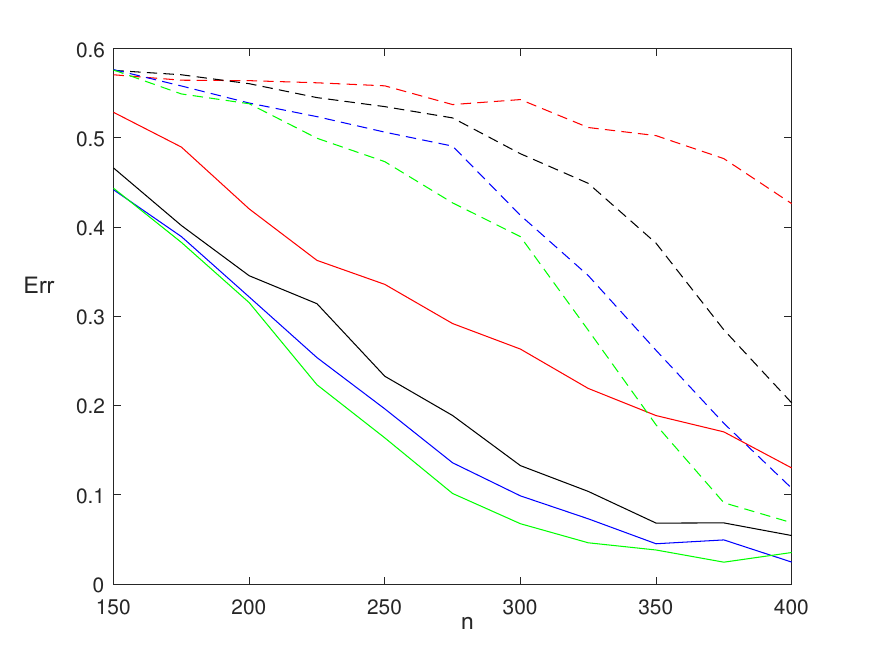}   
\includegraphics[width=8.0cm, height=3.8cm]{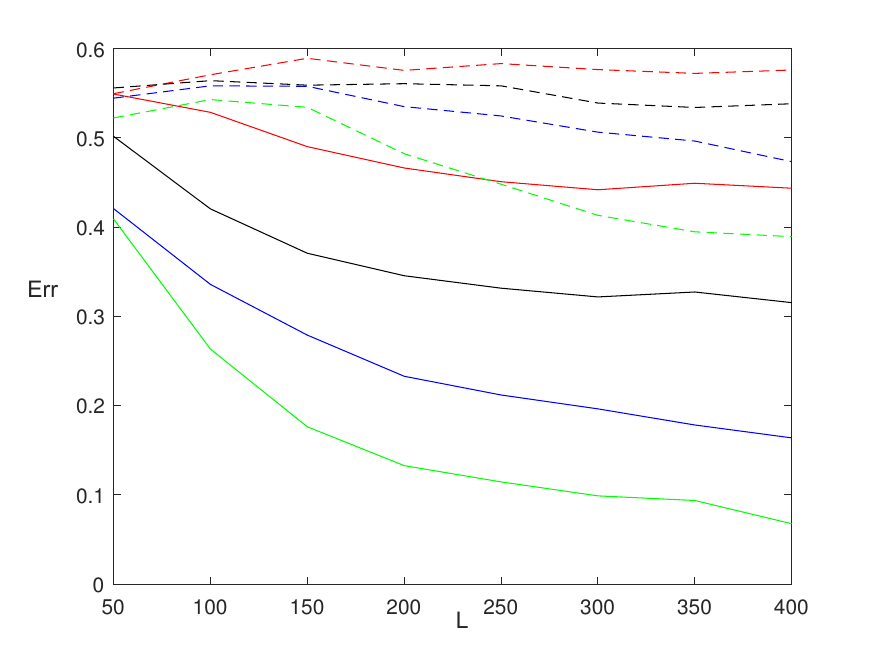}   \]%
\[\includegraphics[width=8.0cm, height=3.8cm]{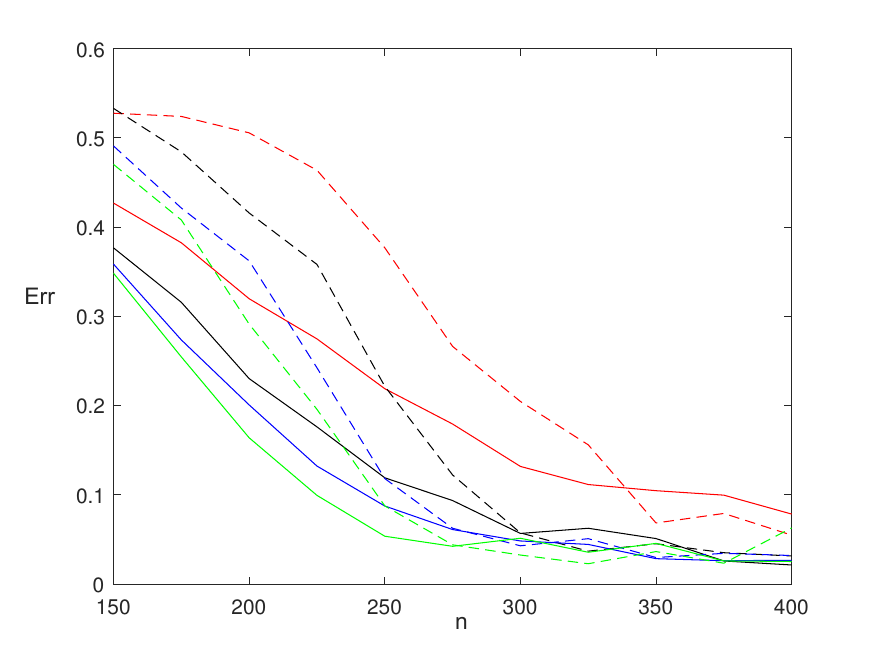}   
\includegraphics[width=8.0cm, height=3.8cm]{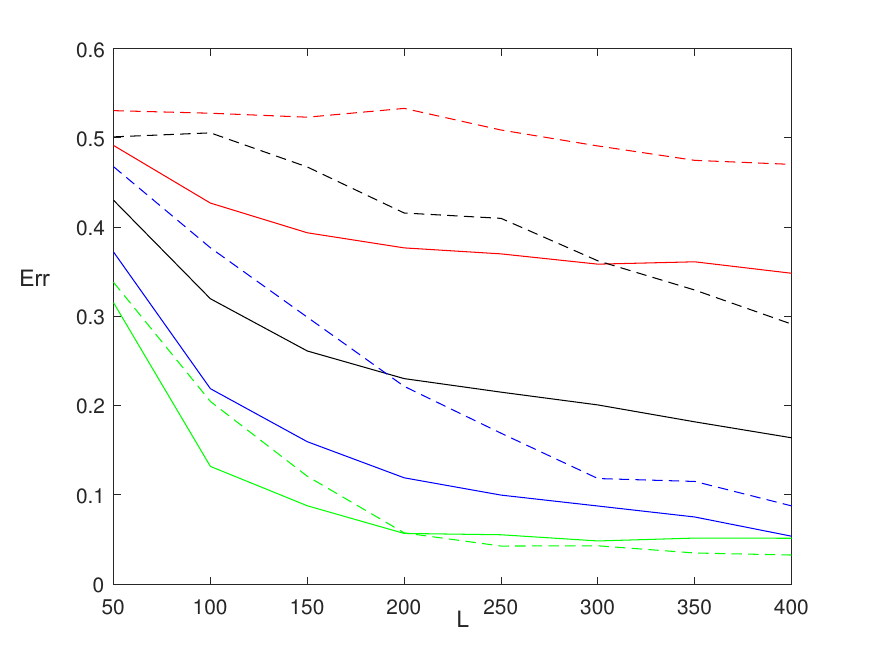}   \]%
\[\includegraphics[width=8.0cm, height=3.8cm]{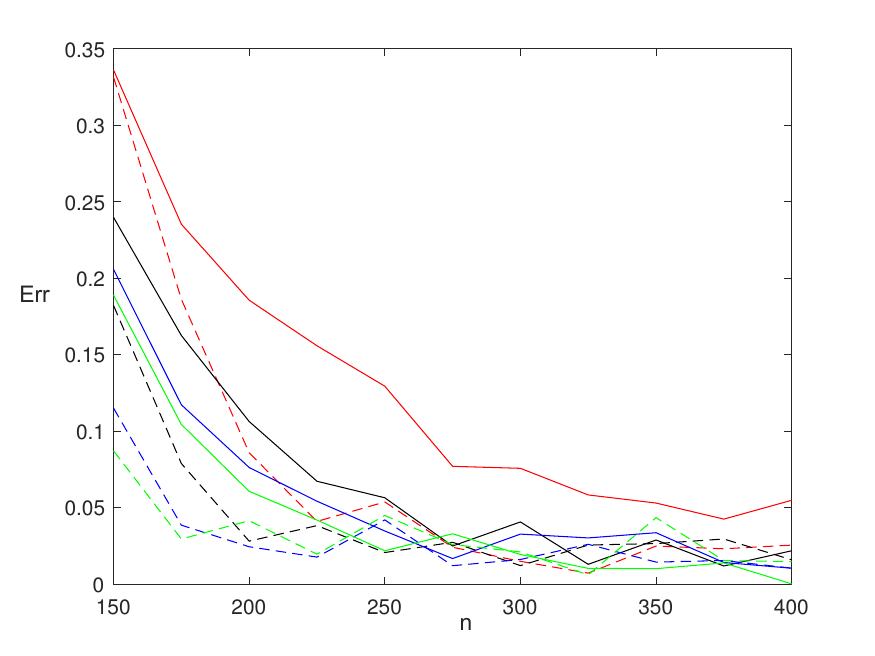}   
\includegraphics[width=8.0cm, height=3.8cm]{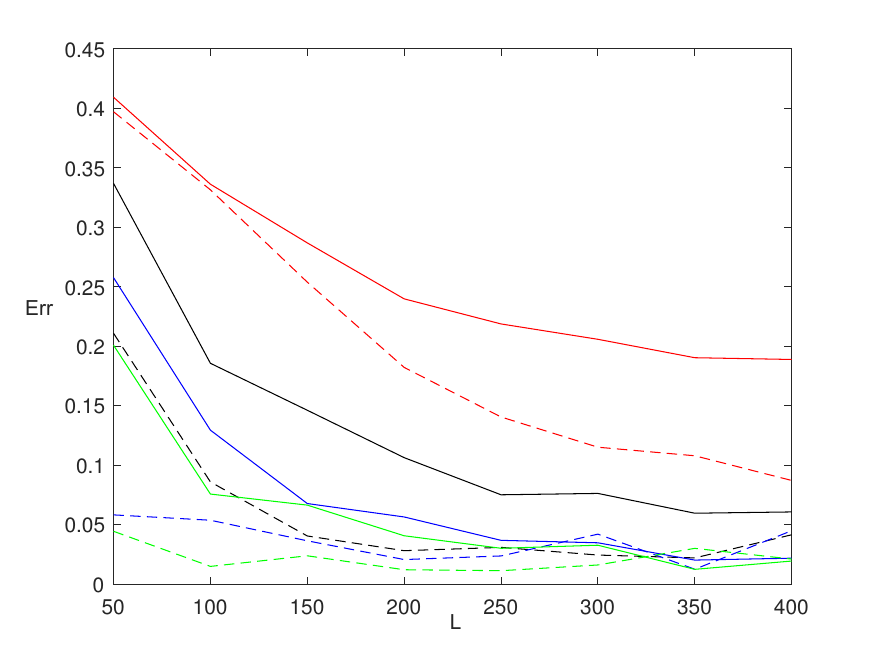}   \]%
\caption{\footnotesize{The between-layer clustering errors $R_{BL}$  in \fr{eq:err_betw_def} of Algorithms~\ref{alg:between}
 (solid lines) and  \ref{alg:between_old} (dash lines). 
Matrices $X\upm$  are generated by the Dirichlet distribution  with $\alpha=0.1$,    $M=3$ and $K_m=K=3$.
The entries of $B\upl$ are generated  as uniform random numbers  between   $c=-0.02$ and
$d = 0.02$ (top),  $c=-0.03$ and $d = 0.03$  (middle) and  $c=-0.05$ and $d = 0.05$  (bottom). 
%
Left panels: varying $n$;   $L=100$ (red),  $L=200$ (black),     $L=300$ (blue), $L=400$ (green).
Right panels: varying $L$;  $n=150$ (red),  $n=200$ (black),   $n=250$ (blue), $n=300$ (green).
Errors are averaged over 100 simulation runs.  
}}
\label{fig:Dirichlet}
\end{figure*}

Figures~\ref{fig:normal}~and~\ref{fig:Dirichlet} 
present  the between-layer clustering errors of Algorithm~\ref{alg:between}
and~Algorithm~\ref{alg:between_old} when the matrices $X\upm$  are generated using, 
respectively, the truncated normal distribution (Case~1)
and the Dirichlet distribution  (Case~2) with $M=3$, $K_m=K=3$. 
The  entries of $B\upl$ are generated  as uniform random numbers  in  $[-0.02,0.02]$ (top rows), 
$[-0.03,0.03]$ (middle rows) and  $[-0.05,0.05]$ (bottom rows).  
%
%
The left panels of Figures~\ref{fig:normal}~and~\ref{fig:Dirichlet} 
display the between-layer clustering errors when the number of nodes $n$  in the networks  grows for a fixed $L$,
while the right panels display the   between-layer clustering errors when the numbers of layers $L$  in the networks 
grows for a fixed $n$.

The top rows represent the cases of extremely sparse networks, where the entries of $B\upl$ are in $[-0.02, 0.02]$.
It is easy to see from the top left panels of  Figures~\ref{fig:normal}~and~\ref{fig:Dirichlet} that, while the errors 
of Algorithm~\ref{alg:between} decline steadily as $n$
grows, the errors of  Algorithm~\ref{alg:between_old}  remain almost constant when $n \rhon$ is 
small and start decreasing rapidly as soon as $n \rhon$ becomes large enough. 
The right panels of Figures~\ref{fig:normal}~and~\ref{fig:Dirichlet} demonstrate that 
Algorithm~\ref{alg:between} manages to take advantage of the additional data,
so that larger values of $L$ lead to better precision, but Algorithm~\ref{alg:between_old}
fails to efficiently utilize this information, so its precision remains very poor as $L$ grows.
In the case of  extremely  sparse networks,  Algorithm~\ref{alg:between}  always outperforms 
Algorithm~\ref{alg:between_old} of \cite{pensky2024signed}. 
Not only does the former allow strongly consistent clustering for very sparse networks, 
but it also takes advantage of the growing number of layers, which is not true for Algorithm~\ref{alg:between_old}.

The middle rows of Figures~\ref{fig:normal}~and~\ref{fig:Dirichlet} 
show the same results for  less sparse networks, where the entries of $B\upl$ belong to  $[-0.03, 0.03]$. 
Although the errors exhibit behavior similar to that of the extremely sparse case, one observes that as $n$ grows,
the errors of   Algorithm~\ref{alg:between} become somewhat larger than those of  
Algorithm~\ref{alg:between_old}, especially when $n$ is large and $L$ is small (e.g., $L=50$ and $n \geq 225$
in the case of the normal distribution and $n \geq 350$ in the case of the Dirichlet distribution).

 The bottom rows of Figures~\ref{fig:normal}~and~\ref{fig:Dirichlet} 
present the between-layer clustering errors for moderately sparse networks, 
where the entries of $B\upl$ lie in $[-0.05, 0.05]$. For such networks, as soon as the number of nodes
becomes reasonably large, Algorithm~\ref{alg:between_old} has precision similar 
to or better than that of  Algorithm~\ref{alg:between}. 
\\

\begin{figure*} [t!] 
\centering
\[\includegraphics[width=8.0cm, height=3.8cm]{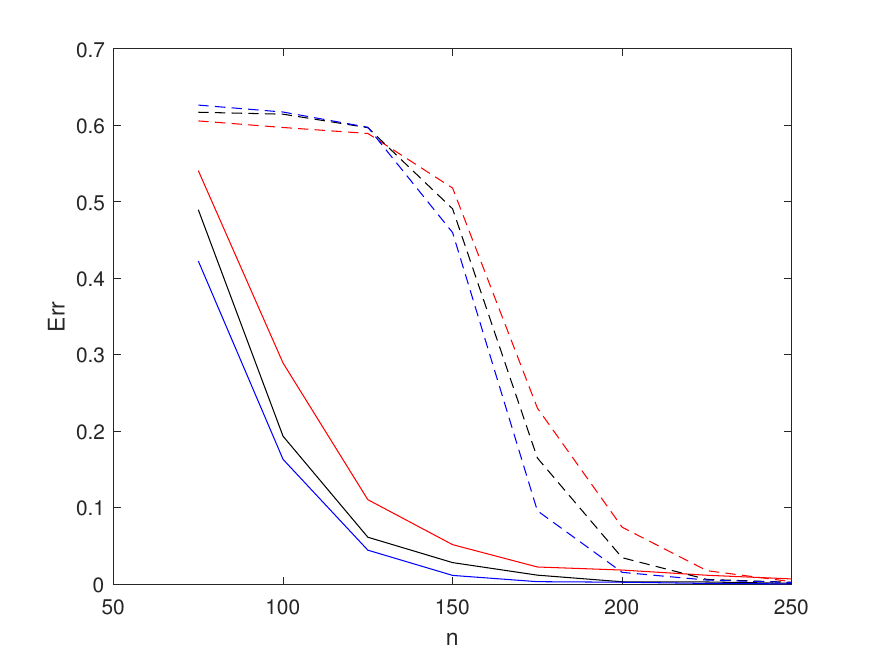}   
\includegraphics[width=8.0cm, height=3.8cm]{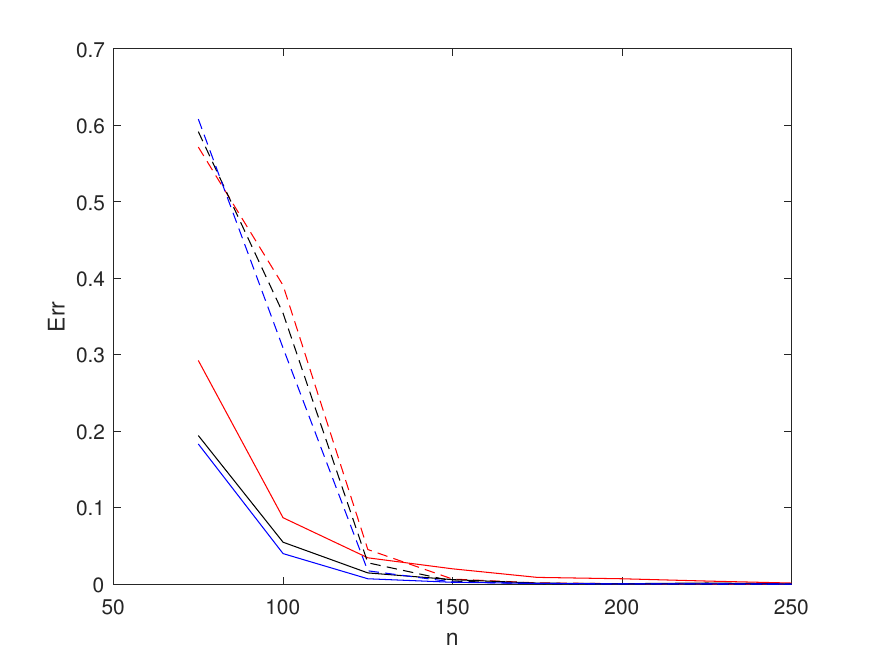}   \]%
\[\includegraphics[width=8.0cm, height=3.8cm]{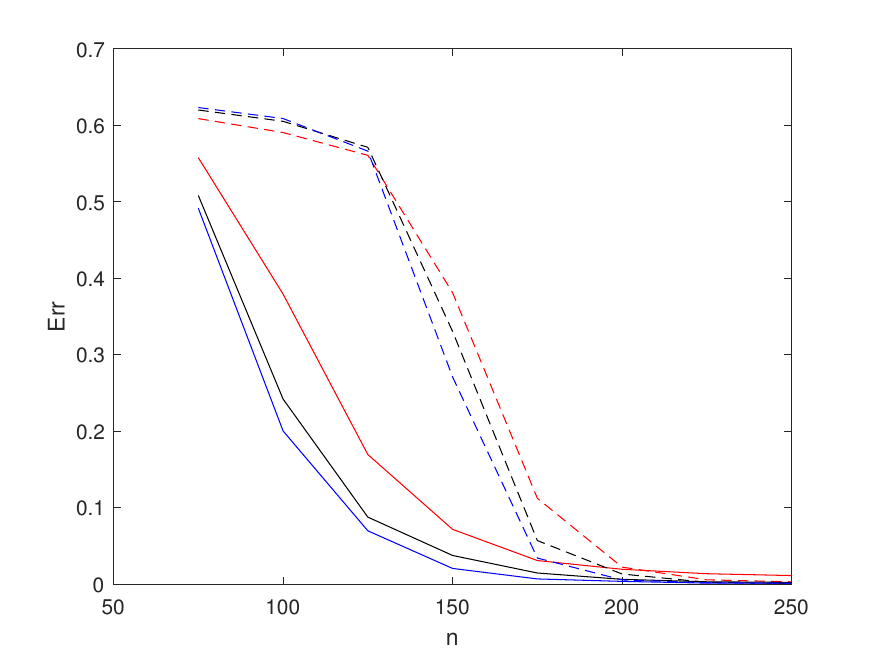}   
\includegraphics[width=8.0cm, height=3.8cm]{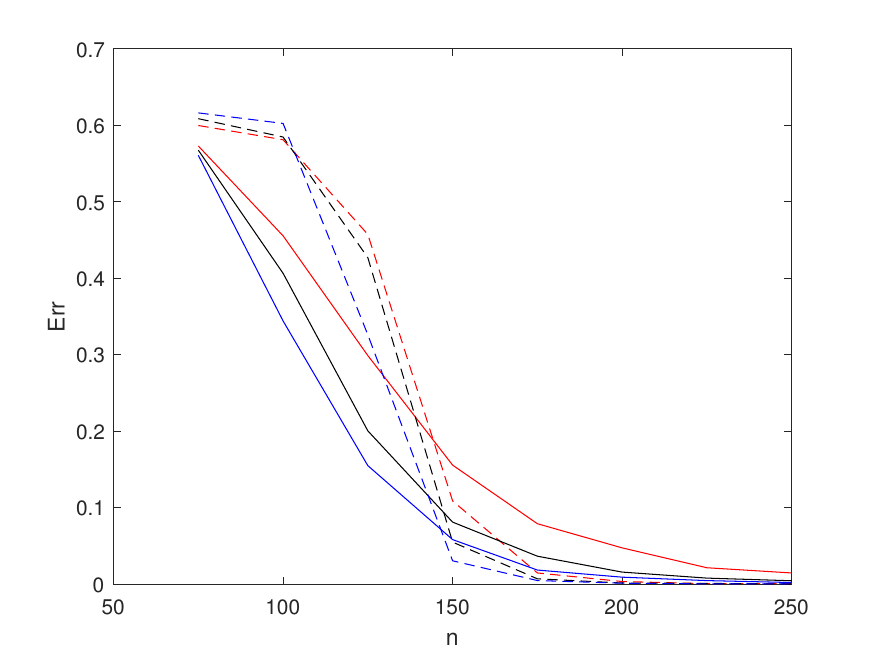}   \]%
\caption{\footnotesize{The between-layer clustering errors $R_{BL}$  in \fr{eq:err_betw_def} of Algorithms~\ref{alg:between}
 (solid lines) and  \ref{alg:between_old} (dash lines). 
Matrices $X\uplm$  are generated according to \eqref{eq:Xlm}  with $\sig =1$, $\mu =0$,   $M=3$ and $K_m=K=3$.
The entries of $B\upl$ are generated  as uniform random numbers  between   $c=-0.05$ and
$d = 0.05$ (top right), and between $c=-0.03$ and $d = 0.03$  (everywhere else).  
Top row: $\eps = 0.2$. Bottom row: $\eps = 0.5$ (left), $\eps = 0.9$ (right).
Varying $n$;   $L=50$ (red),  $L=150$ (black),   $L=250$ (blue), $L=350$ (green).
Errors are averaged over 50 simulation runs.  
}}
\label{fig:approx}
\end{figure*}

{\it The case where subspace structure is not perfectly shared within a group.}\ 
While our paper assumes that  the layers can be perfectly grouped such
that the subspace structure is fully shared within each group,
below we carry out a limited  simulation study of the situation where the subspaces within  the groups are 
close to each other but not identical. In particular, we replace $X\upm$ with
\be \label{eq:Xlm}
X\uplm = X\upm + \eps E\upl, \quad  m = s(l), \ l \in [L],\ m \in [M],
\ee
where rows of $X\upm$ and $E\upl$ are generated independently from the truncated 
normal distribution. Subsequently, we generate the matrices $B\upl$  as before and set 
$P\upl =  X\uplm  B\upl (X\uplm)^T$. Results of our limited simulation study 
are presented in Figure~\ref{fig:approx}. The top row of Figure~\ref{fig:approx} represents the effect
of sparsity while the bottom row shows the impact of   dissimilarity among the shared structures.

It is easy to see that for moderate values of $\eps$ ($\eps = 0.2$ and $\eps=0.5$), the tensor-based 
methodology delivers more accurate results for both sparsity settings, with 
a less sparse case corresponding to smaller errors. However, as $\eps$ grows,
the Tucker-based structure of the tensor becomes weaker and, when the signal is strong enough   ($n> 125$), 
the layer-by-layer method delivers more accurate results (bottom right panel). 
This assessment matches the evaluations reported in \cite{cong_ma_2025_AJIVE} and \cite{MinhTang_arxiv2022}.
However, we would like to emphasize that this is just a very brief analysis and that 
the issue requires a much more thorough investigation.


\section{Real Data Analysis}
\label{sec:real_data}
\setcounter{equation}{0}

In order to demonstrate the practical usefulness of our algorithm, we apply it to real data. We use the 
Worldwide Food Trading Networks data collected by the Food and Agriculture Organization of the United Nations.
The data contain trading volumes among more than 200 countries for more than 500 food items, and are available
from the Food and Agriculture Organization  at \url{https://www.fao.org/faostat/en/#data/TM}.   
A portion of these data for the year 2010  was  
described  in   \cite{doi:10.1038/ncomms7864} and subsequently 
 studied in some papers devoted to the analysis of multilayer networks 
(see, e.g., \cite{fan2021alma}, \cite{TWIST-AOS2079}, \cite{pensky2021clustering}). 
In all of those papers, the nodes are countries, the  layers are commodities, and the trade volumes 
are converted into binary outcomes. In addition, in order to avoid sparsity,   
all of the above papers examined only a small portion of the data,  
removing some of the nodes (countries) and choosing a  very limited subset of layers (food items): 
30 for  \cite{fan2021alma} and \cite{TWIST-AOS2079}, and 58 for \cite{pensky2021clustering}.

To distinguish ourselves from those efforts, we chose to investigate the trade network for the year 2018, since this was 
the year   before COVID-19  and   the war between Russia and Ukraine,  both of which caused a great deal of 
disruption to trading patterns. We decided to process the data keeping all countries that 
appear on both the importer and exporter lists and keeping practically all commodities traded by those countries.
Similarly to the previous publications, we constructed a multilayer network where countries are nodes and commodities are layers 
and an edge is drawn between two countries if they have an import or export relation. However, we did not place any 
threshold on the trade volumes, obtaining a multilayer network with {\bf 136 nodes and 503 layers}.  
The layer sparsity ranges from extremely sparse (0.0000541 for tangerine juice) 
to rather dense (0.298172578 for food preparations, n.e.c.), so 
the network does not really abide by our assumptions. After that, we applied    
Algorithms~\ref{alg:between}~and~\ref{alg:between_old} to the resulting multilayer network. In order to increase
precision, we used    10 replicates of the k-means clusterings  
for both algorithms and applied them with various values of  $M$, $K$  (for Algorithm~\ref{alg:between_old}), and $M\, \barK$ and $r$
(for Algorithm~\ref{alg:between}).

Since   no ``ground truth'' exists for the dataset, we evaluated the results of our analysis in terms of their stability. 
Due to  the k-means algorithm  delivering random outcomes, we investigated the stability of the clustering assignments 
produced by the two algorithms. In particular, we looked at the instability index 
for clustering assignments defined as 
\be \label{eq:instability}
\Ins  =  \lkr L (L-1)/2 \rkr^{-1}\  
\sum   4\,  J(\ell_1, \ell_2)\, \lkv 1 -   J(\ell_1, \ell_2)\rkv \, I(\ell_1 < \ell_2),
\ee 
where $J(\ell_1, \ell_2)$ is the empirical frequency with which layers $\ell_1$  and $\ell_2$ 
are assigned to the same cluster,
$\di J(\ell_1, \ell_2) = N^{-1} \, \sum_{j=1}^N I \lkr z^{(j)} (\ell_1) = z^{(j)} (\ell_2) \rkr$.
Here,  $z^{(j)} (\ell)$  is the $j$-th clustering label of layer $\ell$. If clusterings are consistent, 
then the layers $\ell_1$ and $\ell_2$ are 
either in the same cluster or in different clusters, which results in  
$4\, J(\ell_1, \ell_2)\, \lkv 1 -   J(\ell_1, \ell_2)\rkv  =0$. On the other hand, if the clusterings are extremely 
inconsistent for the layers $\ell_1$ and $\ell_2$, then  $J(\ell_1, \ell_2) \approx 1/2$ and 
$4\, J(\ell_1, \ell_2)\, \lkv 1 -   J(\ell_1, \ell_2)\rkv  \approx 1$. If the latter occurs for many pairs of layers, 
the instability can be close to 1. 
We ran the clustering algorithms with $M=4$    and a variety of values of $M\, \barK$ and $r$ 
for Algorithm~~\ref{alg:between}, and   of $K$ and $M$  for Algorithm~\ref{alg:between_old}.
Results are presented in, respectively, Tables~\ref{tab:instabH} and  \ref{tab:instabS} 
in Section~\ref{sec:addition_realdata} in the Supplementary Material.

Our analysis shows that clustering with the tensor-based Algorithm~\ref{alg:between} is  very consistent with 
 the average instability being close to zero (see Table~1 in the Supplementary Material). For the comparison,
we evaluated the instability index for Algorithm~\ref{alg:between_old}  with $M=4$ and $K=4$, and obtained values around 0.40. 
The latter means that, due to sparsity,  Algorithm~\ref{alg:between_old} essentially delivers 
random clustering results.

In addition, we chose  $M=4$, $K=4$, $M\, \barK=9$ and $r = 15$ for further investigation. For  this set
of parameters, Algorithms~\ref{alg:between}~and~\ref{alg:between_old}  yield, respectively,   $\Ins = 0$  and $\Ins = 0.3785$,    
but results for other parameter values show similar patterns.  Specifically, we examined   
the consistency of cluster sizes across clustering assignments.
In order to match the cluster labels across different runs, for each of the two algorithms, 
we created a reference clustering of the layers. We ran the algorithms $N=100$ times and, for each run, 
found the permutation that best matched the reference clustering.
Then, we calculated the mean and standard deviation of the cluster sizes.
The tensor-based Algorithm~\ref{alg:between} was perfectly consistent 
with clusters having the same sizes (41, 27, 141, 294) in all runs.  
On the other hand, since the sparser layers do not allow reliable spectral analysis, 
the layer-by-layer Algorithm~\ref{alg:between_old} was much less consistent, 
with the means and standard deviations of the cluster sizes equal to, respectively
(18.65,  8.50, 153.07, 322.78) and   (20.77, 6.98, 45.87, 49.63).

Finally,  for the clustering obtained using  Algorithm~\ref{alg:between},
 we created a list of commodities in each of the layers
as clustered by the algorithm  (see Table~\ref{tab:food_items} in the Supplementary Material). 
Those commodities can be broadly labeled as  {\it Fresh Produce} (41 items), 
 {\it Highly processed foods}  (27  items), {\it Animal products and grains} (141 items) and 
{\it Plant derivatives, juices, oils and crop products}  (294 items). 
Since some of the assignments by  Algorithm~\ref{alg:between} were not consistent with these categories, 
we created a list of the ``true'' categories for each layer and examined the differences between the ``true'' 
 clusters and those assigned by Algorithm~\ref{alg:between}, identifying 134 misclassified layers out of 503.
Here, we treated any commodity  with multiple possible labels as correctly classified.  
We also emphasize again that, since there is no ``ground truth'' associated with the data,
we cannot say whether the differences between  the two clusterings are meaningful. However, the 
fact that clustering with the tensor-based algorithm is much more stable is certainly important.
We did not produce a list of items using Algorithm~\ref{alg:between_old} since it yields very inconsistent clustering results.




\section{Discussion}
\label{sec:discussion}
\setcounter{equation}{0}

In this paper, we study clustering of layers in a  diverse multiplex network, where groups of layers are spanned by distinct subspaces 
and the loading matrices can all be different. Clustering of layers is a key  task in this setting  
since multiplex networks with a common subspace structure have been studied previously in a variety of publications. 
Until now (see, e.g., \cite{pensky2024daviskahan}, \cite{pensky2024signed}), clustering has been based on layer-by-layer analysis, 
and therefore consistent clustering  required that $\rhon n = \Om(\log n)$   for large $n$.
This was due to the observation that the probability tensor of such a network falls into the category of 
partially multilinear low-rank tensors   \cite{JMLR:AZhang21}.

Nevertheless, in this paper we succeeded in pooling information across all layers together and establishing  
perfect clustering under much weaker assumptions requiring that $\rhon\, n\,  \minnL$ and $\rhon \, n \sqrt{L}$ 
grow faster than some powers of $\log n$ (see formula \fr{eq:RnL}). 
As  \cite{Lei_Zhang_AOS2024_Comp_Lower_Bounds} shows,   
up to  logarithmic factors, those constraints coincide with the computational lower bounds for much simpler model than ours.  
Therefore, our main contribution is achieving strongly consistent clustering of layers,
even in the case of extreme sparsity of the layer networks, in the very general DIMPLE-SGRDPG multilayer 
network model (which includes the majority of multilayer network models as special cases).
In spite of the fact that the model is much more complex than the MMLSBM studied in  
\cite{fan2021alma}  and  \cite{TWIST-AOS2079}, we recover, up to logarithmic factors that are difficult to compare, 
the results of those papers.

The highly accurate clustering achieved in our paper is based on the novel use of the orthonormal 
basis of the tensor in the mode associated with layer clustering, and the application of the HOOI algorithm. 
We emphasize that both the clustering technique and the analysis of the HOOI algorithm are significantly 
different from those in  \cite{TWIST-AOS2079}.

We also compared our new between-layer clustering Algorithm~\ref{alg:between}  
with Algorithm~\ref{alg:between_old} developed in  \cite{pensky2024signed}.
Our theoretical analysis and simulation studies lead to several key take-away messages 
When a network is extremely sparse, so that $n \rhon = o(\log n)$, there is no alternative to 
HOOI-based Algorithm~\ref{alg:between}. Indeed, Algorithm~\ref{alg:between_old} does not deliver 
consistent layer clustering in this case. However, as $n \rhon$   grows,
the  performance of Algorithm~\ref{alg:between_old} becomes increasingly accurate.
Indeed, if 
$n \, \rhon = \Om (\barK \, M^2\, \log n) \quad \mbox{as} \quad n \to \infty$,
Algorithm~\ref{alg:between_old}  in \cite{pensky2024signed}
also achieves perfect clustering with high probability.

For such relatively dense networks, Algorithm~\ref{alg:between_old} may be preferable since it has several other advantages.
It has no tuning parameters while the HOOI requires specification of the threshold $T$, the
regularization parameters $\delu$ and $\delw$ and a stopping rule. Algorithm~\ref{alg:between_old} is not iterative and runs faster.
The most time-consuming part of Algorithm~\ref{alg:between_old}, , namely, finding the leading singular vectors 
of the layers' adjacency matrices, can be easily parallelized. It is specifically designed for the 
clustering of layers of a diverse multiplex network, so that, unlike   Algorithm~\ref{alg:between},
it clusters the rows of the matrix $\hW \in \calO_{L,M}$, which is directly related to the estimated 
clustering matrix $\hS$ of layers: $\hW = \hS (\hS^T \hS)^{-1}$. On the contrary, 
 Algorithm~\ref{alg:between}  applies clustering to the matrix $\hW \in \calO_{L,r}$, which is 
only indirectly associated with the clustering matrix $\hS$. 
%




\begin{funding}
The author of the paper gratefully acknowledges partial support by National Science Foundation (NSF) grant DMS-2310881. 
This material is partially based upon work supported by the NSF grant DMS-1928930, while the author was in residence at the Simons
Laufer Mathematical Sciences Institute in Berkeley, California, during the Spring 2025 semester.
\end{funding}


\vspace{-10pt}
 \begin{supplement}
\stitle{ }
\vspace{-10pt}
\sdescription{{\bf Supplementary Material} contains additional details of the simulations and a real data example,  and proofs of 
all statements in the paper.}
\end{supplement}


\bibliographystyle{imsart-nameyear} 
\bibliography{MultilayerNew_no_doi.bib}       


\newpage

\newpage



\section{Supplementary Material} 
\label{sec:proofs}
\setcounter{equation}{0} 

\renewcommand{\theequation}
{A\arabic{equation}}


\subsection{Assessment of incoherent  norms of the error tensor}
\label{sec:incoherent_norm}

This section contains a concentration inequality for tensor and matrix  incoherent norms.
Following \cite{TWIST-AOS2079}, we introduce tensor incoherent norm as follows:
for $\bX \in \RR^{n_1 \times n_2 \times n_3}$, $\del \in (0,1]$ and $k=1,2,3$, define
\be \label{eq:incoh_norm}
\|\bX\|\lkd = \sup \lfi \bU \in \calU_k(\del_k):\ \lan \bX, \bU\ran  \rfi,
\ee
where 
\be \label{eq:calukd} 
 \calU_k(\delk) = \lfi u_1 \otimes u_2 \otimes u_3:\ \|u_j\|\leq 1,\, j=1,2,3; \  \|u_k\|_{\infty} \leq \del_k \rfi.  
\ee 
Assume that 
\be \label{eq:rho_bound_nL}
\rho\, n \,  (n \wedge L)  \geq C_\rho\,  \log   \maxnL  .
\ee

\begin{prop}  \label{prop:incoherent_error}
Let $\bA  \in \RR^{n  \times n  \times L}$  have a signed Bernoulli distribution defined in \fr{eq:SGDPG_adjacen}
with $\EE \bA = \bP$.  Denote $\bXi = \bA - \bP$, $n_1, n_2 = n$, $n_3 = L$, $\delo = \delt=\delu$, 
$\delr = \delw$.  If $|\bP(i_1, i_2, i_3)| \leq \rho$ for any $i_1, i_2, i_3$,  
and inequalities  \fr{eq:nLtau} and \fr{eq:rho_bound_nL}  hold, then,
for any $\tau>0$, there exist  an absolute constant $c$ and a 
constant  $C_{\tau}$  that depends  only on $\tau$, such that 
\be  \label{eq:incoherent_error}
\PP \lfi \|\bXi\|\lkd \geq \Del_k \rfi \leq c n^{-\tau}, \quad \Del_k = \Del_k(n,L,\delk),
\ \  k = 1,2,3, 
\ee
where
\be \label{eq:Delk}
\Del_k \leq  
\Ctau\,    \sqrt{\rho\, \maxnL}\ (\log n)^2 \, \delk  \sqrt{\nk}\, \log(\delk^2\, \nk).
\ee
Moreover, if condition \fr{eq:delu_delw} holds, then, due to   $\log L = O(\log n)$,  
derive
\be \label{eq:Delk_simple}
\Del_k \leq  
\Ctau\,  \sqrt{K\, M}\,  \sqrt{\rho\, \maxnL}\ (\log n)^3 \,  \log \log n.
\ee
\end{prop}


\medskip

\noindent
{\bf  Proof  of Proposition~\ref{prop:incoherent_error}. }
Validity of the Proposition follows directly from the Lemma 13 of 
 \cite{lyu2023optimalclusteringdiscretemixtures}, which provides an extension of Theorem~4.4 in
\cite{TWIST-AOS2079} to the case where $L$ is possibly larger than $n$. 
\\


\medskip


\begin{cor}  \label{cor:incoherent_error}
Let Assumptions of Proposition~\ref{prop:incoherent_error}  hold and let $\tilbXi = \tilbA - \tilbP$,
where $\tilbP$ and $\tilbA$ are defined in \fr{eq:tilbPA}. Then, 
\be  \label{eq:incoherent_error_new}
\PP \lfi \|\tilbXi\|\lkd \geq \Del_k \rfi \leq c n^{-\tau}, \quad \Del_k = \Del_k(n,L, 2\,\delk),
\ \  k = 1,2,3, 
\ee
where the upper bounds for $\Del_k$ are given in \fr{eq:Delk} and \fr{eq:Delk_simple}.  
\end{cor}


\medskip

\noindent
{\bf  Proof  of Corollary~\ref{cor:incoherent_error}. }
Validity of Corollary~\ref{cor:incoherent_error}  follows from the fact that, for any vector $u$,  
$\|u\| \leq 1$ implies $\|\Pi^{\bot}\, u\| \leq 1$ and also
$\|u\|_{\infty} \leq \del$ implies $\|\Pi^{\bot}\, u\|_{\infty} \leq 2\, \del$.
Therefore, 
\bes
\|\tilbXi\|_{1, \delu} \leq \|\bXi\|_{1, 2 \delu}, \quad 
\|\tilbXi\|_{3, \delw} \leq \|\bXi\|_{3, 2 \delw},
\ees 
which completes the proof.

\medskip


\begin{prop}  \label{prop:matrix_error}  
Let entries of matrix $A  \in \RR^{n  \times n}$  have a signed Bernoulli distribution defined in \fr{eq:SGDPG_adjacen}
with $\EE  A =  P$, $\|P\|_\infty \leq \rho$,  $\Xi =  A -  P$. Let inequalities  \fr{eq:nLtau} and \fr{eq:rho_bound_nL}  hold.
Consider a set 
\bes
 \calV (\del) = \lfi v \in \RR^n:\ \|v\|\leq 1, \, \|v\|_{\infty} \leq \del \rfi.
\ees
Then, for any $\tau >0$, there exist an absolute constant 
 $C_{\tau}$  that depends  only on $\tau$, such that 
\be  \label{eq:scalar_product_error}
\PP \lfi \sup_{u,v \in \calV (\del)}\, \langle  \Xi, u v^T   \rangle  \geq \Del (n, \del) \rfi \leq c n^{-\tau},
\ee
where
\be \label{eq:Delndel}
\Del (n,\del) \leq  
\Ctau\, \lkv  \sqrt{\rho \, (n \rho + \log n)\log n \log \log (n \del^2)}
+ \del^2  \, (n \rho + \log n)\log n \log \log (n \del^2)\rkv.
\ee
\end{prop}


\medskip

\noindent
{\bf  Proof  of Proposition~\ref{prop:matrix_error}. }
Let $\Up \in \{-1, 1\}^{n \times n}$ be a matrix of Rademacher 
random variables with 
$\PP\lfi \Up(i,j) = 1 \rfi = \PP\lfi \Up(i,j)= -1 \rfi = 1/2$.
Denote the Hadamard product of matrices  $\Up$ and $A$ by $Y = \Up \circ A$.
Then, by standard symmetrization argument, for any $x >0$, one has
\be \label{eq:for1}
\calP_{n,\del} := \PP\lfi   \langle  \Xi, u v^T   \rangle \geq   x \rfi  
\leq \calP\upone   + 4 \calP\uptwo
\ee
where 
\be \label{eq:for2}
\calP\upone  = \PP \lfi \lan \Xi, u v^T  \ran \geq x/3 \rfi, \quad
\calP\uptwo =  \PP \lfi \sup_{u,v \in \calV (\del)}\, \lan Y, u v^T  \ran  \geq x/3 \rfi.
\ee
Below, we obtain upper bounds for $\calP\upone$ and    $\calP\uptwo$  separately.
\\


\underline{Upper bound for $\calP\upone$.}  Consider sets 
\bes
\calI^2 = \lfi (\io,\itw): 1 \leq \io \leq \itw \leq n \rfi, \quad \calI_{\io,\itw} = \{ (\io,\itw), (\itw, \io)\}.
\ees
Define 
\be  \label{eq:sumi1i2k}
\xi =   \lan \Xi,  u v^T \ran =   \sum_{(\io,\itw) \in \calI^2} \Xi(\io,\itw)\, 
\sum_{\icalI} u(\ipo) v (\ipt)
\ee  
and apply Bernstein inequality to $\xi$. For this purpose, observe that 
\begin{align*} 
 \left| \Xi(\io,\itw)\, \sum_{\icalI} u(\ipo) v (\ipt) \right| \leq 2\, \del^2, \quad 
  \Var (\xi) \leq  2\, \rho \,   \sum_{(\io,\itw) \in \calI^2} u^2(\ipo) v^2(\ipt) m \leq 2 \, \rho.
\end{align*} 
Then, Bernstein inequality yields that, for any $x>0$,
\bes 
\calP\upone = \max_{u,v \in \calV (\del)}\, \PP \lfi \xi  \geq x/3 \rfi \leq \exp \lkr - \frac{x^2}{72\, \rho} \rkr 
+ \exp \lkr - \frac{x}{8\, \del^2} \rkr. 
\ees 
Hence,  for some constant $\Ctau$ that depends only on $\tau$, one has 
\be \label{eq:calP1}
\calP\upone \leq n^{-\tau} \quad \mbox{if} \quad x \geq  \Del\upone  =  \Ctau \lkr \sqrt{\rho\, \log n} + \del^2  \, \log n \rkr
\ee

\underline{Upper bound for $\calP\uptwo$.} \   Define a discretization $\calvs (\del)$ of $\calV (\del)$:
\bes
\calvs(\del) = \lfi  \pm\, 2^{l} \cdot (2 n)^{-1/2},\ l = 0,...,\mdel \rfi^n, \quad
\mdel = \log_2 (n\, \del^2) + 1.
\ees
Consider  $\calG_1 = \lfi i:\, Y(i,:) \neq 0\rfi$ and 
$\calG_2 = \lfi i :\, Y(:,i) \neq 0\rfi$. Then, due to the symmetry, 
\bes
 \PP(|\calG_1\|> \nus)= \PP(|\calG_2\|> \nus) < n^{-\tau},
\ees
 where
\be \label{eq:nus} 
\nus = \Ctau\, (\rho\, n + \log n).
\ee
Consider a set $\calF = \lfi \om:\ \lkr \|\calG_1| \leq \nus \rkr \bigcap  \lkr \|\calG_1| \leq \nus \rkr \rfi$
with $\PP(\calF) \geq 1 - 2\, n^{- \tau}$.

Denote projection on support of a set $B$ by $\Pi_B$ and   
\bes
\calQ_j (\del) = \lfi \tilu: \,  \tilu = \Pi_{B_j} u,\, u \in \calvs (\del),\,
\|u\| \leq 1,  \, \|B_j\| \leq \nus \rfi, \quad j=1,2,
\ees
\bes
\calQ  (\del) = \lfi (\tilu, \tilv):\,  \tilu = \Pi_{B_1} u, \, \tilv = \Pi_{B_2} v, \, u, v \in \calvs (\del),\,
\|u\| \leq 1, \|v\| \leq 1, \, \|B_1\| \leq \nus, \|B_2\| \leq \nus \rfi.
\ees
Then, $\calQ (\del) = \calQ_1 (\del) \times \calQ_2 (\del)$  
with $\log \card(\calQ_j  (\del)) \leq 2\, \nus\, \log (\mdel)$.
%
Therefore, with high probability, one has 
\be  \label{eq:logcard}
\log \card(\calQ   (\del)) \leq \tilnu_n:= \Ctau\, (\rho\, n + \log n) \, \log \log (n \del^2).
\ee
Now, we derive that 
\be \label{eq:for3}
\calP\uptwo \leq 2\, n^{-\tau} + \calP\uptwo_0,\quad  
\calP\uptwo_0 = \PP \lfi \sup_{(\tilu, \tilv) \in \calQ  (\del)}\, \lan Y, \tilu \tilv^T  \ran  \geq x/3 \rfi.
\ee 
Apply Bernstein inequality to $\lan Y, u v^T  \ran$ in a similar manner to how it was done 
in the case of $\calP\upone$. Derive 
\bes
\PP \lfi \lan Y, \tilu \tilv^T  \ran  \geq x/3 \rfi \leq \exp \lkr - \frac{x^2}{72\, \rho} \rkr 
+ \exp \lkr - \frac{x}{8\, \del^2} \rkr. 
\ees 
Now, apply the union bound over $\calQ (\del)$. Obtain that 
\bes
\calP\uptwo_0
 \leq  \exp \lkr \tilnu_n - \frac{x^2}{72\, \rho} \rkr 
+ \exp \lkr \tilnu_n -  \frac{x}{8\, \del^2} \rkr. 
\ees
Consequently, 
\be \label{eq:calP2}
\calP\uptwo \leq c\, n^{-\tau} \quad \mbox{if} \quad x \geq  
\Del\uptwo  =  \Ctau \lkr \sqrt{\tilnu_n\, \rho\, \log n} + \tilnu_n\, \del^2  \, \log n \rkr.
\ee
Now, \fr{eq:for1}, \fr{eq:for2}, \fr{eq:calP1} and \fr{eq:calP2} imply that 
$\calP_{n,\del} \leq c\, n^{-\tau}$ provided
\be \label{eq:for5}
x \geq \Ctau\, \lkr \sqrt{\rho\, \log n} + \del^2  \, \log n + \sqrt{\tilnu_n\, \rho\, \log n} + \tilnu_n\, \del^2  \, \log n \rkr.
\ee 
Since according to \fr{eq:logcard}, $\tilnu_n \geq C\, \log n$, the last two terms in \fr{eq:for5} dominate, so that 
$\Del (n, \del)$ in \fr{eq:scalar_product_error} satisfies \fr{eq:Delndel}.


\subsection{Proof of statements in Section~\ref{sec:HOOI_Perform_Accuracy}}
This section contains results related to the accuracy of HOOI algorithm.

\vspace{4mm}
 

\noindent
{\bf Proof of Theorem~\ref{thm:HOOI_accuracy}.\ }  
In order to simplify the proof, we denote $V_1 = V_2 = U, V_3 = W$, $n_1 = n_2 = n, n_3  = L$, 
$\del_1 = \del_2 = \del_u$, $\del_3 = \del_w$, 
$r_1 = r_2 = r_u$, $r_3 = r_w$. Also, for $k = 1,2$ or 3, let $\jo$ and $\jt$ be the other two indices, 
i.e., if $k=1$, then $j_1 = 2$ and $\jt = 3$.  Denote
\be \label{eq:sigosigr}
\sigo (n,L,K,M) = \sigt (n,L,K,M) = \sig_u (n,L,r_u,r_w), \quad
\sigr (n,L,K,M) = \sig_w (n,L,r_u,r_w), 
\ee 
\be \label{eq:calD}
\cald(V, \tilV) = \inf_{O \in \calO_r}\, \|\tilV - V\, O\|, \quad \mbox{where} \quad V, \tilV \in \calO_{m,r},\ m \geq r.
\ee  
Fix $k = 1$ or $k=3$, $t \geq 1$, and consider 
\bes
 \tilV_k\upt = \SVD_{r_k} \lkr \calM_k(\tilbA \timjo [\hV_\jo\upto]^T  \timjt [\hV_\jt\upto]^T)\rkr = 
\SVD_{r_k} \lkr \calM_k(\tilbA)\,  (\hV_\jo\upto \otimes \hV_\jt\upto) \rkr.
\ees
It is known (see, e.g., \cite{ke2019_hypergraph}) that
\be \label{eq:hV_ineq}
\cald (\hV_j\upto, V_j) \leq  2\sqrt{2}\,  \cald (\tilV_j\upto, V_j), \quad
\|\hV_j\upto\|\tinf \leq \sqrt{2}\, \del_j, \quad j=1,2,3.
\ee  
Note that, due to \fr{eq:bP_total}, one has $\calM_k(\tilbP) = V_k\, \calM_k(\bTe)\, (V_\jo  \otimes V_\jt)^T$,
so that 
\bes
\calM_k(\tilbP) \,  (\hV_\jo\upto \otimes \hV_\jt\upto) = V_k\, \calM_k(\bTe)\, \lkr  V_\jo^T\, \hV_\jo\upto  
\otimes V_\jt^T \, \hV_\jt\upto \rkr^T.
\ees
Therefore,  due to \fr{eq:hV_ineq}, if $\Err(t-1) \leq 1/4$, one has $\sig_{\min} (V_j^T\, \hV_j\upto) \geq 1/\sqrt{2}$ and 
\be \label{eq:sigrk}
\sig_{r_k} \lkr \calM_k(\tilbP) \,  (\hV_\jo\upto \otimes \hV_\jt\upto)\rkr \geq  1/2\, \sig_k (n,L,K,M).
\ee 
Recall that, by Davis-Kahan theorem 
\be \label{eq:dav_kahan1}
\cald (\tilV_k\upt, V_k) \leq \frac{C\, \left\| \calM_k (\tilbXi) \, (\hV_\jo^T\, \hV_\jt\upto) \right\|}{\sig_k (n,L,K,M)}
= \frac{C\, \left\| \calM_k \lkr \tilbXi  \timjo [\hV_\jo\upto]^T  \timjt [\hV_\jt\upto]^T  \rkr \right\|}{\sig_k (n,L,K,M)}.
\ee 
Hence, it is necessary to derive an upper bound for 
\be \label{eq:Rk_err}
\Ruk = \left\| \calM_k \lkr \tilbXi  \timjo [\hV_\jo\upto]^T  \timjt [\hV_\jt\upto]^T  \rkr \right\| \leq \Ruk_1 + \Ruk_2 + \Ruk_3,
\ee
where 
\begin{align*}
\Ruk_1 & =   \left\| \calM_k \lkr \tilbXi  \timjo  [\hV_\jo\upto - V_\jo \, O_\jo\upto]^T  \timjt  [\hV_\jt\upto]^T  \rkr \right\|, \\
\Ruk_2 & = \left\| \calM_k \lkr \tilbXi  \timjo [V_\jo \, O_\jo\upto]^T  \timjt [\hV_\jt\upto - V_\jt \, O_\jt\upto]^T  \rkr \right\|, \\
\Ruk_3 & = \left\| \calM_k \lkr \tilbXi  \timjo [V_\jo \, O_\jo\upto]^T  \timjt [V_\jt \, O_\jt\upto]^T  \rkr \right\|. 
\end{align*}
Here, $\di O_j\upto = \displaystyle \argmin_{O \in  \calO_r} \, \|\hV_j - V_j \, O\|$, $j=1,2,3$.
\\

First, we derive an upper bound for $\Ruk_1$. 
Note that it follows from Lemma~C1 of \cite{TWIST-AOS2079} that, for any tensor $\bX$ 
with respective ranks $r_1 = r_2$ and $r_3$, one has $\|\calM_k(\bX)\| \leq \sqrt{r_1}\, \|\bX\|$.
Therefore, 
\bes
\Ruk_1   \leq  \sqrt{r_1}\, \left\| \tilbXi  \timjo [\hV_\jo\upto - V_\jo \, O_\jo\upto]^T  \timjt [\hV_\jt\upto]^T  \right\|. 
\ees 
Observe that, for unit vectors $v_1, v_2, v_3$, 
\begin{align*}
& \| \tilbXi  \timjo [\hV_\jo\upto - V_\jo \, O_\jo\upto]^T  \timjt [\hV_\jt\upto]^T  \| = \\
& \max_{v_1, v_2, v_3}\, \lan \tilbXi  \timjo [\hV_\jo\upto - V_\jo \, O_\jo\upto]^T  \timjt [\hV_\jt\upto]^T, 
v_1 \otimes v_2 \otimes v_3 \ran,
\end{align*}
where,  for   $\tilv_{\jo} =  \|\hV_\jo\upto - V_\jo \, O_\jo\upto\|^{-1/2}\, \lkr \hV_\jo\upto - V_\jo \, O_\jo\upto \rkr$, derive
\begin{align*}
& \lan \tilbXi  \timjo [\hV_\jo\upto - V_\jo \, O_\jo\upto]^T  \timjt [\hV_\jt\upto]^T, 
v_1 \otimes v_2 \otimes v_3 \ran = \\
& \|\hV_\jo\upto - V_\jo \, O_\jo\upto\|\, 
\lan \tilbXi, v_k \otimes \tilv_{\jo} \otimes (\hV_\jt\upto\, v_\jt) \ran.
\end{align*}
Since  $\|\hV_\jo\upto - V_\jo \, O_\jo\upto\| = \cald(\hV_\jo\upto, V_\jo)$, $\tilv_{\jo}$ is a unit vector and 
$\|\hV_\jt\upto\|\tinf \leq \sqrt{2}\, \del_\jt$ due to \fr{eq:hV_ineq}, derive that 
\be \label{eq:Ruk1_bound}
\Ruk_1 \leq \sqrt{r_1}\, \cald(\hV_\jo\upto, V_\jo)\, \|\tilbXi\|_{\jt,\sqrt{2}\, \del_\jt},
\ee
where $\|\tilbXi\|_{\jt,\sqrt{2}\, \del_\jt}$ is defined in \fr{eq:incoh_norm} and \fr{eq:calukd}.
Also, by an almost identical calculation, derive
\be \label{eq:Ruk2_bound}
\Ruk_2 \leq \sqrt{r_1}\, \cald(\hV_\jt\upto, V_\jt)\, \|\tilbXi\|_{\jo,\sqrt{2}\, \del_\jo}.
\ee

Now, consider $\Ruk_3$. Recall that, due to \fr{eq:tilbp1}, one can write 
 $\tilbXi = \bXi \tio \Pibot \tit \Pibot$, where $\Pi$ is defined in \fr{eq:tX}.
Hence, if $k=1$ or 2, then 
\beqns
\Ruk_3 & = & \|\calM_1 (\bXi \tio \Pibot \tit \Pibot \tit [V_2 \, O_2\upto]^T  \tir [V_3 \, O_3\upto]^T \|\\
& = &
\|\Pibot\, \calM_1 (\bXi) \lkr (V_3 O_3\upto) \otimes (\Pibot\, V_2\, O_2\upto)\rkr \|.
\eeqns
Since, by construction, $\Pibot\, V_1 = V_1$ and $\Pibot\, V_2 = V_2$, derive that for $k=1$ or 2,
\bes
\Ruk_3 \leq \| \calM_1(\bXi)\, (V_3 \otimes V_2)\| = \|\calM_1 (\bXi \tit V_2^T \tir V_3^T)\|.
\ees
If $k=3$, then obtain a similar result, so that overall
\be \label{eq:Ruk3_1}
\Ruk_3 \leq \| \calM_k(\bXi)\, (V_\jo \otimes V_\jt)\| = 
\| S^{(k)}\|,
\ee 
where, with the correct permutation of indices $\aleph(\ijo, \ijt, \ik)$ 
\be \label{eq:Supk} 
S^{(k)} =  \sum_{(\ijo, \ijt, \ik)} \bXi\lkr \aleph(\ijo, \ijt, \ik) \rkr   \, e_{\ik} \, 
(V_\jo^T e_\jo)^T \otimes (V_\jt^T e_\jt)^T.
\ee  
All components of the sum $S^{(k)}$ are rank one matrices and $|\bXi(\aleph(\ijo, \ijt, \ik))| \leq 1$, 
so that 
\bes
\left\|e_{\ik} \, (V_\jo^T e_\jo)^T \otimes (V_\jt^T e_\jt)^T \right\|^2 = 
\|e_{\ik}\|\, \|V_\jo^T e_\jo\|^2 \, \|V_\jt^T e_\jt\|^2
\leq \del_\jo^2\, \del_\jt^2.
\ees
Also,  by direct calculations obtain that
\begin{align*}
\left\|\EE \lkr S^{(k)} \, (S^{(k)})^T \rkr \right\| & \leq C\, \rhon \, (\del_\jo^2\, n_\jo)\, (\del_\jt^2\, n_\jt),\\
\left\|\EE \lkr (S^{(k)})^T\, S^{(k)} \rkr \right\| & \leq C\, \rhon \, n_k\, (\del_\jo^2\, n_\jo)\, (\del_\jt^2\, n_\jt).
\end{align*}
Hence, application of Bernstein inequality to $S^{(k)}$ in \fr{eq:Supk} yields that, 
with probability at least $1 - n^{-\tau}$,
\be \label{eq:Ruk3_bound}
\Ruk_3 \leq \| S^{(k)}\| \leq \Ctau\, \sqrt{\rhon\, n_k\, (\del_\jo^2\, n_\jo)\, (\del_\jt^2\, n_\jt)\, \log n}.
\ee
Now, combine \fr{eq:Ruk1_bound}, \fr{eq:Ruk2_bound} and \fr{eq:Ruk3_bound}, in a view of Corollary~\ref{cor:incoherent_error}
in Section~\ref{sec:incoherent_norm}. 
Obtain that, with probability at least $1 - c\, n^{-\tau}$,
\begin{align}  
\Ruk \leq \Ctau\, \lfi \sqrt{\rhon}\, \sqrt{\log n}\,  \sqrt{n_k}\, (\del_\jo\, \sqrt{n_\jo})\, (\del_\jt\, \sqrt{n_\jt}) \right.
& +   \sqrt{K\, M}\,  \lkv \cald(\hV_\jo\upto, V_\jo)\, \Del_{\jt} \right.   \nonumber \\
&  \left. \left. + \cald(\hV_\jt\upto, V_\jt)\, \Del_{\jo} \rkv \rfi,   \label{eq:Ruk_total}
\end{align}
where   upper bounds for  $\Del_k$ are given in \fr{eq:Delk} and \fr{eq:Delk_simple}.
Now, plugging \fr{eq:Rk_err} and \fr{eq:Ruk_total}   into \fr{eq:dav_kahan1}, derive that 
\begin{align} \label{eq:dist_err}
\cald (\tilU\upt, U) + \cald  (\tilW\upt, W) & \leq \eps (n,L) +  
\sqrt{K\, M}\, \lkv \cald (\hU\upto, U)\, \Del_3 (n, L, \del_w) \right. \\
& + \left.
\cald (\hW\upto, W)\, \Del_1 (n, L, \del_u)\rkv,
\nonumber
\end{align} 
where 
\bes 
 \eps (n,L) = \Ctau \, \lkv \frac{\sqrt{\rhon\, n\, \log n\, (\del_u^2\, n)\, (\del_w^2\, L)}}{\sig_1 (n,L,K,M)} 
+ \frac{\sqrt{\rhon\, L\, \log n}\, (\del_u^2\, n)}{\sig_3 (n,L,K,M)} \rkv,
\ees
and $\sigo (n,L,K,M)$ and $\sigr (n,L,K,M)$ are defined in \fr{eq:sigosigr}.
Recall that  $r_u = \barK \, M \leq K\, M$ and, by Lemma~\ref{lem:sig_strength}, one has, 
with probability at least $1  - c\, n^{-\tau}$, that
\bes
\sig_u^2 (n,L,r_u,r_w) \geq C \,\rhon^2\, n^2\, L, \quad 
\sig_w^2 (n,L,r_u,r_w) \geq C\,  M^{-1} \rhon^2 n^2  L. 
\ees
Hence, using \fr{eq:hV_ineq}  and plugging $\Del_1 (n, L, \del_u)$ and $\Del_3 (n, L, \del_w)$ into \fr{eq:dist_err}, obtain
\fr{eq:err_rel} where $\eps (n,L)$ and $\delsnl$ are given  in \fr{eq:epsnl} and \fr{eq:delsnL}, respectively.

Validity of the rest of the Proposition follows from the fact that, if $0<s<1$ and $d_k \leq s \, d_{k-1}  + \eps$,
then $d_k \leq s^k\, d_0 + \eps\, (1-s)^{-1}$. Moreover, if $s \leq 1/2$   and $d_0 \leq 1/2$, then 
$d_k \leq 3\, \eps$ provided $k \geq k_{\max} = \log_2 (\eps^{-1})$.
\\

\medskip


\noindent
{\bf Proof of Theorem~\ref{thm:initial}.\ }  
First, we obtain an upper bound on $\sinTe (\tilUo, U)$. 
Note that  $\sig_{M\barK} (\scrT) \geq \sig^2_{\min} (\bTe)$
and $\|\Pibot\, (\hscrT -  \scrT)\, \Pibot\| \leq \|\hscrT - \scrT\|$,
where $\sig_{\min} (\bTe)$ is defined in \fr{eq:sig_stren_G}.
Lemma~\ref{lem:sig_strength} yields
\be \label{eq:lower_sig_ini} 
\PP \lfi \sig^2_{\min}(\bTe) \geq C\,  M^{-1} \rhon^2 n^2  L \rfi \geq 1 - c\, n^{-\tau}.  
\ee
Therefore, by Davis-Kahan theorem, obtain that 
\be \label{eq:dk_ini_U}
\PP \lfi \left\| \sinTe (\tilUo, U) \right\| \leq \Ctau\, M\, (\rhon^2 n^2  L)^{-1}\, \|\hscrT - \scrT\| \rfi \geq 1 - c\, n^{-\tau}. 
\ee
Hence, we need to construct an upper bound for $\|\hscrT - \scrT\|$.  
Observe that 
\be    \label{eq:err_hscrT_ini}
\|\hscrT - \scrT\| \leq     \|\calH(\hscrT - \scrT)\| + \max_{i \in [n]}\, |\scrT(i,i)| 
\ee   
Here,   
\bes   
\|\calH(\hscrT - \scrT)\| =  \left\| \sum_{\linL} \, \calH \lkr (A\upl)^2 - (P\upl)^2 \rkr \right\|. 
\ees 
If $n \rhon \geq c_0$ for some constant $c_0 >0$, then, similarly to \cite{pensky2021clustering},
obtain that, with probability at least $1 -  c\, n^{-\tau}$, one has
\be \label{eq:R1_strong_sig}
\|\calH(\hscrT - \scrT)\|  \leq \Ctau \lkr \rhon\, n\,  \sqrt{L}\, \log n + (\rhon\, n)^{3/2}\, \sqrt{L\,  \log n} + (\log n)^2  \rkr.
\ee  
On the other hand, under condition \fr{eq:rhon_main_cond},  if $n \rhon \leq c_0$, then, by Theorem~5 of \cite{lei2021biasadjusted}, obtain
with probability at least $1 -  c\, n^{-\tau}$, that
\be \label{eq:R1_weak_sig}
\|\calH(\hscrT - \scrT)\| \leq \Ctau \lkr \rhon\, n\, \sqrt{L\, \log n} +  L\, n\, \rhon^2 \rkr
\ee    
Combining \fr{eq:R1_strong_sig} and \fr{eq:R1_weak_sig} and recalling that $L \geq \tilc\, (\log n)^2$ by Assumption~{\bf A6}, derive 
\be \label{eq:R1_str_weak_sig} 
\PP \lfi \|\calH(\hscrT - \scrT)\| \leq \Ctau \lkr \rhon\, n\, \log n\,  \sqrt{L} +  (\rhon\, n)^{3/2}\, \sqrt{L\,  \log n}  \rkr \rfi \geq 
1 -  c\, n^{-\tau}.
\ee 
Therefore, combining \fr{eq:dk_ini_U}, \fr{eq:err_hscrT_ini}, \fr{eq:R1_str_weak_sig} and the fact that 
$\displaystyle \max_{i \in [n]}\, |\scrT(i,i)| \leq C \, \rhon^2\, n\, L$, obtain that 
with probability at least $1 - c\, n^{-\tau}$ one has
$\|\sinTe (\tilUo, U)\| \leq \Ctau\, \tileps(n,L)$, where 
\be \label{eq:tepsnL}
\tileps(n,L) = \frac{M\, \log n}{\rhon n \sqrt{L}} + \frac{M\, \log n}{\sqrt{\rhon\, n\, L}}
+ \frac{M}{n}.
\ee 
Then, since $\tileps(n,L)/\delsnl \to 0$ as $n \to \infty$, one has  $\|\sinTe (\tilUo, U)\| \to 0$ as $n \to \infty$.
Hence, by \fr{eq:hV_ineq}, obtain that $\|\hU^{(0)}\|\tinf \leq \sqrt{2}\,  \delu$ and $\|\sinTe (\hUo, U) \| \leq 1/4$.
\\

Now, we study the errors of $\tilWo$ and $\hWo$. Recall that 
\bes
\tilWo = \SVD_r (\calM_3(\tilbA \tio [\hUo]^T  \tit [\hUo)^T]),
\ees 
so that repeating the steps of the proof of 
Theorem~\ref{thm:HOOI_accuracy}
with $k=3$, $t=1$,  $\tilV_k^{(1)}  = \tilWo$ and  
$\tilV_\jo^{(0)}  =  \tilV_\jt^{(0)} = \tilUo$,  using \fr{eq:sigrk}--\fr{eq:Rk_err} , derive
that, for $\cald (\tilWo, W)$ defined in \fr{eq:calD}, one has
\be \label{eq:th2_tilWo} 
\cald (\tilWo, W) \leq  C\, \sqrt{M}\, (\rhon \, n \, \sqrt{L})^{-1}\,  (R_1\upo +  R_2\upo + R_3\upo).
\ee 
Here, similarly to the proof of Theorem~\ref{thm:HOOI_accuracy}
with $\del_\jo = \del_\jt = \del_u$,
\bes
R_1\upo +  R_2\upo \leq 2\, \sqrt{K\, M}\, \cald(\hUo, U)\, \|\tilbXi\|_{1,\sqrt{2}\, \delu}.
\ees
It follows from \fr{eq:Delk_simple} of Proposition~\ref{prop:incoherent_error} that, with probability at least $1 - c\, n^{-\tau}$,
\bes
\|\tilbXi\|_{1,\sqrt{2}\, \delu}  
\leq \Ctau \,  \sqrt{K\, M}\,  \sqrt{\rho\, \maxnL}\ (\log n)^3 \,  \log \log n.
\ees
Hence, with probability at least $1 - c\, n^{-\tau}$,
\be \label{eq:th2_R1R2upo}
R_1\upo +  R_2\upo \leq \Ctau\, K\, M\, \sqrt{\rho\, \maxnL}\ (\log n)^3 \,  \log \log n \  \tileps (n,L),
\ee
where $\tileps (n,L)$ is defined in \fr{eq:tepsnL}.
Also, using \fr{eq:delu_delw} and   \fr{eq:Ruk3_bound}, obtain that,  with probability at least $1 - c\, n^{-\tau}$,
\be  \label{eq:Ruk3_boundW}
R_3\upo  \leq \Ctau\,   M\,K\,  \sqrt{\rhon\, L}\, (\log n)^{5/2}.  
\ee 
Plugging \fr{eq:th2_R1R2upo} and \fr{eq:Ruk3_boundW} into \fr{eq:th2_tilWo} and noting that $(n L)^{-1}\, \maxnL = (\minnL)^{-1}$, derive
\bes
\PP \lfi \| \sin \Te(W, \tilWo) \| \leq \Ctau\, \breve{\eps} (n,L) \rfi \geq 1 - c\, n^{-\tau}
\ees
where
\bes  
\breve{\eps} (n,L)   =  K\,   M^{3/2} \,  (\log n)^{5/2} \lkv \frac{\log \log n\, \sqrt{\log n}}{\sqrt{\rhon \, n\, \minnL}}\ 
\tileps (n,L) + \frac{1}{n\, \sqrt{\rhon}} \rkv,
\ees 
which completes the proof.
\\

\medskip 


\noindent
{\bf Proof of Corollary~\ref{cor:initial}.\ }  
Validity of Theorem~\ref{thm:initial} is guaranteed by $R_i \to 0$ as $n,L \to \infty$, $i=1,2,3$,  where
\begin{align*}
R_1 & = M\, \log n \,   (\rhon\, n \sqrt{L})^{-1},\\
R_2 & =  M\, \log n \,  (\rhon\, n\, L)^{-1/2},\\
R_3 & = K\,   M^{3/2} \,  (\log n)^{5/2}\, \lkr n^2\, \rhon \rkr^{-1/2}.
\end{align*}
Note that $\delsnl <1/2$ implies that
\be  \label{eq:delsnl_cond12}
R = \Ctau\, K\, M^{3/2}\, (\log n)^3\,  \log \log n  \, \lkr \rhon\, n\, \min(n,L) \rkr^{-1/2} < 1/2.
\ee  
It is easy to see that $R_2 = o(R)$ and $R_3 = o(R)$, so that, by \fr{eq:delsnl_cond12}, $R_2 = o(1)$ and $R_3 = o(1)$ as 
$n,L \to \infty$. Hence, Corollary holds if $R_1  = o(1)$ as $n,L \to \infty$, which completes the proof. 
\\

\medskip


\noindent
{\bf Proof of Theorem~\ref{thm:clust_between}.\ }  
%
In order to prove that clustering is perfect, one needs to ensure that
\bes
 \min_{O \in \calO_{L,r}}\, \|\hW - W O\|\tinf = o(\sqrt{M}/\sqrt{L}), \quad n,L \to \infty.
\ees 
For this purpose, we consider the SVD $W^T \hW = O_1 \Lam_W O_2^T$ with $O_1, O_2 \in \calO_r$ and set 
$O_w = O_1 O_2^T$. We shall derive an upper bound for $\|\hW - W \, O_w\|\tinf$ using Theorem~4 of \cite{pensky2024daviskahan}.
For the readers' convenience, we present a version of this theorem here.

\begin{thm}\label{thm:nonsym_up_bound} {\bf (Theorem 4 of \cite{pensky2024daviskahan})}.\  
Let  $X, \hX \in \RR^{n \times m}$ have the  SVD expansions 
\bes
X = U D  V^T + U_{\perp} D_{\perp} V_{\perp}^T, \quad 
\hX = \hU \hD  \hV^T + \hU_{\perp} \hD_{\perp} \hV_{\perp}^T,
\ees
where $U, \hU \in \calO_{n,r}$,   $V, \hV \in \calO_{m,r}$,
$U_{\perp}, \hU_{\perp} \in \calO_{n,(m \wedge n)-r}$,   $V_{\perp}, \hV_{\perp} \in \calO_{m,(m \wedge n)-r}$, 
$D = \diag(d_1, ...,  d_r)$, $\hD = \diag(\hd_1, ...,  \hd_r)$,  
$d_k = \sigma_k(X)$,  and $\Xi = \hX - X$.
Let $\tDel_0 = d_r ^{-1}\, \|\Xi\|$, $ \tDel_{2,\infty} = d_r ^{-1}\, \|\Xi\|_{2,\infty}$.
If,  for any $\tau>0$,
there exists a constant $\Ctau$ and deterministic quantities $\tepso$  and $\tepstinf$
that depend on $n$, $m$, $r$ and possibly $\tau$,   such that simultaneously, 
\bes 
\PP \lfi  \tDelo \leq \Ctau\, \tepso, \    \tDeltinf \leq \Ctau\,  \tepstinf  \rfi  
  \geq 1 - n^{-\tau} 
\ees
for $n$ large enough, and $\tepso < 1/4$, then, 
\be \label{eq:D-K_th}
\PP \lfi \|\hU - U W_U\|\tinf \leq  \Ctau \lkv  \|U\|\tinf \,   \tepso +
\tepstinf +  \tepso\, (\tepstinf + d_{r+1}\, d_r^{-1}) \rkv    \rfi  
 \geq 1 - n^{-\tau}.   
\ee 
\end{thm}

\medskip
\medskip

\noindent
We apply this theorem with $X = \calM_3 (\tilbP)\, (\hU \otimes \hU)$, $\hX = \calM_3 (\tilbA)\, (\hU \otimes \hU)$ 
and $\Xi = \calM_3 (\tilbXi)\, (\hU \otimes \hU)$,
where $\tilbXi = \tilbA - \tilbP$.
Note that, due to $\tilbP = \bTe \tio U \tit  U \tir W$, one has 
\bes  
X = \calM_3 (\tilbP)\, (\hU \otimes \hU) = W \,\calM_3(\bTe) \, \lkr (U^T \hU) \otimes  (U^T \hU)\rkr,
\ees 
so that the SVD of $X$ is of the form 
\bes
X = W\, \tilO\, \tilD\, \tilV^T, \quad \tilO \in \calO_r, \ \ \tilV \in \calO_{(m \barK)^2, r}. 
\ees
Let the SVD of $\hX$ be of the form 
\bes  
\hX = \hW \hD \hV^T, \quad \hV \in \calO_{(m \barK)^2, r}.
\ees 
Observe that, by Corollary~\ref{cor:initial}, one has  $\sig_{\min} (U^T \hU) \geq 1/\sqrt{2}$. 
Therefore, by Lemma~\ref{lem:sig_strength} 
\be \label{eq:th3_1}
\sig_{\min}\, (X) \geq 0.5\, \sig_{r} (\calM_3(\bTe)) \geq \Ctau\, M^{-1/2}\, \rhon\, n\, \sqrt{L}.
\ee 
Recall also  that, by Theorem~\ref{thm:HOOI_accuracy}, $\|\hU\|\tinf \leq \sqrt{2}\, \del_u$.
Since $\rank(\hU) = M\,K$, obtain that 
\bes
\|\hX - X\| =  \left\| \calM_3 \lkr \tilbXi  \tio \hU^T  \tit \hU^T  \rkr \right\|
\leq (M\, K)^2 \|\tilbXi\|_{1, \sqrt{2}\,\delu} \leq  (M\, K)^2 \, \Del_1,
\ees 
where  $\Del_1$ is of the form \fr{eq:Delk_simple}.
Then,   Wedin's theorem yields that, with probability at least $1 - c\, n^{-\tau}$,
\bes
\tepso = \|\sinTe (\hW,W)\| + \|\sinTe (\hV,\tilV)\| \leq  \frac{\Ctau\, (M\, K)^2 \Del_1}{\sig_{\min}(X)},
\ees 
which, due to $(n\,L)/\max(n,L) = \min(n,L)$ and \fr{eq:th3_1}, appears as 
\be \label{eq:tepso}
\tepso = \|\sinTe (\hW,W)\|  \leq 
\frac{\Ctau\, M^{5/2}\, K^2\, (\log n)^{3}\, \log \log n}{\sqrt{\rhon\, n\,  \minnL }}.
\ee

Now, in order to to apply Theorem~4 of \cite{pensky2024daviskahan},
we need to find an upper bound for 
\bes
\|\tilXihu\|\tinf \equiv \left\| \calM_3 \lkr \tilbXi  \timjo \hU^T  \timjt \hU^T  \rkr \right\|\tinf =
\max_{\linL} \|\xi\upl\, (\hU \otimes \hU)\|,
\ees
where  $\xi\upl = \vect(\tilXi\upl) = (\calM_3(\tilbXi))^T\, e_l$. 
Due to $\|\hU\|\tinf \leq \sqrt{2}\, \delu$ and $\rank(\hU) \leq M\,K$,  
application of Proposition~\ref{prop:matrix_error}  yields, for any $x>0$ and $l \in [L]$
\begin{align*}
\PP \lkr \|\xi\upl\, (\hU \otimes \hU)\| > x \rkr & \leq 
\PP \lkr \max_{i,j}\, \left\|\xi\upl\, (\hU(:,i) \otimes \hU(:,j))\right\| >  (K\, M)^{-1}\, x \rkr \\
 & \leq \PP \lkr \sup_{u,v \in \calV (\sqrt{2}\, \delu)}\, \langle  \Xi\upl, u v^T   \rangle   >  (K\, M)^{-1}\, x  \rkr  \leq  c n^{-\tau},
\end{align*} 
provided $x > K\, M \Del (n, \sqrt{2}\, \delu)$, where $\Del (n,\del)$ is defined in \fr{eq:Delndel}.
Plugging in the values and taking a union bound over $\linL$,  derive that
\bes 
\PP \lfi \|\tilXihu\|\tinf > \Ctau\, \tilR(n,\del) \rfi \leq  c n^{-(\tau- \tau_0)},
\ees 
where
\bes 
\tilR(n,\del) = K M\, \sqrt{\rhon  (n \rhon + \log n)\log n \log \log n}
+  n^{-1} (K M)^2 \, (n \rhon + \log n) (\log n)^2 \log \log n.
\ees 
Hence, using \fr{eq:th3_1}, obtain that  
\be \label{eq:probab_tepstinf}
\PP \lfi   \frac{\|\tilXihu\|\tinf}{\sig_{\min}(X)} \leq  \teps\tinf \rfi \geq 1- c n^{-(\tau- \tau_0)},
\ee
where
\begin{align}  \label{eq:tepstinf}
\teps\tinf = 
 \frac{\Ctau\, K M^{3/2}}{\sqrt{L}} 
 \lkv \frac{\sqrt{\log n\, \log \log n}}{\sqrt{n}} \right. & + \frac{\log n  \sqrt{\log \log n}}{n\, \sqrt{\rhon}}\\
 & + \left. \frac{KM\, (\log n)^3\, \log \log n}{n^2\, \rhon} \rkv. \nonumber
\end{align}

Now, we apply Theorem~4  
 with $X = \calM_3 (\tilbP)\, (\hU \otimes \hU)$, $\hX = \calM_3 (\tilbA)\, (\hU \otimes \hU)$,  
$\Xi = \calM_3 (\tilbXi)\, (\hU \otimes \hU)$,
matrices $U$, $\hU$, $W_U$ and $V$ replaced by $W$, $\hW$, $O_w$ and $\tilV$, respectively,
 $d_r = \sig_{\min} (X)$ given by \fr{eq:th3_1}, $d_{r+1} =0$ 
and $\|W\|\tinf = C\, L^{-1/2} \, (M K)^{1/2}$. 
Using \fr{eq:D-K_th}, obtain
\be \label{eq:hWW}
\|\hW - W\, O_w\|\tinf  \leq \Ctau\, \lkv  \sqrt{M}/\sqrt{L}\,   (\tepso + \tepso^2) + \teps\tinf (1 + \tepso)\rkv,
\ee 
where the upper bounds for  $\tepso$ and $\teps\tinf$ are given in \fr{eq:tepso} and \fr{eq:tepstinf},
respectively.  Since $\tepso \to 0$ as $n,L \to \infty$ under conditions of the theorem,  
plugging  $\tepso$ and $\teps\tinf$ into \fr{eq:hWW}, obtain that 
\be  \label{eq:mainW_ineq}
\PP \lfi \|\hW - W\, O_w\|\tinf  \leq \Ctau\,   \sqrt{M}/\sqrt{L}\,  \tilR_0 (n,L) \rfi \geq 1 - c\, n^{-\tau},
\ee
where
\begin{align*} 
\tilR_0  (n,L) & = \frac{(KM)^{3/2}\, (\log n)^3 \, \log \log n}{\sqrt{\rhon\, n\, \minnL}}
  +   \frac{K\, M\,   \log n\, (\log \log n)^{1/2} }{n \sqrt{\rhon}\, \sqrt{n}} \\
 & + \frac{(K\, M)^2\, (\log n)^3\, \log \log n }{n^2\, \rhon}   
+ \frac{K\, M\, \sqrt{\log n\, \log \log n}}{\sqrt{n}}. 
\end{align*} 
Re-write $\tilR_0  (n,L)$ as 
\begin{align} 
\tilR_0  (n,L) & = \frac{(KM)^{3/2}\, (\log n)^3 \, \log \log n}{\sqrt{\rhon\, n\, \minnL}}
\lkv 1 + \frac{\sqrt{\min(n,L)}\, (K\, M\, \log \log n)^{-1/2} }{(\log n)^{2} \sqrt{n}}   \right. \label{eq:tilR0} \\
& + \left. \frac{\sqrt{K\, M}}{\sqrt{n^2\, \rhon}}\, \frac{\sqrt{\min(n,L)}} {\sqrt{n}} \rkv
+ \frac{K\, M\, \sqrt{\log n\, \log \log n}}{\sqrt{n}}. \nonumber
\end{align} 
Since $\delsnl \leq 1/2$, where $\delsnl$  is defined in \fr{eq:delsnL}, implies that
$(n^2\,\rhon)^{-1}\, K\,M \to 0$ as $n \to \infty$,
the second and the third terms in the square brackets in \fr{eq:tilR0} tend to zero as $n,L \to \infty$.
Therefore,  $\tilR_0  (n,L) =  (\log n)^{-1}\, R(n,L) (1 + o(1))$, where $R(n,L)$ is defined in \fr{eq:RnL}.

Now, for matrices $Y$ and $\hY$ with elements defined in \fr{eq:Y_hY},
due to Lemma~\ref{lem:matr_W_new}, \fr{eq:mainW_ineq} and the relation between  $\tilR_0 (n,L)$ and $R(n,L)$, one has
\bes
|\hY(l_1, l_2) -  Y(l_1, l_2)| \leq \Ctau L^{-1}\, M \, (\log n)^{-1}\, R (n,L)).
\ees
Taking a union bound in the last inequality and adjusting the value of $\Ctau$, derive that 
\begin{align*}
& \min \hY(l_1, l_2)   \geq \Ctau\, L^{-1}\, M \, [1 - \tilCtau \, (\log n)^{-1}\, R (n,L)], & \mbox{if} \quad s(l_1) = s(l_2);\\
& \max |\hY(l_1, l_2)|   \leq \Ctau\, L^{-1}\, M \,   (\log n)^{-1}\, R (n,L), & \mbox{if} \quad s(l_1) \neq s(l_2).
\end{align*}
Due to $R(n,L) = o(1)$ as $n,L \to \infty$,  for $n$ large enough and $T(n,L)$, defined in \fr{eq:T-value}, one has  
$\hY(l_1, l_2)  > T$ if $s(l_1) = s(l_2)$, and $|\hY(l_1, l_2)| < T(n,L)$ if $s(l_1) \neq s(l_2)$.
Therefore, with high probability, for matrix $\scrX$ in Algorithm~\ref{alg:between}, one has
$\scrX(l_1, l_2) =1$  if $s(l_1) = s(l_2)$, and $|\scrX(l_1, l_2)|= 0$ if $s(l_1) \neq s(l_2)$,
which guarantees perfect clustering.
\\

\medskip 


\noindent
{\bf Proof of Corollary~\ref{cor:comparison}.\ }  
Note that in the setting above, one derives that
\begin{align}
  \Err(t_{\max}) & \asymp (\log n)^{3/2}\, \lkr \rhon\, n\, \minnL \rkr^{-1/2}, \label{eq:ertmax}\\
  \Err(0) & \asymp \log n \, (\rhon\, n\, \sqrt{L})^{-1} + \log n \, (\rhon\, n\, L)^{-1/2} 
+ (\log n)^{5/2} \, (n \sqrt{\rhon})^{-1}.  \label{eq:err0}
\end{align}
Now we consider two cases: $n \rhon = o \lkr (\log n)^{-1/2} \rkr$ and $C \, (\log n)^{-1/2} \leq n \rhon \leq C\, \log n$.

If  $n \rhon = o \lkr (\log n)^{-1/2} \rkr$, then,
due to \fr{eq:nLtau}
\bes
\Err(0) \geq   \Om \lkr   (\log n)^{3/2}\, \lkr \rhon\, n\, L \rkr^{-1/2} \rkr +  
O \lkr (\log n)^{5/2} \, (n^2 \, \rhon)^{-1/2} \rkr,
\ees
hence, it follows from \fr{eq:ertmax} and \fr{eq:err0} that $\Err(t_{\max}) = o(\Err (0))$ for any relationship between $n$ and $L$.

If $C \, (\log n)^{-1/2} \leq n \rhon \leq C\, \log n$, then
in order to be more specific, we  assume that $n \rhon \asymp (\log n)^\ga$
where $-1/2 \leq \ga \leq 1$.   The latter yields   
\begin{align*}
  \Err(t_{\max})   & \asymp (\log n)^{\frac{3-\ga}{2}}\,  \minnL^{-\frac12},  \\
  \Err(0)  & \asymp [(\log n)^{1- \ga/2} + (\log n)^{1 - \ga}] \,  L^{-\frac12} +  
  (\log n)^{\frac{5-\ga}{2}} \, n^{-\frac12}.   
\end{align*}
Since $(\log n)^{1- \frac{\ga}{2}} + (\log n)^{1 - \ga}  \leq 2\, (\log n)^{(3 -\ga)/2}$, derive that  $L > n$ implies  
 $\Err(t_{\max}) = o(\Err(0))$ as $n, L \to \infty$. 
If $L \leq n$, then 
\bes
\Err(t_{\max})   \asymp (\log n)^{(3-\ga)/2}\,  L^{-1/2}, \quad
\Err(0) \leq C\, \Err(t_{\max}) + (\log n)^{(5-\ga)/2} \, n^{-1/2}.
\ees
Therefore, $\Err(0) = O \lkr \Err(t_{\max}) \rkr$ provided the second condition in \fr{eq:unless} holds.
Otherwise, $\Err(t_{\max}) = o(\Err(0))$ as $n, L \to \infty$.


\subsection{Proofs of Lemmas in Sections~\ref{sec:tensor_components} and~\ref{sec:est_clust}} 
This section presents proofs of supplementary lemmas.

\vspace{4mm}

 
\noindent
{\bf Proof of Lemma~\ref{lem:Tucker_for_tilbP}  }  
Validity of the lemma largely follows from the narrative before it. 
Formulas \fr{eq:bP_total} and \fr{eq:bTe} can be obtained from \fr{eq:tensP}--\fr{eq:tens_pres} 
by standard tensor manipulations. The fact that  $\rank(\barU) = M\, \barK$ follows from 
Lemma~\ref{lem:matr_U_LamU} in Section~\ref{sec:matrU_W}.

Recall that slices $G\upl = \bG(:,:,l)$ of tensor $\bG$ with $s(l) = m$ 
are block diagonal with $(G\upl)_{m,m} = \tilB\upl \neq 0$, 
and all other blocks $(G\upl)_{m_1,m_2}=0$.  
Note also that, due to the symmetry of matrices $(G\upl)_{m,m}$ and $G\upl$, slices of the tensors $\bH\upm$ and $\bH$
are also symmetric. 
Since matrices $H\upm \in \RR^{K_m^2 \times r_m}$ are obtained by vectorizing symmetric matrices 
$\bH\upm(:,:,j)$, $j \in [r_m]$, in \fr{eq:WH_present}  one has 
\be \label{eq:r_rm}
r_m \leq K_m(K_m+1)/2, \quad \sum_m  r_m = r 
\ee
which establishes the upper bounds on $r_m$. 
Finally, observe  that, due to the block-diagonal structure of slices $G\upl$, matrix $W$ also has a block structure.
Specifically, matrix $W$ is formed by scrambled  concatenation of matrices $W\upm  \in \RR^{L \times r_m}$,  
and hence \fr{eq:W_str}  holds. 
\\

\medskip

 
\noindent
{\bf Proof of Lemma~\ref{lem:matr_U_LamU}. }  
It is easy to check that  
$\tilU_P\upl = \tilU\upm \widetilde{O}\upl$ for some $\widetilde{O}\upl \in \calO_{K_m}$, when $s(l)=m$. 
Validity of the first part of \fr{eq:matrU} follows from the fact that 
\bes
\|  (\tilU_P^{(l_1)})^T \tilU_P^{(l_2)} \| = \| \tilU\upm\|^2 =1 \quad \mbox{if} \quad  s(l_1) = s(l_2)=m.
\ees  
To prove validity of the second part, 
recall that, it follows from \fr{eq:X_P_svds} that 
\bes
\tilU\upm = \tilX\upm\, \tilO\upm \, (\tilD\upm)^{-1},
\ees
where $\tilO\upm \in \calO_{K_m}$. Since, by \fr{eq:tilD_upm} of 
Lemma~\ref{lem:tilD_upm}, one has $\|(\tilD\upm)^{-1}\| \leq C n^{-1/2}$, derive that,
for $s(l_1) = m_1$ and $s(l_2) = m_2$  
\begin{align*}
 \|  (\tilU_P^{(l_1)})^T \tilU_P^{(l_2)} \|  = \| (\tilU^{(m_1)})^T \,  \tilU^{(m_2)} \| & = 
\|(\tilD{(m_1)})^{-1}\, (\tilX^{(m_1)})^T\, \tilX^{(m_2)} \, (\tilD{(m_2)})^{-1}\|\\ 
& \leq C \, n^{-1} \, \| (\tilX^{(m_1)})^T \, \tilX^{(m_2)}  \|.
\end{align*} 
To complete the proof of \fr{eq:matrU}, apply  Lemma~5 of \cite{pensky2024signed}, 
which, under the identical assumptions,  establishes that, 
with probability at least $1 - c\,  n^{-\tau}$, one has 
\bes
\| (\tilX^{(m_1)})^T \, \tilX^{(m_2)}  \|   \leq C    \, \sqrt{n\, \log n}.
\ees

Now, we prove \fr{eq:LamU}  and \fr{eq:for_del1}.
Since $(\tilU^{(m)})^T \,  \tilU^{(m)} = I_{K_m}$, matrix $\barU^T \barU$ is of the form
\bes
\barU^T \barU = I_{\barK\, M} + \frZ, \quad \frZ \in \RR^{(\barK M) \times (\barK M)},
\ees 
where $\frZ$ is a block matrix with blocks $\frZ^{(m_1, m_2)} = (\tilU^{(m_1)})^T \, \tilU^{(m_2)} \in \RR^{K_{m_1} \times K_{m_2}}$,
and $\frZ^{(m,m)} = 0$. 
Note also that, due to $\barU^T \barU = O_U\, \Lam_U^2\, O_U^T$, one has
\be \label{eq:Lam_U_minmax}
\lam^2_{\min}(\Lam_U) \geq 1 - \lam_{\max}(\frZ), \quad 
\lam^2_{\max}(\Lam_U) \leq 1 + \lam_{\max}(\frZ).
\ee 
By Theorem~P10.1.1 of \cite{Rao_Rao_1998} and \fr{eq:matrU} , derive that
\bes
\lam_{\max}(\frZ) \leq \sum_{m_1 \neq m_2=1} \|\frZ^{(m_1, m_2)}\| \leq M^2\, \max_{m_1 \neq m_2} \, \|(\tilU^{(m_1)})^T \, \tilU^{(m_2)}\|
\leq C \,  M^2\, n^{-1/2} \, \sqrt{\log n},
\ees
which, together with \fr{eq:Lam_U_minmax}, proves \fr{eq:LamU}. Therefore, for $n$ large enough, one has 
$\lam_{\min}(\Lam_U) \geq 1/2$, and  $\rank(\barU) = \barK\, M$. 

Now, we prove \fr{eq:for_del1}. Note that $U = \barU\, O_U \Lam_U^{-1}$, so that, by  \fr{eq:LamU} 
and the properties of the two-to-infinity norm, obtain
\be \label{eq:Utinf}
\|U\|\tinf \leq \|\barU\|\tinf \, \|\Lam_U^{-1}\| \leq 2\, \|\barU\|\tinf  \leq 2\, \sqrt{M}\, \max_{\minM}\,  \|\tilU\upm\|\tinf.
\ee 
Hence, due to $\tilU\upm = \tilX\upm\, \tilO\upm \, (\tilD\upm)^{-1}$, Assumption~{\bf A2}
and  \fr{eq:tilD_upm}, obtain  
\bes
\|\tilU\upm\|\tinf \leq     \|\tilX\upm\|\tinf \, \|(\tilD\upm)^{-1}\| \leq C\, n^{-1/2},
\ees
which, together with \fr{eq:Utinf}, completes the proof of \fr{eq:for_del1}.
\\

\medskip

\begin{figure*}[t!]   
\centering
\[\includegraphics[width=14cm, height=6cm]{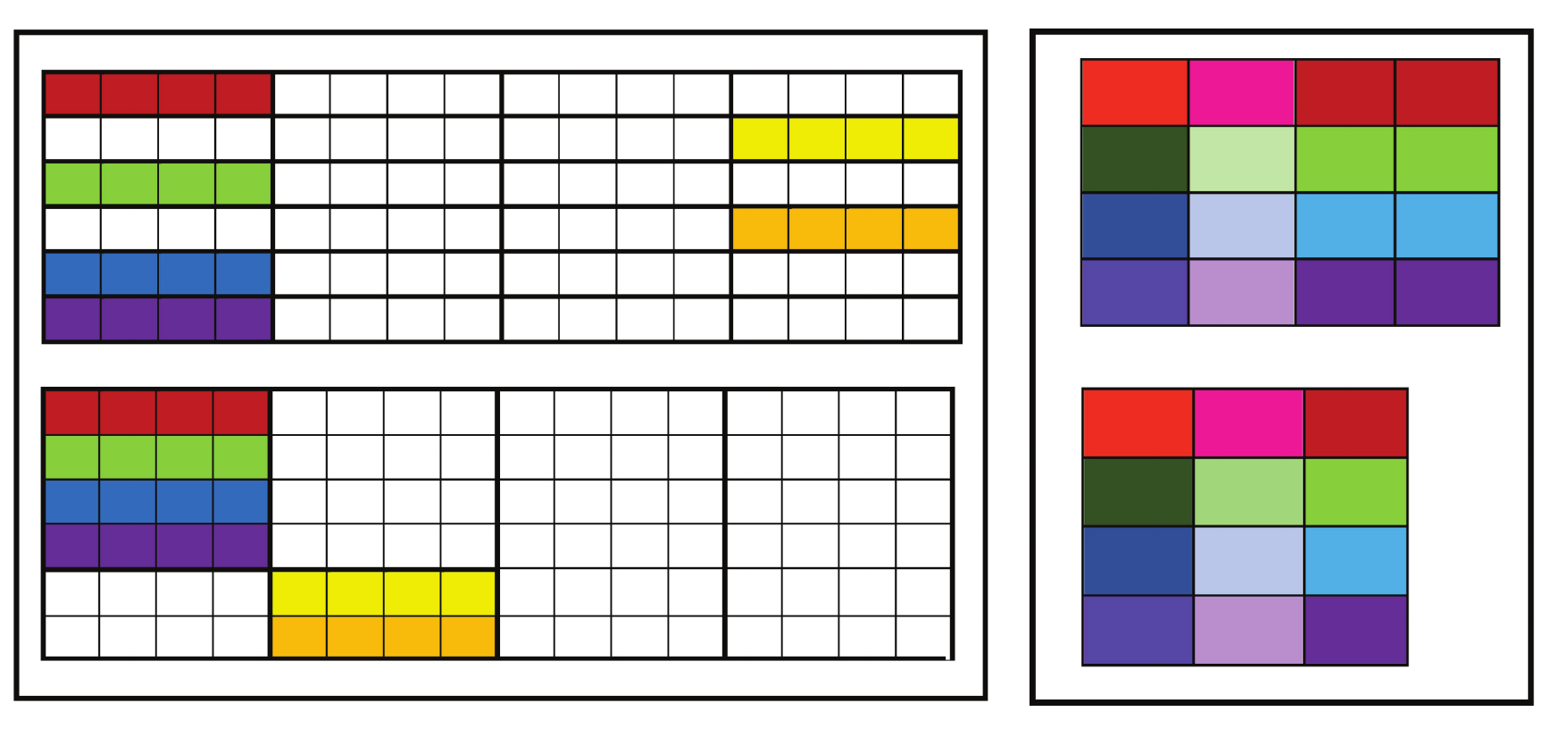}  \]
%
\caption{\small {\bf Tensor structure} with $M=2$, $L=6$, $L_1=4$, $L_2 =2$, $K_1 = K_2 = \barK = 2$, 
$r_1 = 3$, $r_2=2$, $r=5$.  
In the left  panels, white color means zero value but other colors  are used  for object identification only.
In the right panels, different shades of color identify different values of the elements.
 {\bf Top left panel:}  matrix $\calM_3 (\bG)$ (compare with the bottom right panel of Figure~\ref{fig:tensor}).
  {\bf Bottom left panel:}  matrix $\calM_3 (\bG)$  with the rearranged rows and columns. 
{\bf Top right panel:}   matrix $\Phi^{(1)} \in \RR^{L_1 \times K_1^2}$, corresponding to the top left diagonal block of 
the rearranged matrix $\calM_3 (\bG)$ in the bottom left panel of the figure. Due to symmetry of matrices $B\upl$, matrix  
$\Phi^{(1)}$ has two identical columns that correspond to identical non-diagonal entries of matrices $B\upl$. 
{\bf Bottom right panel:}   matrix $\tilPhi^{(1)} \in \RR^{L_1 \times K_1(K_1+1)/2}$  consisting of unique columns of matrix $\Phi^{(1)}$.
}
\label{fig:sig_strength}
\end{figure*}

\noindent
{\bf Proof of Lemma~\ref{lem:matr_W_new}. } 
%
In order to assess properties of matrix $W$, examine  matrix $\calM_3 (\bG)$.
Denote $\bGo = \rhon^{-1}\, \bG$. 
Recall that the matrix $\calM_3 (\bG)$ consists of $L$ rows 
$\calM_3 (\bG)(l,:) = \rhon\, \lkr \vect(G_0\upl)\rkr^T \in \RR^{(M \barK)^2}$.  
Here, due to \fr{eq:Guplmm1}, matrices $G_0\upl$ are block-diagonal with blocks 
\be  \label{eq:G0uplmm}
(G_0\upl)_{m,m} =  \tilB_0\upl \quad \mbox{with} \ 
\tilB_0\upl =  (n-1)   \sqrt{ \hSig\upm} B_0\upl \lkr \sqrt{ \hSig\upm} \rkr^{T}  \in \RR^{K_m^2}, \  s(l)=m.
\ee 
Hence, for every $l$, the only nonzero portion of $G_0\upl$ is  $(G_0\upl)_{m,m}$ 
(see the top left panel of Figure~\ref{fig:sig_strength}).
Since the set of singular values is invariant under permutations of rows and columns,
consider a rearranged  version of of $\calM_3 (\bG)$, where
all rows with $s(l)=m$ are consecutive, and all zero columns 
follow non-zero columns (see the bottom left panel of Figure~\ref{fig:sig_strength}).
Then, re-arranged matrix $\calM_3 (\bG)$ is a concatenation of the block matrices with blocks of rows 
$\vect((G\upl)_{m,m}) = \rhon\,  \vect (\tilB_0\upl)$, where $s(l)=m$, and a zero matrix.
Since  the nonzero part of  matrix $\calM_3 (\bG)$ is block-diagonal, one has 
\be \label{eq:sig_min_bG}
\sig_{\min}(\calM_3 (\bG)) = \sig_{r}(\calM_3 (\bG)) =  \rhon\, \min_{\minM} \sig_{r_m}  \lkr \calM_3 (\bG_0\upm) \rkr,
\ee
where $\calM_3 (\bG_0\upm) \in \RR^{L_m \times K_m^2}$ has rows $[\vect((G_0\upl)_{m,m})]^T$ defined in \fr{eq:G0uplmm}
and $r$ is given by \fr{eq:r_rm}.

Consider matrices $\Phi\upm \in \RR^{L_m \times K_m^2}$ with rows $[\vect(B_0\upl)]^T$, $s(l) =m$ 
(the top right panel of  Figure~\ref{fig:sig_strength} exhibits matrix $\Phi^{(1)}$).
Then, it follows from Theorem 1.2.22 of \cite{GuptaNagar1999} and from \fr{eq:G0uplmm} that 
\be \label{eq:bG0upm}
\calM_3 (\bG_0\upm) = (n-1)\, \Phi\upm \lkr \sqrt{\hSig\upm}  \otimes \sqrt{\hSig\upm} \rkr^T. 
\ee
The above relation implies that the organization of matrices $\bG_0\upm$ and related matrices $W\upm$, $\minM$,
depend on the structure of matrices $\Phi\upm$, which 
have vectorized versions of matrices $B_0\upl$, $s(l)=m$, as their rows. 

Now, consider reductions $\tilW\upm \in \calO_{L_m, r_m}$ of matrices $W\upm \in \calO_{L, r_m}$ 
in \fr{eq:WH_present}  that are obtained by removing zero rows from matrices $W\upm$
(recall Figure~\ref{fig:sig_strength}). 
Let $H_0\upm = \rhon^{-1}\, H\upm$. Then, it follows from  \fr{eq:WH_present}   that
\be \label{eq:calGupm}
\calM_3  (\bG_0\upm) =  \tilW\upm \, H_0\upm,
\ee
 which, together with \fr{eq:bG0upm},  implies that 
$(n-1)\, \Phi\upm \lkr \sqrt{\hSig\upm}  \otimes \sqrt{\hSig\upm} \rkr^T =  \tilW\upm \, H_0\upm$.
Here, due to Assumption~{\bf A3} and Lemma~\ref{lem:tilD_upm}, matrix 
$$
\tilSig\upm = \sqrt{\hSig\upm}  \otimes \sqrt{\hSig\upm} \in R^{K_m^2 \times K_m^2}
$$
has rank $K_m^2$ and eigenvalues bounded above and below by constants. Hence,
\be \label{eq:Phiupm}
\Phi\upm = \tilW\upm\, \calF\upm, \quad \mbox{with} \quad  
\calF\upm = (n-1)^{-1}\, H_o\upm \, (\tilSig\upm)^{-1} \in \RR^{r_m \times K_m^2}.
\ee
%
Before proceeding any further, we note that, due to symmetry, the above diagonal and the below diagonal portions of matrices $B_0\upl$ 
are identical and, therefore, $K_m(K_m-1)/2$  columns of matrices $\Phi\upm$ are repeated twice. 
This repetition does not affect either the rank $r_m$ of $\Phi\upm$ or its smallest nonzero singular value $\sig_{r_m} (\Phi\upm)$. 
Hence, matrices $\Phi\upm$ can be replaced by $\tilPhi\upm \in  \RR^{L_m \times K_m(K_m+1)/2}$, $\minM$   
(see the bottom right panel of Figure~\ref{fig:sig_strength}). Consider sub-matrix $\tilcalF\upm$, obtained from $\calF\upm$ 
by removing the respective columns. Then,  
\be \label{eq:tilPhiupm}
\tilPhi\upm = \tilW\upm\, \tilcalF\upm, \quad \mbox{with} \quad   
\tilW\upm \in \calO_{L_l, r_m}, \ \tilcalF\upm \in \RR^{r_m, K_m(K_m+1)/2}.
\ee


Now, validity of the first relation in \fr{eq:W2inf} follows directly from the block structure and \fr{eq:W_str}.
In order to prove the second part of \fr{eq:W2inf}, consider the SVD 
\be \label{eq:calFupm}
\tilcalF\upm = U_{\calF}\upm \Lam_{\calF}\upm  (V_{\calF}\upm)^T, \quad U_{\calF}\upm \in \calO_{r_m}, \ 
V_{\calF}\upm \in \calO_{K_m(K_m +1)/2, r_m}.
\ee 
Then, \fr{eq:tilPhiupm} and \fr{eq:calFupm} yield 
\be \label{eq:SVD_tilPhi} 
\tilPhi\upm = \tilW\upm U_{\calF}\upm \Lam_{\calF}\upm  (V_{\calF}\upm)^T,
\ee
which, due to $\rank(\tilPhi\upm) = r_m$, provides a version of the SVD for $\tilPhi\upm$. 
Hence,  by Assumption~{\bf A5}, one has
\bes  
\minmM \sig_{r_m} (\Phi\upm) = \minmM \sig_{r_m} (\tilPhi\upm) = \minmM  \Lam_{\calF}\upm (r_m, r_m) 
\geq \lkr \lowC_{\Phi} \, M^{-1}\,L \rkr^{1/2}.
\ees
Consequently, 
\be \label{eq:Lam_calF}
\|(\Lam_{\calF}\upm)^{-1}\|   \leq C (L/M)^{-1/2}.
\ee
On the other hand, by Lemma~\ref{lem:balanced_groups},  one has
\be\label{eq:norm_Lamupm}
\|\Lam_{\calF}\upm \| \leq \|\tilPhi\upm\|_F \leq 
\sqrt{L_m} \|\Phi\upm\|\tinf \leq C\, \sqrt{L}/\sqrt{M}\, \max_{\linL} \|B_0\upl\|_F.
\ee
Now, observe that, by Assumption~{\bf A4}, one has 
\be  \label{eq:Fnorm_Bol}
\lowC \, \sqrt{K} \leq \|B_0\upl\|_F \leq \highC\, \sqrt{K}, \quad \linL,
\ee
so that 
$\maxmM\, \|\tilPhi\upm\|\tinf \leq \maxL\, \|B_0\upl\|_F \leq \highC\, \sqrt{K}$.
Since \fr{eq:SVD_tilPhi}  yields
\be \label{eq:tilW_upm_pr} 
\tilW\upm = \tilPhi\upm\, V_{\calF}\upm \, (\Lam_{\calF}\upm)^{-1}\,   (U_{\calF}\upm)^T,
\ee
derive  
\bes 
\|W\|\tinf = \maxmM\, \|\tilW\upm\|\tinf \leq \maxmM\, \lkv \|\tilPhi\upm\|\tinf\, \|(\Lam_{\calF}\upm)^{-1}\| \rkv,
\ees
which implies  \fr{eq:W2inf}.

Finally, we prove \fr{eq:sc_prod_W_same}. First, consider the case when $r_m = \tilK_m$. In this case, 
remove zero rows from $V_{\calF}\upm$  and obtain $V_{\calF}\upm \in \calO_{r_m}$ in \fr{eq:SVD_tilPhi}.  
Also, for any $i \in [L_m]$, by \fr{eq:Fnorm_Bol}, one has
\be \label{eq:norm_eq}
 \|\tilPhi\upm(i,:)\, V_{\calF}\upm \| = \|\tilPhi\upm(i,:) \|  \geq \lowC \, \sqrt{K}.
\ee 
Consider $l_1, l_2 \in [L]$ with $s(l_1) = s(l_2)=m$, and denote by $i_1, i_2$ the corresponding indices within $\tilW\upm$. 
Then
\begin{align*}
\lan W(l_1,:),  W(l_2,:)\ran & = \lan \tilW\upm(i_1,:),  \tilW\upm(i_2,:)\ran \\
& = 
\tilPhi\upm(i_1,:)  V_{\calF}\upm  (\Lam_{\calF}\upm)^{-2} (V_{\calF}\upm)^T (\tilPhi\upm(i_2,:))^T,
\end{align*} 
so that 
\bes
|\lan W(l_1,:),  W(l_2,:)\ran| \geq \min_{i}  \|\tilPhi\upm(i_1,:)\|^2 \, 
\|\Lam_{\calF}\upm\|^{-2}.
\ees
Therefore, combination of \fr{eq:norm_Lamupm}, \fr{eq:Fnorm_Bol} and \fr{eq:norm_eq} implies that 
\bes
|\lan W(l_1,:),  W(l_2,:)\ran| \geq C \,  M\,L^{-1}  \quad {\rm if} \quad s(l_1) = s(l_2).
\ees

Now, consider the case when $B_0\upl$ with $s(l)=m$  come from a dictionary $\calB_{0,m,t}$  of size $t_m  \leq t_0$. 
Then, matrix $\tilPhi\upm$ contains  only $t_m \leq  t_0$ distinct rows  and $r_m = t_m$, $\minM$. 
Moreover, $\tilW\upm \in \calO_{L_m, t_m}$ are such that
$\tilW\upm(i,t) =0$   if row $i$ of $\tilW\upm$ is not of the type $t$, and 
$W\upm(i,t) = (L_{t,m})^{-1/2}$  if row $i$ is of the type $t$.
Therefore, due to \fr{eq:Ltm}, one has
$\|\tilW\upm(i,:)\| \geq \lowC_t\,  M^{1/2}\,  L^{-1/2}$, which completes the proof. 
\\

\medskip


\noindent
{\bf Proof of Lemma~\ref{lem:matr_Phim}. }    
Consider matrix $\tilPhi\upm \in \RR^{L_m \times K_m(K_m+1)/2}$ in \fr{eq:SVD_tilPhi}. Remove $K_m(K_m+1)/2 - \tilK_m$
zero columns from this matrix and, with some abuse of notations,  denote the new matrix   by $\tilPhi\upm \in \RR^{L_m \times \tilK_m}$.
Construct estimators $\bar{\mu}_0\upm = L_m^{-1} \, 1_{_{L_m}}^T \tilPhi\upm$ and 
\be \label{eq:Psiupm}
\hPsi\upm = (L_m-1)^{-1}\, (\tilPhi\upm -  1_{_{L_m}}\, \bar{\mu}_0\upm)^T \, (\tilPhi\upm -  1_{_{L_m}}\, \bar{\mu}_0\upm)
\ee
of the mean $\mu_0\upm$ and the covariance matrix $\Psi\upm$. Let 
\be \label{eq:brPhi_svd}
(\tilPhi\upm)^T \tilPhi\upm = V_0\upm (\Lam_0\upm)^2 (V_0\upm)^T, \quad V_0\upm \in \calO_{r_m}, \ r_m  = \rank(\tilPhi\upm),
\ee 
 be the SVD of $(\tilPhi\upm)^T \tilPhi\upm$,  where   $r_m \leq \tilK_m$, and 
$\sig_{r_m} (\tilPhi\upm) = \Lam_0\upm(r_m, r_m)$. 
It follows from \fr{eq:Psiupm} that 
\be   \label{eq:sig_rm}
\tilPhi\upm (\tilPhi\upm)^T = (L_m-1)\, \lkv \hPsi\upm + L_m\, (\bar{\mu}_0\upm)^T \bar{\mu}_0\upm \rkv.
\ee
Now consider the two cases:   $r_m=1$,  or  $r_m \geq 2$  and $\lowlam_0 >0$.

If  $r_m=1$, then rows of matrix $\tilPhi\upm$ are equal to each other and equal to
$\mu_0\upm = \bar{\mu}_0\upm$,  therefore, $(\bar{\mu}_0\upm)^T \bar{\mu}_0\upm \geq \lowd_0^2>0$
and, by \fr{eq:sig_rm}, $\sig_{r_m}^2 (\tilPhi\upm) \geq L_m \, \lowd_0^2$.

If  $r_m \geq 2$   and  $\lowlam_0 >0$, then, applying Theorem~6.5 of \cite{wainwright_2019},
obtain that 
\bes \label{eq:sig_min_Wain}
\PP \lfi  \|\hPsi\upm - \Psi\upm\|  \geq   C\, L_m^{-1/2}\, \lkv \sqrt{\tilK_m} + \sqrt{\log L_m}\rkv \rfi
\leq 1 - c\, L_m^{-\tau},
\ees 
 so that $ \|\hPsi\upm - \Psi\upm\| \leq \lowlam_0/2$, if $L_m$ is large enough. By Weyl inequality, one has
$\sig_{r_m} (\hPsi\upm) \geq \sig_{r_m} (\hPsi\upm) - \|\hPsi\upm - \Psi\upm\| \geq \lowlam_0/2$.
Hence, \fr{eq:sig_rm} implies that $\sig_{r_m}^2 (\tilPhi\upm) \geq (L_m-1)\, \lowlam_0^2 /4$, and, thus,  
in both cases,
\be \label{eq:intermed}
\PP \lfi  \sig_{r_m}^2 (\tilPhi\upm) \geq C \, L_m \rfi    \geq 1 - c(L/M)^{-\tau}.
\ee
Then, \fr{eq:sig_hPsiupm} follows from the fact that $\sig_{r_m}^2 (\Phi\upm) \geq \sig_{r_m}^2 (\tilPhi\upm)$ 
and from inequality \fr{eq:hLm}.
\\

\medskip


\noindent
{\bf Proof of Lemma~\ref{lem:sig_strength}.}  
%
Recall that    tensor $\bG_0$  has block-diagonal slices $(G_0\upl)_{m,m}$
given by \fr{eq:G0uplmm}. Hence, Assumption~{\bf A4} and \fr{eq:tilSig_upm}   of Lemma~\ref{lem:tilD_upm} 
yield 
\bes
\sig_{\min} \lkr (G_0\upl)_{m,m} \rkr \geq C\, n\,  \sig_{\min} (B_0\upl) \geq C \, n.
\ees
Consequently, by Courant-Fisher theorem (P 10.3.1 of \cite{Rao_Rao_1998}), one has
\be \label{eq:sigmin1G} 
\sig_{\min}^2(\calM_1 (\bG)) = \rhon^2\   \lam_{\min} \lkr \sumlL  G_0\upl\, (G_0\upl)^T \rkr
\geq \rhon^2\  \sumlL   \sig_{\min}^2 \lkr (G_0\upl)_{m,m} \rkr \geq C \,\rhon^2\, n^2\, L.
\ee  
In order to assess $\sig_{\min}(\calM_3 (\bG))$, recall that \fr{eq:sig_min_bG}, in combination with \fr{eq:bG0upm}
and Assumption~{\bf A5}, imply that 
\be \label{eq:sig_min_Phi}
\sig^2_{\min}(\calM_3 (\bG)) \geq C\, \rhon^2\, n^2 \min_{\minM} \sig_{r_m} (\Phi\upm).
\ee
Therefore, Lemma~\ref{lem:matr_Phim}  yields that, with high probability, 
$\sig^2_{\min}(\calM_3 (\bG)) \geq C\, \rhon^2\, n^2 L\, M^{-1}$. 
The latter results, together with  \fr{eq:sigmin1G}, complete the proof of the lemma.



\subsection{Supplementary Lemmas and Validity of Examples in Section~\ref{sec:assump_A5}}
\label{sec:suppl_lemmas}


\begin{lem} \label{lem:balanced_groups}   
Let Assumption~{\bf A1} hold and $L \geq \di \frac{2\,  M^2}{\lowc_\pi^{2}} \, \log(2 \, M\, n^{\tau})$.  Denote 
$ \di L_m =  \sum_{l=1}^M I(s(l)=m)$. Then, 
\be \label{eq:lem_hLm}
\PP  \lfi \bigcap_{m=1}^M\, \lkr   
\om:\ \lowc_{\pi}\, L/(2\, M)  \leq  L_m \leq  3\, \highc_{\pi}\ L/(2\, M) \rkr
\rfi  \ge 1- n^{-\tau}. 
\ee 
\end{lem} 


\noindent
{\bf Proof of Lemma~\ref{lem:balanced_groups}.\ }  
Note that  $L_m  \sim \text{Binomial}(\pi_m , L)$  for a fixed $m$. By Hoeffding inequality, for any $x > 0$
\bes
\PP \left \{ \left |L_m/L  - \pi_m \right | \ge x \right \} \le 2 \exp\{-2L x^2\}.
\ees
Then, applying Assumption~{\bf A1} and the union bound, obtain
\begin{align*}
& \PP \lfi  \bigcap_{m=1}^M\,  \lkr  \lowc_{\pi}\,L/M - L x \leq L_m  \leq   \highc_{\pi}\,\,L/M + L x \rkr  \rfi \\
& \ge 1- 2 M\, \exp\{-2L x^2\}.
\end{align*}
Set $x = (2M)^{-1} \lowc_\pi$. Then, \fr{eq:lem_hLm} holds provided 
$$
2M \, \exp\lfi - \frac{L (\lowc_\pi)^2}{2 M^2} \rfi \leq n^{-\tau}.
$$
The latter is guaranteed by the condition   $L \geq \di \frac{2\,  M^2}{\lowc_\pi^{2}} \, \log(2 \, M\, n^{\tau})$
of the Lemma.
\\

\medskip


\begin{lem}  \label{lem:tilD_upm}  
Let Assumptions {\bf A1} - {\bf A3} hold  and $n$ be large enough.  
If for $C$  large enough one has
\be \label{eq:n_lowbound_Lemma7}
n \geq C\, (K + \tau \log n),
\ee
then, with probability at least $1 - c\, n^{-\tau}$ derive
\begin{align} \ 
& \min_m \lam_{\min} (\hSig\upm) \propto \max_m \lam_{\max} (\hSig\upm) \propto 1, \label{eq:tilSig_upm}\\
& \min_m \lam_{\min} (\tilD\upm) \propto \max_m \lam_{\max} (\tilD\upm) \propto \sqrt{n}.
\label{eq:tilD_upm}
\end{align}
\end{lem}

\medskip


\noindent
{\bf Proof of Lemma~\ref{lem:tilD_upm}.\ }
Apply Theorem 6.5 of \cite{wainwright_2019} which states that, under conditions of the Lemma, for some $\Ctau$ that depends on $\tau$, 
one has
\be  \label{eq:hsig_oper_er}
\PP \lfi \|\hSig\upm - \Sig\upm\| \leq  \frac{C_\tau\, \highc}{\lowc} \, \sqrt{\frac{K + \tau \log n}{n}} \rfi 
\geq 1 -    n^{-\tau-1}
\ee  
Since 
\bes
\lam_{\min} (\Sig\upm) - \|\hSig\upm - \Sig\upm\| \leq \lam_{\min} (\hSig\upm) \leq \lam_{\max} (\hSig\upm) \leq 
\lam_{\max} (\Sig\upm) + \|\hSig\upm - \Sig\upm\|,
\ees
 by Assumption~{\bf A3},  obtain that for $\om \in \Omtau$, one has
\bes  \label{eq:temp}
\lam_{\min} (\hSig\upm)  \geq \lowc \lkv 1 -   \frac{C_\tau\, \highc}{\lowc^2} \, \sqrt{\frac{K + \tau \log n}{n}}\rkv,
\quad
\lam_{\max} (\hSig\upm)  \leq \highc \lkv 1 +  \frac{C_\tau\}}{\lowc} \, \sqrt{\frac{K + \tau \log n}{n}}\rkv.
\ees 
If $n$ satisfies \fr{eq:n_lowbound_Lemma7} with $C = C_\tau\, \highc\, \lowc^{-2}$, the expressions in the square brackets above
are  bounded below by 1/2.  We complete the proof of \fr{eq:tilSig_upm} by taking the union bound.

Now, validity of  \fr{eq:tilD_upm} follows directly  from \fr{eq:hSig_def}  and \fr{eq:tilSig_upm}.
\\

\medskip

\newpage


\noindent{\bf Justification of Validity of Assumption {\bf A5} in Examples in Section~\ref{sec:assump_A5}}.\\

\noindent
{\it Example~1.\ }
It is easy to see that   condition~\fr{eq:Ltm} in Lemma~\ref{lem:matr_W_new}  holds with high probability due 
to Lemma ~\ref{lem:balanced_groups}.
Also, since $\sig_{r_m}^2 (\Phi\upm) = \|\Phi\upm\|^2_F \geq d_0^2 \, L_m$, Assumption~{\bf A5} is valid by the same lemma.
\\

\medskip

\noindent
{\it Example~2.\ }
Note that $\tilPhi\upm$ has $L_{t,m}$ rows equal to $\tilb_0\upmt$, where, by a version of Lemma~\ref{lem:balanced_groups},
inequality \fr{eq:Ltm} holds with $\lowC_{t,0}$ and $\highC_{t,0}$ that depend on constants $t_0$, $\lowvarpi_0$, $\tilsig_0$
and constants in  Assumptions~{\bf A1}-{\bf A6} provided $L \geq C\, M^3\, \log(M\, n^{\tau})$.
Hence, 
  \bes
  \sig_{t_m}^2 (\Phi\upm) \geq \sig_{t_m}^2 (\calQ\upm)\, \min_{t,m}\, L_{t,m} \geq C\,  \tilsig_0^2 \, M^{-1}\, L,
  \ees
so that Assumption~{\bf A5} is satisfied.  
\\

\medskip

\noindent
{\it Example~3.\ }
Note that, since layers are equipped with the Signed GDPG model, itis possible that $\lowa= -\higha$, 
which leads to $\mu_0 \approx \EE(B_0\upl) = 0$.
Nevertheless, condition    \fr{eq:assump_A5} still holds with high probability since  
eigenvalues of $\Cov(b_0\upl) = \Psi\upm$ in \fr{eq:eig_Psi_upm} are bounded below.
Indeed, consider $\xi \sim \Uniform(\lowa,\higha)$. Then, $\xi = (\lowa+\higha)/2 + (\higha - \lowa)\,\eta$,
where $\eta \sim \Uniform(-0.5, 0.5)$ with $\EE \eta =0$ and $\Var(\eta) = 1/12$. Then, 
$\Psi\upm = (\higha - \lowa)^2/12 * \Psi_0\upm$ where $\Psi_0\upm$ is the $K_m(K_m+1)/2$-diagonal matrix with 
$K_m$ ones followed by  $\om * I_{K_m(K_m-1)/2}$ on the diagonal.
The latter yields that 
$\lam_{\min}(\Psi_0\upm) = \min(\om^2,1)\, (\higha - \lowa)^2/12$.
Of course, if $\om=0$, then $r_m = K_m$ and the lower part of the covariance matrix is omitted,
leading to $\lam_{\min}(\Psi_0\upm) =   (\higha - \lowa)^2/12$.


\newpage




\section{Sensitivity to the misspecification of parameter values}
\label{sec:sensitivity}

Since in reality, the values of $\barK$, $r$ and $M$ are unknown, we analyze numerically how sensitive our procedure is to those parameters.
Our study consists of two parts. In the first part, we examine  the clustering accuracy when $M$ is known but $M\, \bar K$ and $r$ 
are misspecified. In the second part, we assume that parameters  $M\, \bar K$ and $r$  are known 
(since they can be estimated  by  standard methods), but parameter $M$ is unknown and is possibly estimated incorrectly.
\\

\begin{figure*} [t!] 
\centering
\[\includegraphics[width=8.0cm, height=3.8cm]{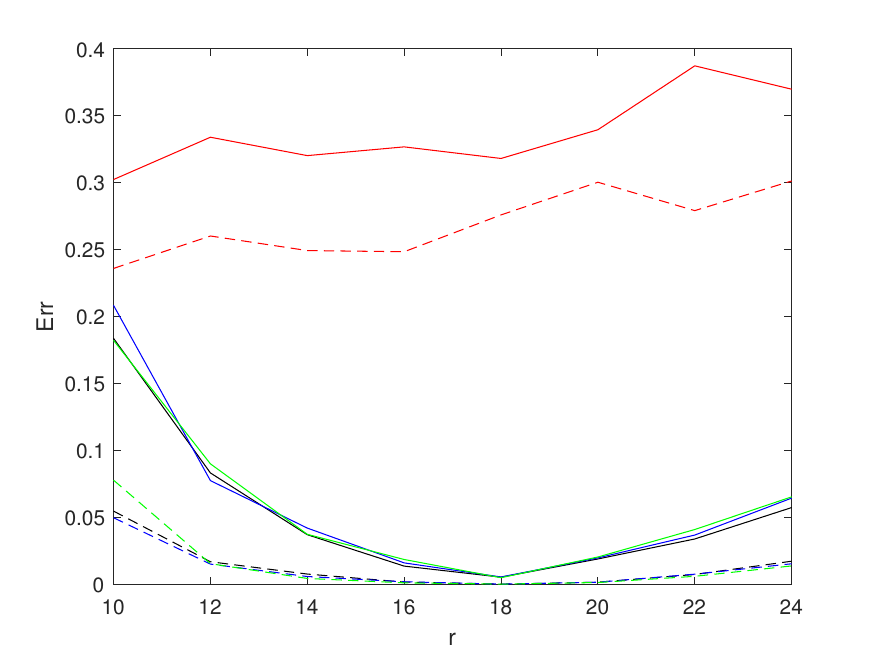}   
\includegraphics[width=8.0cm, height=3.8cm]{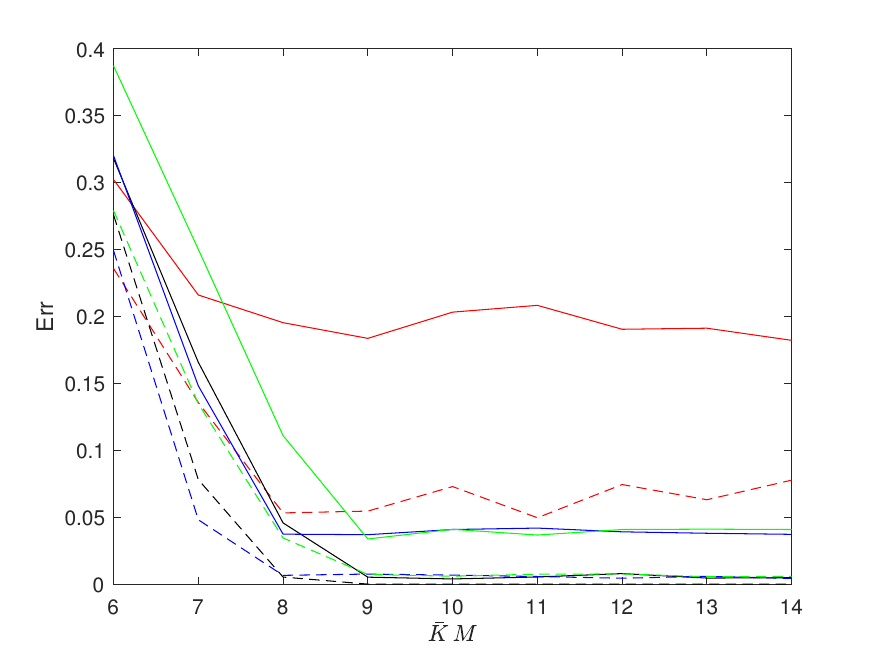}   \]%
\[\includegraphics[width=8.0cm, height=3.8cm]{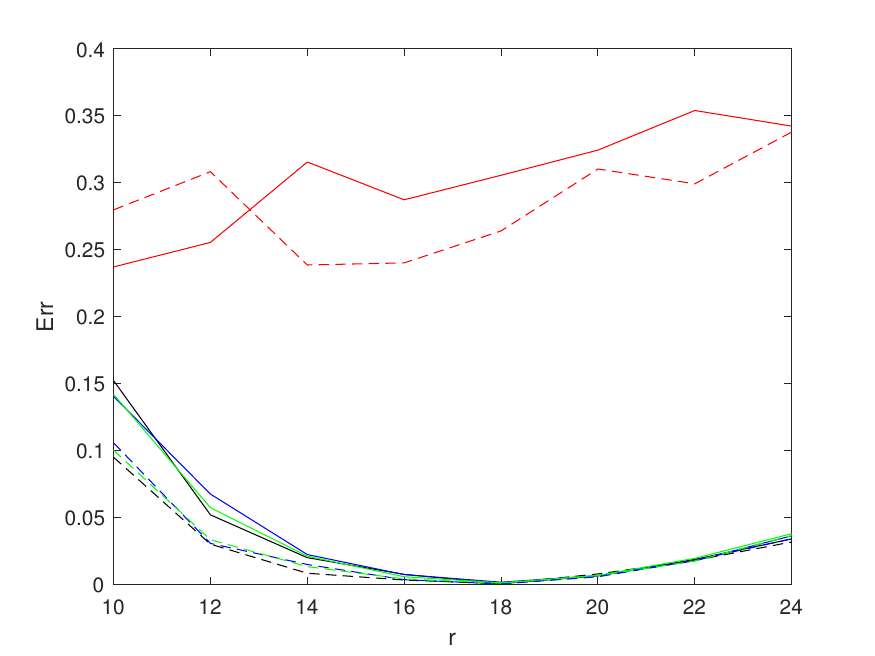}   
\includegraphics[width=8.0cm, height=3.8cm]{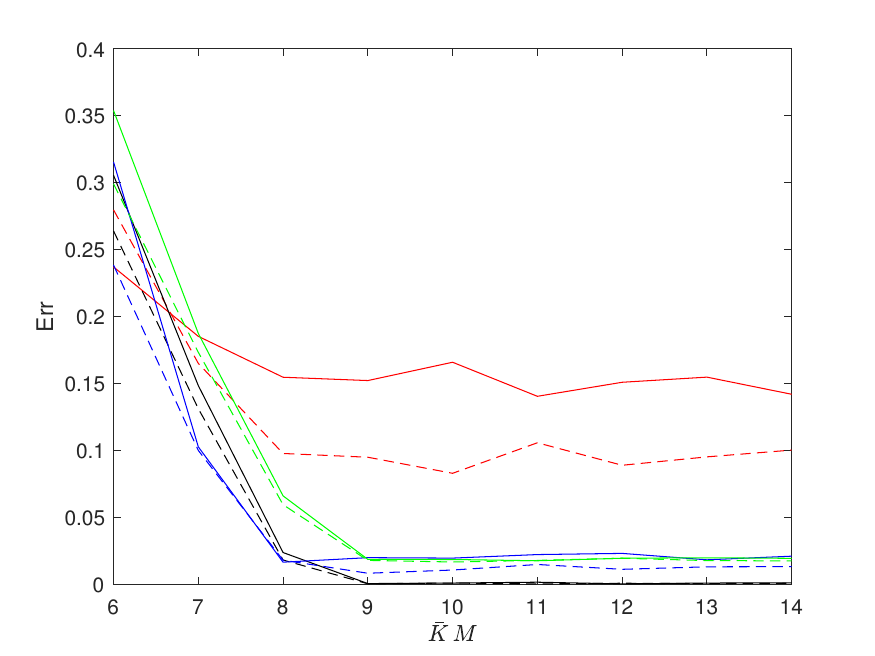}   \]%
\caption{\footnotesize{The between-layer clustering errors of Algorithms~\ref{alg:between} 
for correct $M$ and misspecified values of $M\, \barK$ and $r$.  
Matrices $X\upm$  are generated by the truncated normal distribution  with $\sig =1$, $\mu =0$,   $M=3$ and $K_m=K=3$.
The entries of $B\upl$ are generated  as uniform random numbers  between   $c=-0.05$ and
$d = 0.05$.   \\
Top row: $n = L =150$ (solid lines),  $n = L =250$ (dash lines). 
Bottom row: $n = 150$, $L =250$ (solid lines),  $n = 250$, $L =150$ (dash lines). 
Left panels: varying $\hr$;   $\hKM =6$ (red),  $\hKM=9$ (black),   $\hKM=11$ (blue), $\hKM=14$ (green).
Right panels: varying $\hKM$;  $\hr=10$ (red),  $\hr=18$ (black),   $\hr=14$ (blue), $\hr=22$ (green).
Errors are averaged over 50 simulation runs. In all graphs, correct values are exhibited in black color. 
}}
\label{fig:correctM_wrong_KK_r}
\end{figure*}

\underline{Misspecified $\barK$ and $r$.}\ 
For the first scenario, we fix $M=3$, $\barK = K = 3$, so that $M \, \barK =9$,  and generate elements of matrices $X\upm$  
using \fr{eq:Xm_gen}, where $f_m$ is the truncated normal distribution (Case~1). 
We generate the above diagonal entries of symmetric matrices $B\upl$, $\linL$, in \fr{eq:DIMPnew_GDPG} 
as uniform random numbers  between $c= -0.05$ and $d=0.05$, and form the
 connection probability matrices $P\upl$, $\linL$, using \fr{eq:DIMPnew_GDPG}.  
In this case, according to Example~\ref{example3}, $r =  M*K*(K+1)/2 = 18$.
We examine  this situation for $n$ and $L$ taking values   150   and 250, and estimators $\hKM$ and 
$\hr$  of $\barK M$ and $r$ taking values, respectively, between 6 and 14, and between 10 and 24. 
For each of these scenarios, we evaluate the misclassification   error given by \fr{eq:err_betw_def}.
Results of this investigation are presented in Figure~\ref{fig:correctM_wrong_KK_r}. The black curves correspond to 
the correct parameter values. It is easy to see that for any combination of $n$ and $L$,
the minimum errors are attained for the correct parameter values $\hKM =9$ and $\hr = 18$.

It follows from the left panels of Figure~\ref{fig:correctM_wrong_KK_r} that the clustering accuracy 
is sensitive to underestimating $\barK\, M$. Indeed, the clustering accuracy for $\hKM = 6$ is much worse than for 
$\hKM = 9, 11$ and $14$, where $\hKM = 9$  is the correct value. The minimum error is attained at the correct value of 
$\hr = 18$; however, the clustering errors are not significantly higher when the ``nearby'' values of $\hr$ are used,
especially, if $n$ is large enough ($n=250$ and $L=250$ or $L=150$). Left panels of Figure~\ref{fig:correctM_wrong_KK_r}
also indicate that underestimating $\hr$ seems to lead to somewhat higher errors than overestimating it.

The right panels of Figure~\ref{fig:correctM_wrong_KK_r} appear to agree with this message. Indeed, the red curves 
that correspond  to severely underestimated value $\hr = 10$ are, in general, well above the black (correct $\hr$) 
curves. However, the blue  ($\hr = 14$ instead of 18) and the green curves ($\hr = 22$ instead of 18)
are close to the black curves when $\hKM$ is large enough. 
Nevertheless, when $\barK\, M$ is underestimated ($\hKM = 6$ or 7), overestimating $\hr$ can be very detrimental:
for those values of  $\hKM$, the green curve ($\hr=22$) lies above the red curve ($\hr = 10$).


\begin{figure*} [t!] 
\centering
\[\includegraphics[width=8.0cm, height=3.8cm]{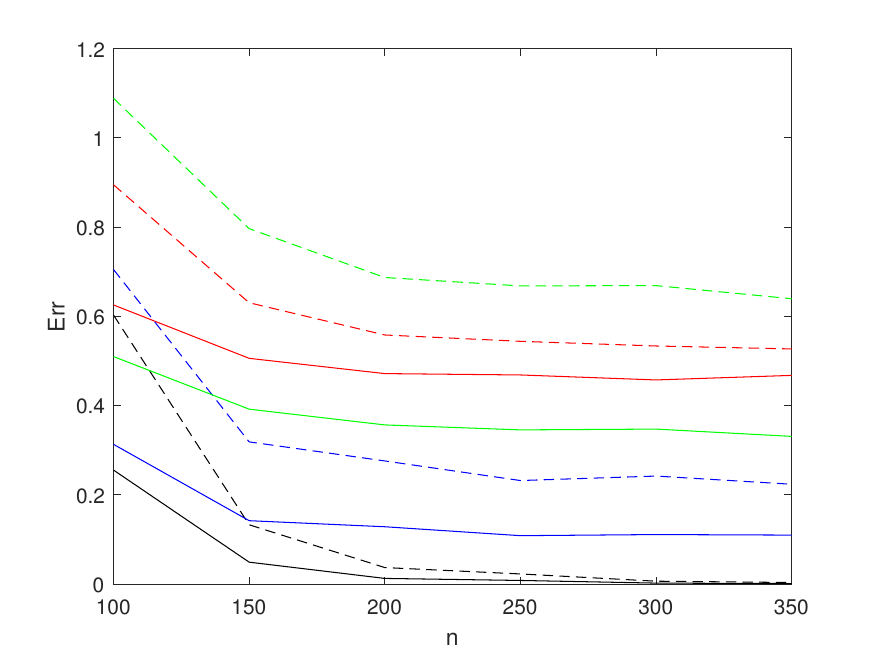}   
\includegraphics[width=8.0cm, height=3.8cm]{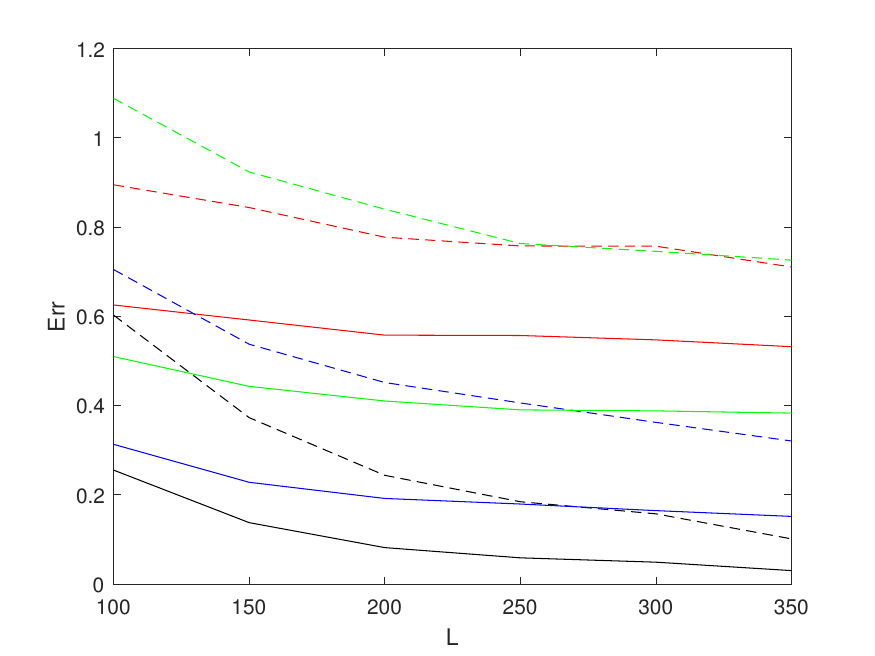}   \]%
\[\includegraphics[width=8.0cm, height=3.8cm]{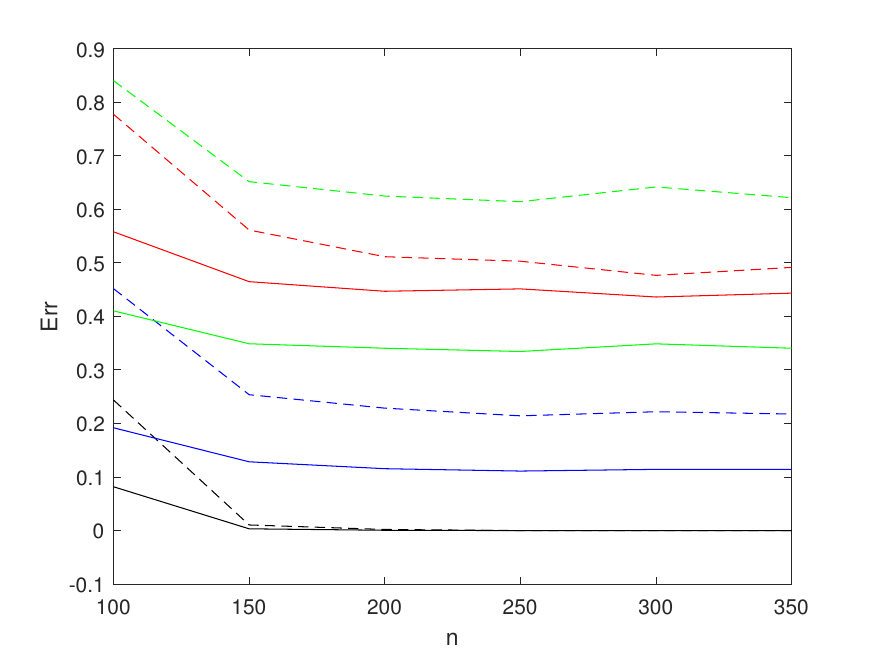}   
\includegraphics[width=8.0cm, height=3.8cm]{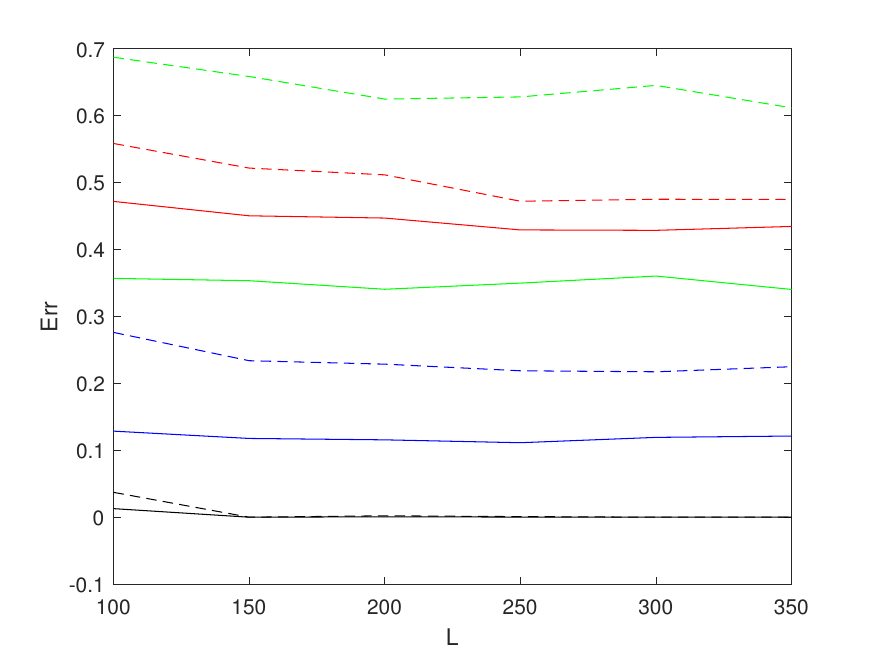}   \]%
\[\includegraphics[width=8.0cm, height=3.8cm]{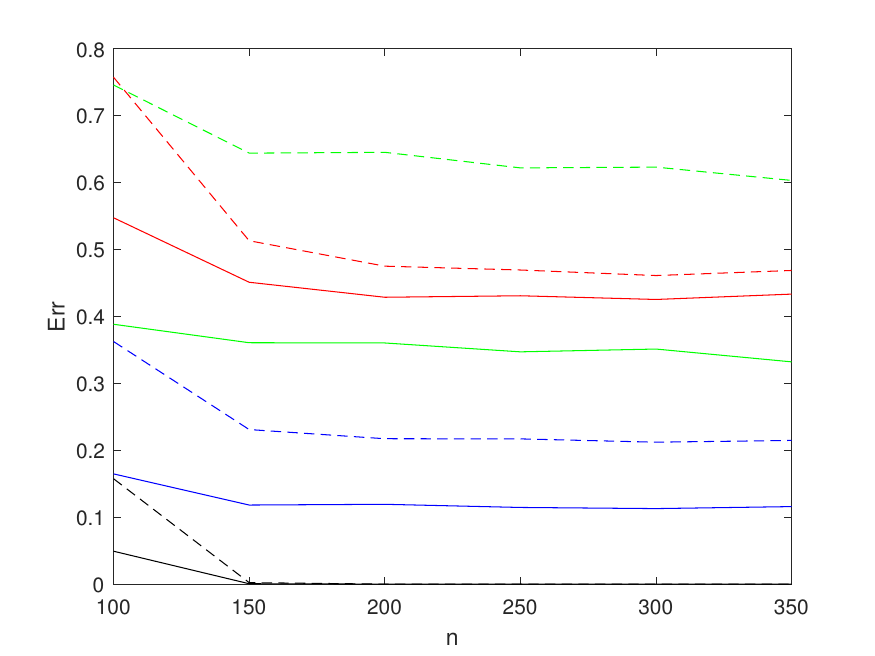}   
\includegraphics[width=8.0cm, height=3.8cm]{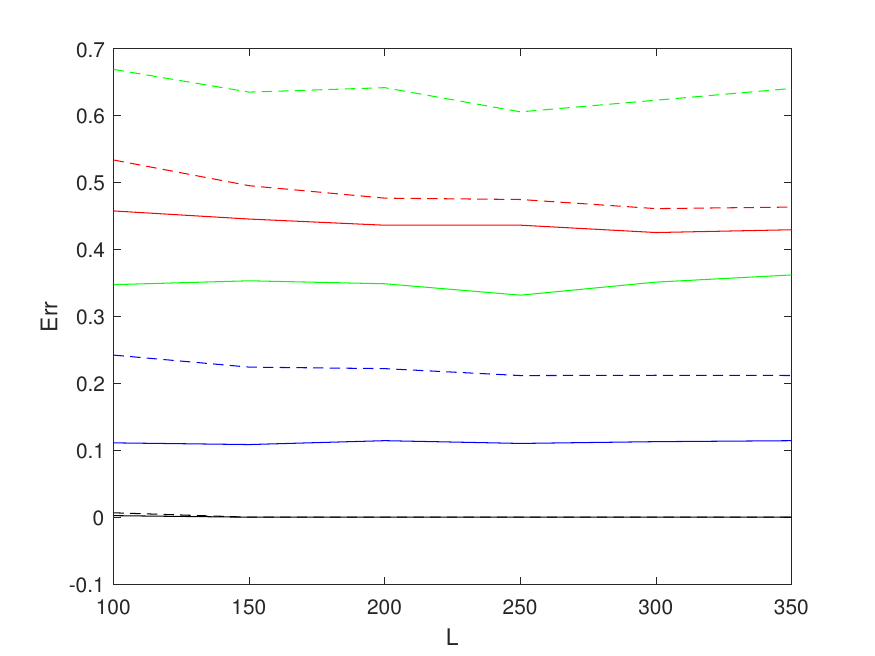}   \]%
\caption{\footnotesize{The $\barARI$ and $\VI$ discrepancies between clustering assignments by
Algorithms~\ref{alg:between} for correct  values of $M\, \barK$ and $r$ and misspecified $M$.
Matrices $X\upm$  are generated by the truncated normal distribution  with $\sig =1$, $\mu =0$,   $M=3$ and $K_m=K=3$.
The entries of $B\upl$ are generated  as uniform random numbers  between   $c=-0.05$ and
$d = 0.05$.  The values of $\barARI$ and $\VI$  are presented by, respectively,  solid and dash lines:
 $\hM =2$ (red),  $\hM =3$ (black),   $\hM =4$ (blue), $\hM =6$ (green).
Left panels: varying $n$;  $L=100$ (top row), $L=200$ (middle row), $L=300$ (bottom row).
Right panels: varying $L$;  $n=100$ (top row), $n=200$ (middle row), $n=300$ (bottom row).
Errors are averaged over 50 simulation runs.  
}}
\label{fig:correct_KK_r_wrong_M}
\end{figure*}


\medskip

\underline{Misspecified $M$.}\ 
Since for a misspecified $M$ one cannot use the misclassification  error as a measure of clustering accuracy,    we use 
the Adjusted Rand Index ({\bf ARI}) and the Variation of Information ({\bf VI}) to evaluate clustering precision. 
If there are two clustering assignments, $A$ and $B$,  of $n$ nodes with respective cluster sizes $K_A$ and $K_B$, 
define $n_{ij}$ to be the number of elements that are simultaneously in cluster $i$ 
according to $A$ and in cluster $j$ according to $B$:
\bes
n_{ij}=|\mathcal A_i\cap\mathcal B_j|, \quad i \in [K_A], \ j \in [K_B].
\ees
Let $\nAi$ and $\nBj$ be the number of elements in clusters $i$ and $j$
according to $A$ and $B$, respectively:
\bes
\nAi=\sum_{j=1}^{K_B}n_{ij},
\qquad
\nBj=\sum_{i=1}^{K_A}n_{ij}.
\ees
Recall that the  total number of pairs of observations is $\binom{n}{2} = n (n-1)/2$.
The Adjusted Rand Index (ARI) is defined as
\be \label{eq:ARI_def}
\ARI(A,B) = \lkr S_{A,B} - S_A\, S_B \rkr \Big/ \lkr \binom{n}{2} -  S_A\, S_B \rkr
\ee
where 
\be \label{eq:SAB}
 S_{A,B} =  \sum_{i=1}^{K_A}\, \sum_{j=1}^{K_B}\binom{n_{ij}}{2}, \quad
S_A = \sum_{i=1}^{K_A}\binom{\nAi}{2}, \quad 
S_B = \sum_{j=1}^{K_B}\binom{\nBj}{2}.
\ee 
Since the  Adjusted Rand Index is a measure of the agreement between two clustering assignments,
the error is measured as 
\be \label{eq:barARI}
\barARI(A,B) = 1 - \ARI(A,B).
\ee
To get some insight for this definition, note that  $S_{A,B}$
 in \fr{eq:ARI_def} is the number of pairs of observations that are assigned to the same cluster in both partitions,
while  $S_A\, S_B$ is the expected value of this quantity under the generalized hypergeometric model 
in which the cluster sizes are fixed. Since $\binom{n}{2}$ is the maximum number of pairs,  $\barARI$ can be interpreted as
the fraction between the difference  of the ''Maximum agreement'' minus the ``Observed agreement''  
over the ''Maximum agreement'' minus the ``Expected agreement''.
Hence, $\barARI =0$ for identical (up to a permutation of the labels) clusterings, 
and  $\barARI \approx 1$ for independent random clusterings. Note 
that $\ARI < 0$ and $\barARI >1$ when the agreement is worse than expected by chance.

Another cluster partition characteristic is an information-theoretic metric,
 Variation of Information (VI), defined as 
\be \label{eq:VI} 
\VI(A,B)= \Ent(A)+\Ent(B)-2\, \Inf(A;B),
\ee
where $\Inf(A;B)$  is the mutual information between $A$ and $B$ and $\Ent(A), \Ent(B)$ are the entropies of 
partitions of $A$ and $B$, given by
\be \label{eq:InfEnt}  
\Inf(A;B)=  - \sum_{i,j} p_{ij}  \log  \lkr \frac{p_{ij}}{p_i\,q_j}\rkr, 
\  
\Ent(A) = \sum_{i} p_i   \log p_i,  
\ 
\Ent(B = \sum_{j} q_j  \log q_j,  
\ee
where
\bes
 p_{ij} =  n_{ij}/n, \quad p_i = \nAi/n, \quad q_j =  \nBj/n.
\ees
Here, $\VI(A,B)=0$  for identical clusterings and $\VI(A,B) = \Ent(A)+\Ent(B)$
for independent clusterings.


Results of our investigation are presented in Figure~\ref{fig:correct_KK_r_wrong_M}
which paints a very different picture from Figure~\ref{fig:correctM_wrong_KK_r}.
Not surprisingly, misspecification of $M$ is much more detrimental to the clustering agreement than 
inaccurate $\hKM$ or $\hr$. Indeed, the black curves that correspond to the correct value $\hM=3$ are 
firmly below all other curves, with the blue curves ($\hM=4$)  having significantly lower agreement 
(higher  $\barARI$ and $\VI$), and the red/green curves ($\hM = 2$ and $\hM = 6$, respectively) also delivering very poor 
agreement with the truth. Unfortunately, the situation does not change substantially  when $n$ and $L$ grow:
even for $n=L=300$ one has $\barARI \approx \VI \approx 0.5$ when $\hM =2$, 
and $\barARI \approx 0.35$. and $\VI \approx 0.7$ when $\hM = 6$.
One can also conclude that underestimating $M$ may be more detrimental than overestimating it since in the latter case
the true clusters may be just partitioned into smaller, relatively pure groups by the clustering algorithm.




\newpage




\section{Additional results for the real data analysis}
\label{sec:addition_realdata}

In this section, we present additional details of the real data analysis.

Tables below reveal   instability levels of Algorithm~~\ref{alg:between}  and~\ref{alg:between_old}
for a variety of parameter values. In particular, Table~\ref{tab:instabH} 
 present instability indices for $M=4$, $M\, \barK$  ranging from 8 to 15 and $r$ ranging from 
$M\, \barK +1$  to $M\, \barK+12$. In comparison, Table~~\ref{tab:instabS} exhibits 
instability indices for Algorithm~\ref{alg:between_old} with $K$ and $M$ ranging  from 2  to 8.
As it is easy to see, Algorithm~~\ref{alg:between}  delivers very stable clustering results with
instability index being close to zero. On the contrary, the instability indices for 
Algorithm~\ref{alg:between_old} are much higher and exceeding 40\% for many pairs of parameters $K$ and $M$.

In addition, Table~\ref{tab:food_items} below presents clustering results obtained using  Algorithm~\ref{alg:between}
with  $M=4$,  $M\, \barK=9$ and $r = 15$. The clusters correspond to commodities broadly labeled as  {\it Fresh Produce} (41 items), 
 {\it Highly processed foods}  (27  items), {\it Animal products and grains} (141 items) and 
{\it Plant derivatives, juices, oils and crop products}  (294 items). Column 2 of   Table~\ref{tab:food_items} 
 contains the cluster labels resulting from application of  Algorithm~~\ref{alg:between}. 
The last column corresponds to the ``true clustering'' manually obtained using the commodity group labels. 
According to Table~\ref{tab:food_items}, 134  out of 503 items, i.e. about 26.6\%, were misclassified.
We did not produce a list of items using Algorithm~\ref{alg:between_old} since it yields very inconsistent clustering results.


\begin{table}[h!]
\centering
\caption{Instability index values  $\Ins$, defined in \fr{eq:instability},  of Algorithm~~\ref{alg:between} 
 as a function of $M\bar K$  and $r$.}
\label{tab:instabH}
\renewcommand{\arraystretch}{1.15}
\begin{tabular}{c|cccccccc}
\hline
$r\ \ \ \ \backslash M\bar K$ & 8 & 9 & 10 & 11 & 12 & 13 & 14 & 15 \\
\hline
 $M\bar K + 1$  & 0.0441 & 0.0293 & 0.0276 & 0.0018 & 0.0060 & 0.0033 & 0.0012 & 0.0556 \\
 $M\bar K + 2$  & 0.0035 & 0.0012 & 0.0014 & 0.0176 & 0.0000 & 0.0073 & 0.0213 & 0.0114 \\
 $M\bar K + 3$  & 0.0000 & 0.0092 & 0.0000 & 0.0000 & 0.0019 & 0.0052 & 0.0021 & 0.0751 \\
 $M\bar K + 4$  & 0.0105 & 0.0003 & 0.0003 & 0.0220 & 0.0010 & 0.0188 & 0.0056 & 0.0152 \\
 $M\bar K + 5$  & 0.0107 & 0.0041 & 0.0046 & 0.0136 & 0.0284 & 0.0162 & 0.0155 & 0.1566 \\
 $M\bar K + 6$  & 0.0007 & 0.0000 & 0.0030 & 0.0377 & 0.0000 & 0.0111 & 0.1006 & 0.0125 \\
 $M\bar K + 7$  & 0.0000 & 0.0062 & 0.0073 & 0.0271 & 0.0115 & 0.0168 & 0.0254 & 0.0877 \\
 $M\bar K + 8$  & 0.0086 & 0.0000 & 0.0000 & 0.0021 & 0.0335 & 0.0018 & 0.0111 & 0.0472 \\
 $M\bar K + 9$  & 0.0440 & 0.0000 & 0.0000 & 0.0531 & 0.0119 & 0.0165 & 0.0038 & 0.0373 \\
 $M\bar K + 10$ & 0.0066 & 0.0025 & 0.0000 & 0.0176 & 0.0055 & 0.0728 & 0.0411 & 0.0159 \\
 $M\bar K + 11$ & 0.0005 & 0.0003 & 0.0111 & 0.0203 & 0.0044 & 0.0802 & 0.0799 & 0.0341 \\
 $M\bar K + 12$ & 0.0863 & 0.0110 & 0.0324 & 0.0008 & 0.0107 & 0.0000 & 0.0009 & 0.0231 \\
\hline
\end{tabular}
\end{table}


\begin{table}[h!]
\centering
\caption{Instability index values  $\Ins$, defined in \fr{eq:instability},   of Algorithm~\ref{alg:between_old}
 as a function of  $M$  and $K$.}
\label{tab:instabS}
\begin{tabular}{c|ccccccc}
\hline
$K \backslash M$ & 2 & 3 & 4 & 5 & 6 & 7 & 8 \\
\hline
2 & 0.4432 & 0.1994 & 0.4895 & 0.5648 & 0.5441 & 0.4536 & 0.4403 \\
3 & 0.6662 & 0.3641 & 0.3047 & 0.3681 & 0.4265 & 0.4344 & 0.4775 \\
4 & 0.5641 & 0.6057 & 0.4626 & 0.3850 & 0.4138 & 0.4442 & 0.4570 \\
5 & 0.4956 & 0.6158 & 0.5366 & 0.4337 & 0.4018 & 0.3999 & 0.4250 \\
6 & 0.2816 & 0.4738 & 0.6470 & 0.5686 & 0.4347 & 0.3986 & 0.3971 \\
7 & 0.2040 & 0.3675 & 0.5737 & 0.6617 & 0.5924 & 0.5249 & 0.4633 \\
8 & 0.1451 & 0.3251 & 0.3679 & 0.6033 & 0.6686 & 0.6277 & 0.5522 \\
\hline
\end{tabular}
\end{table}


\newpage


 
\begin{longtable}{p{0.47\textwidth}cc}
\caption{Clustering of commodities (layers) in the world trade network from 2018
into $M=4$ categories: {\it Fresh Produce} (cluster 1), 
 {\it Highly processed foods}  (cluster  2), {\it Animal products and grains} (cluster 3) and 
{\it Plant derivatives, juices, oils and crop products} (cluster 4).
}
\label{tab:food_items}\\
\hline
\textbf{Item's name} & \textbf{estimated } & \textbf{true} \\
                     & \textbf{cluster } & \textbf{cluster} \\
\hline
\endfirsthead

\hline
\textbf{Item's name} & \textbf{estimated } & \textbf{true} \\
                     & \textbf{cluster } & \textbf{cluster} \\
\hline
\endhead

\hline
\endfoot
Cabbages & 1 & 1 \\
Chillies and peppers, green (Capsicum spp. and Pimenta spp.) & 1 & 1 \\
Cucumbers and gherkins & 1 & 1 \\
Tomatoes & 1 & 1 \\
Watermelons & 1 & 1 \\
Cantaloupes and other melons & 1 & 1 \\
Cauliflowers and broccoli & 1 & 1 \\
Eggplants (aubergines) & 1 & 1 \\
Figs & 4 & 1 \\
Green garlic & 1 & 1 \\
Leeks and other alliaceous vegetables & 1 & 1 \\ 
Lettuce and chicory & 1 & 1 \\
Onions and shallots, dry (excluding dehydrated) & 1 & 1 \\
Peaches and nectarines & 1 & 1 \\
Potatoes & 1 & 1 \\
Pumpkins, squash and gourds & 1 & 1 \\
Strawberries & 1 & 1 \\
Tangerines, mandarins, clementines & 1 & 1 \\
Apples & 1 & 1 \\
Chestnuts, in shell & 1 & 1 \\
Apricots & 1 & 1 \\
Cherries & 1 & 1 \\
Grapes & 1 & 1 \\
Other vegetables, fresh n.e.c. & 1 & 1 \\
Peas, green & 4 & 1 \\
Plums and sloes & 1 & 1 \\
Other beans, green & 1 & 1 \\
Asparagus & 4 & 1 \\
Lemons and limes & 1 & 1 \\
Pears & 1 & 1 \\
Other berries and fruits of the genus vaccinium n.e.c. & 1 & 1 \\
Oranges & 1 & 1 \\
Walnuts, in shell & 4 & 1 \\
Kiwi fruit & 1 & 1 \\
Pomelos and grapefruits & 1 & 1 \\
Bananas & 4 & 1 \\
Mushrooms and truffles & 3 & 1 \\
Other fruits, n.e.c. & 1 & 1 \\
Carrots and turnips & 1 & 1 \\
Hazelnuts, in shell & 4 & 1 \\
Persimmons & 1 & 1 \\
Sour cherries & 3 & 1 \\
Spinach & 1 & 1 \\
Almonds, in shell & 4 & 1 \\
Avocados & 1 & 1 \\
Other tropical fruits, n.e.c. & 4 & 1 \\
Sweet potatoes & 1 & 1 \\
Currants & 4 & 1 \\
Other citrus fruit, n.e.c. & 1 & 1 \\
Raspberries & 1 & 1 \\
Artichokes & 1 & 1 \\
Cashew nuts, in shell & 4 & 1 \\
Pineapples & 4 & 1 \\
Plantains and cooking bananas & 4 & 1 \\
Yams & 4 & 1 \\
Brazil nuts, in shell & 4 & 1 \\
Papayas & 4 & 1 \\
Green corn (maize) & 3 & 1 \\
Quinces & 1 & 1 \\
Blueberries & 4 & 1 \\
Mangoes, guavas and mangosteens & 4 & 1 \\
Onions and shallots, green & 4 & 1 \\
Taro & 4 & 1 \\
Cranberries & 4 & 1 \\
Grapes, marc & 4 & 1 \\
Yautia & 4 & 1 \\
Gooseberries & 4 & 1 \\
Okra & 4 & 1 \\
Cassava, fresh & 4 & 1 \\
Other stone fruits & 4 & 1 \\
Olives preserved & 4 & 2 \\
Vegetables preserved nes (o/t vinegar) & 2 & 2 \\
Chocolate products nes & 2 & 2 \\
Pastry & 2 & 2 \\
Breakfast cereals & 2 & 2 \\
Coffee, decaffeinated or roasted & 2 & 2 \\
Food preparations n.e.c. & 2 & 2 \\
Ice and snow & 2 & 2 \\
Other non-alcoholic caloric beverages & 2 & 2 \\
Sugar confectionery & 2 & 2 \\
Uncooked pasta, not stuffed or otherwise prepared & 4 & 2 \\
Undenatured ethyl alcohol of an alcoholic strength by volume of less than 80\% vol; spirits, liqueurs and other spirituous beverages & 2 & 2 \\
Bread & 2 & 2 \\
Fruit prepared n.e.c. & 2 & 2 \\
Prepared groundnuts & 4 & 2 \\
Refined sugar & 3 & 2 \\
Wine & 2 & 2 \\
Juice of fruits n.e.c. & 2 & 2 \\
Cigarettes & 4 & 2 \\
Cigars and cheroots & 4 & 2 \\
Extracts, essences and concentrates of tea or mate, and preparations with a basis thereof or with a basis of tea or maté & 4 & 2 \\
Cheese from whole cow milk & 2 & 2 \\
Cider and other fermented beverages & 2 & 2 \\
Dog or cat food, put up for retail sale & 2 & 2 \\
Food preparations of flour, meal or malt extract & 2 & 2 \\
Glucose and dextrose & 4 & 2 \\
Infant food & 3 & 2 \\
Mixes and doughs for the preparation of bakers' wares & 2 & 2 \\
Sugar and syrups n.e.c. & 4 & 2 \\
Animal or vegetable fats and oils and their fractions, chemically modified, except those hydrogenated, inter-esterified, re-esterified or elaidinized; inedible mixtures or preparations of animal or vegetable fats or oils & 4 & 2 \\
Malt, whether or not roasted & 3 & 2 \\
Cereal preparations & 3 & 2 \\
Communion wafers, empty cachets of a kind suitable for pharmaceutical use, sealing wafers, rice paper and similar products. & 2 & 2 \\
Other manufactured tobacco and manufactured tobacco substitutes; 'homogenized' or 'reconstituted' tobacco; tobacco extracts and essences & 4 & 2 \\
Prepared nuts & 2 & 2 \\
Vegetables, pulses and potatoes, preserved by vinegar or acetic acid & 2 & 2 \\
Beer of barley, malted & 2 & 2 \\
Coffee extracts & 2 & 2 \\
Vermouth and other wine of fresh grapes flavoured with plats or aromatic substances & 4 & 2 \\
Extracts and juices of meat, fish, crustaceans, molluscs or other aquatic invertebrates & 4 & 2 \\
Homogenized cooked fruit, prepared & 3 & 2 \\
Malt extract & 3 & 2 \\
Coffee substitutes & 4 & 2 \\
Refined cane or beet sugar, in solid form, containing added flavouring or colouring matter; maple sugar and maple syrup & 4 & 2 \\
Pyrethrum extract & 4 & 2 \\
Liver preparations & 3 & 2 \\
Peanut butter & 3 & 2 \\
Fatty liver preparations & 3 & 2 \\
Rice-fermented beverages & 4 & 2 \\
Hides, wet-salted of cattle & 3 & 3 \\
Hen eggs in shell, fresh & 3 & 3 \\
Wheat and meslin flour & 3 & 3 \\
Raw hides and skins of sheep or lambs, with wool & 3 & 3 \\
beef and veal preparations nes & 3 & 3 \\
Ice cream and other edible ice & 3 & 3 \\
Yoghurt, with additives & 3 & 3 \\
Meat of chickens, fresh or chilled & 3 & 3 \\
Meat of cattle boneless, fresh or chilled & 3 & 3 \\
Skim milk and whey powder & 3 & 3 \\
Wheat & 3 & 3 \\
Whole milk powder & 3 & 3 \\
Edible offal of cattle, fresh, chilled or frozen & 3 & 3 \\
Edible offal of pigs, fresh, chilled or frozen & 3 & 3 \\
Meat of cattle with the bone, fresh or chilled & 3 & 3 \\
Edible offals of horses and other equines,  fresh, chilled or frozen & 3 & 3 \\
Horse meat, fresh or chilled & 3 & 3 \\
Horses & 3 & 3 \\
Meat of rabbits and hares, fresh or chilled & 3 & 3 \\
Fat preparations n.e.c. & 4 & 3 \\
Processed cheese & 3 & 3 \\
Barley & 3 & 3 \\
Butter of cow milk & 3 & 3 \\
Ghee from cow milk & 3 & 3 \\
Tallow & 3 & 3 \\
Whey, dry & 3 & 3 \\
Wool grease and lanolin & 4 & 3 \\
Cattle fat, unrendered & 3 & 3 \\
Eggs, dried & 3 & 3 \\
Hides, cattle, dry salted & 3 & 3 \\
Margarine and shortening & 3 & 3 \\
Poultry meat preparations & 3 & 3 \\
Pig meat preparations & 3 & 3 \\
Meat meal & 3 & 3 \\
Shorn wool, greasy, including fleece-washed shorn wool & 4 & 3 \\
Wool, degreased or carbonized, not carded or combed & 4 & 3 \\
Wool, hair waste & 4 & 3 \\
Bees & 4 & 3 \\
Skins, furs & 3 & 3 \\
Fine animal hair, n.e.c. & 4 & 3 \\
Meat of sheep, fresh or chilled & 3 & 3 \\
Cattle & 3 & 3 \\
Raw milk of cattle & 3 & 3 \\
Sausages and similar products of meat, offal or blood of pig & 3 & 3 \\
Fat of pigs & 3 & 3 \\
Meat of pig boneless, fresh or chilled & 3 & 3 \\
Meat of pig with the bone, fresh or chilled & 3 & 3 \\
Spermaceti & 4 & 3 \\
Bran of cereals n.e.c. & 3 & 3 \\
Bran of wheat & 3 & 3 \\
Oats & 3 & 3 \\
Sheep & 3 & 3 \\
Edible offal of sheep, fresh, chilled or frozen & 3 & 3 \\
Buckwheat & 3 & 3 \\
Buttermilk, curdled and acidified milk & 3 & 3 \\
Whole milk, evaporated & 3 & 3 \\
Cream, fresh & 3 & 3 \\
Homogenized meat preparations & 3 & 3 \\
Meat prepared n.e.c. & 3 & 3 \\
Other meat and edible meat offal, salted, in brine, dried or smoked; edible flours and meals of meat or meat offal & 3 & 3 \\
Turkeys & 4 & 3 \\
Bovine meat, salted, dried or smoked & 3 & 3 \\
Cereal straw, husks, unprepared, ground, pressed, or in the form of pellets & 3 & 3 \\
Offals n.e.c. (excluding mammals), fresh, chilled or frozen & 4 & 3 \\
Other meat of mammals, fresh or chilled & 4 & 3 \\
Chickens & 3 & 3 \\
Dairy products n.e.c. & 3 & 3 \\
Goats & 4 & 3 \\
Meat of turkeys, fresh or chilled & 3 & 3 \\
Other raw skins of other animals, preserved & 4 & 3 \\
Whole milk, condensed & 3 & 3 \\
Millet & 3 & 3 \\
Buffalo & 4 & 3 \\
Skim milk of cows & 3 & 3 \\
Pig meat, cuts, salted, dried or smoked (bacon and ham) & 3 & 3 \\
Animal oils and fats n.e.c. & 3 & 3 \\
Meat of camels, fresh or chilled & 4 & 3 \\
Meat of goat, fresh or chilled & 3 & 3 \\
Edible offals and liver of ducks, fresh, chilled or frozen & 3 & 3 \\
Eggs from other birds in shell, fresh, n.e.c. & 3 & 3 \\
Eggs, liquid & 3 & 3 \\
Lactose & 3 & 3 \\
Meat of ducks, fresh or chilled & 3 & 3 \\
Camels & 4 & 3 \\
Hair, carded/combed & 4 & 3 \\
Rye & 3 & 3 \\
Skins, sheep nes & 4 & 3 \\
Skins, karakul & 3 & 3 \\
Poultry fat, rendered & 3 & 3 \\
Lard stearine and lard oil & 3 & 3 \\
Hair, coarse nes & 4 & 3 \\
Fat of poultry & 3 & 3 \\
Other live animals, n.e.c. & 4 & 3 \\
Asses & 4 & 3 \\
Pig fat, rendered & 3 & 3 \\
Wool, shoddy & 4 & 3 \\
Barley, pearled & 3 & 3 \\
Game meat, fresh, chilled or frozen & 3 & 3 \\
Barley flour and grits & 3 & 3 \\
Cheese from milk of sheep, fresh or processed & 3 & 3 \\
Skim milk, condensed & 3 & 3 \\
Butter of goat milk & 3 & 3 \\
Edible offal of buffalo, fresh, chilled or frozen & 4 & 3 \\
Edible offals and liver of chickens and guinea fowl, fresh, chilled or frozen & 3 & 3 \\
Edible offals and liver of geese, fresh, chilled or frozen & 3 & 3 \\
Swine / pigs & 3 & 3 \\
Whey, condensed & 3 & 3 \\
Beeswax & 3 & 3 \\
Edible offals and liver of turkey, fresh, chilled or frozen & 3 & 3 \\
Meat of geese, fresh or chilled & 3 & 3 \\
Skins, dry-salted of sheep & 4 & 3 \\
Meat of pigeons and other birds n.e.c., fresh, chilled or frozen & 4 & 3 \\
Yoghurt & 4 & 3 \\
Flour of mixed grain & 3 & 3 \\
Sausages and similar products of meat, offal or blood of beef and veal & 4 & 3 \\
Hides nes, dry salted & 4 & 3 \\
Ducks & 3 & 3 \\
Rabbits and hares & 3 & 3 \\
Geese & 4 & 3 \\
Skins, wet-salted of calves & 3 & 3 \\
Hair, goat, fine & 4 & 3 \\
Whey, fresh & 3 & 3 \\
Skins, calve, dry salted & 4 & 3 \\
Skins, pig, wet salted & 4 & 3 \\
Hides, wet salted nes & 4 & 3 \\
Degras & 4 & 3 \\
Whey cheese & 4 & 3 \\
Skins, wet-salted of sheep & 4 & 3 \\
Other rodents & 4 & 3 \\
Hides, horse, wet salted & 4 & 3 \\
Hides, cattle nes & 4 & 3 \\
Skins, rabbit & 4 & 3 \\
Cattle, butcher fat & 4 & 3 \\
Bran of mixed grain & 3 & 3 \\
Meat of buffalo, fresh or chilled & 3 & 3 \\
Sheep fat, unrendered & 4 & 3 \\
Crude organic material n.e.c. & 2 & 4 \\
Broad beans and horse beans, green & 4 & 4 \\
Almonds, shelled & 4 & 4 \\
Olive oil & 4 & 4 \\
Compound feed n.e.c. & 4 & 4 \\
Paste of tomatoes & 4 & 4 \\
Sunflower-seed oil, crude & 3 & 4 \\
Essential oils n.e.c. & 2 & 4 \\
Flour of maize & 4 & 4 \\
Maize (corn) & 4 & 4 \\
Beans, dry & 4 & 4 \\
Cake of  soya beans & 4 & 4 \\
Chick peas, dry & 4 & 4 \\
Groundnuts, shelled & 4 & 4 \\
Soya bean oil & 3 & 4 \\
Sunflower seed & 3 & 4 \\
Grape juice & 4 & 4 \\
Lemon juice, concentrated & 4 & 4 \\
Maté leaves & 4 & 4 \\
Natural honey & 4 & 4 \\
Rape or colza seed & 3 & 4 \\
Plums, dried & 3 & 4 \\
Unmanufactured tobacco & 4 & 4 \\
Apple juice, concentrated & 4 & 4 \\
Chillies and peppers, dry (Capsicum spp., Pimenta spp.), raw & 4 & 4 \\
Food wastes & 2 & 4 \\
Linseed & 3 & 4 \\
Other forage products, n.e.c. & 3 & 4 \\
Raisins & 4 & 4 \\
Rice, milled & 4 & 4 \\
Rice, paddy (rice milled equivalent) & 4 & 4 \\
Sorghum & 4 & 4 \\
Soya beans & 4 & 4 \\
Tea leaves & 4 & 4 \\
Anise, badian, coriander, cumin, caraway, fennel and juniper berries, raw & 4 & 4 \\
Apricots, dried & 4 & 4 \\
Canary seed & 4 & 4 \\
Cotton lint, ginned & 4 & 4 \\
Husked rice & 4 & 4 \\
Other fruit n.e.c., dried & 4 & 4 \\
Peas, dry & 3 & 4 \\
Potatoes, frozen & 3 & 4 \\
Vegetable products for feed n.e.c. & 4 & 4 \\
Vegetables frozen & 4 & 4 \\
Vegetables, dehydrated & 4 & 4 \\
Walnuts, shelled & 3 & 4 \\
Other oil seeds, n.e.c. & 4 & 4 \\
Brewing or distilling dregs and waste & 3 & 4 \\
Cake of groundnuts & 4 & 4 \\
Cake of sunflower seed & 3 & 4 \\
Cotton seed & 4 & 4 \\
Flour of pulses & 4 & 4 \\
Gluten feed and meal & 3 & 4 \\
Industrial monocarboxylic fatty acids; acid oils from refining & 4 & 4 \\
Lucerne (alfalfa) meal and pellets & 4 & 4 \\
Oil of maize & 3 & 4 \\
Other oil of vegetable origin, crude n.e.c. & 4 & 4 \\
Pepper (Piper spp.), raw & 4 & 4 \\
Rice, broken & 4 & 4 \\
Soya sauce & 4 & 4 \\
Groundnut oil & 4 & 4 \\
Other live animals non food, n.e.c. & 4 & 4 \\
Dried mushrooms & 4 & 4 \\
Grapefruit juice, concentrated & 4 & 4 \\
Orange juice, concentrated & 4 & 4 \\
Pistachios, in shell & 4 & 4 \\
Sesame seed & 4 & 4 \\
Flour of mustard seed & 3 & 4 \\
Other fructose and syrup & 4 & 4 \\
Tomatoes, peeled (o/t vinegar) & 4 & 4 \\
Lentils, dry & 4 & 4 \\
Raw cane or beet sugar (centrifugal only) & 4 & 4 \\
Cake of maize & 4 & 4 \\
hydrogenated oils and fats & 4 & 4 \\
Oats, rolled & 3 & 4 \\
Sweet corn, prepared or preserved & 4 & 4 \\
Apple juice & 3 & 4 \\
Cashew nuts, shelled & 4 & 4 \\
Dates & 4 & 4 \\
Hazelnuts, shelled & 3 & 4 \\
Orange juice & 4 & 4 \\
Vegetable products, fresh or dry n.e.c. & 4 & 4 \\
Flour of cereals n.e.c. & 4 & 4 \\
Other stimulant, spice and aromatic crops, n.e.c. & 4 & 4 \\
Pineapple juice & 4 & 4 \\
Tomato juice & 4 & 4 \\
Fruit, nuts, peel, sugar preserved & 4 & 4 \\
Figs, dried & 4 & 4 \\
Sugar beet & 4 & 4 \\
Canned mushrooms & 4 & 4 \\
Cinnamon and cinnamon-tree flowers, raw & 4 & 4 \\
Ginger, raw & 4 & 4 \\
Nutmeg, mace, cardamoms, raw & 4 & 4 \\
Grapefruit juice & 4 & 4 \\
Brazil nuts, shelled & 4 & 4 \\
Oil of linseed & 4 & 4 \\
Cocoa butter, fat and oil & 4 & 4 \\
Coffee, green & 4 & 4 \\
Cotton, carded or combed & 4 & 4 \\
Flax, tow and waste & 3 & 4 \\
Homogenized vegetable preparations & 4 & 4 \\
Juice of pineapples, concentrated & 4 & 4 \\
Jute, raw or retted & 4 & 4 \\
Other vegetables provisionally preserved & 4 & 4 \\
Pineapples, otherwise prepared or preserved & 4 & 4 \\
Vegetables preserved (frozen) & 4 & 4 \\
Other nuts (excluding wild edible nuts and groundnuts), in shell, n.e.c. & 4 & 4 \\
Coconut oil & 4 & 4 \\
Hop cones & 4 & 4 \\
Poppy seed & 3 & 4 \\
Vanilla, raw & 4 & 4 \\
Broad beans and horse beans, dry & 4 & 4 \\
Cake of copra & 4 & 4 \\
Coconuts, desiccated & 4 & 4 \\
Coconuts, in shell & 4 & 4 \\
Flour of fruits & 4 & 4 \\
Germ of wheat & 4 & 4 \\
Oil of sesame seed & 4 & 4 \\
Quinoa & 4 & 4 \\
Sweet corn, frozen & 4 & 4 \\
Cake, oilseeds nes & 4 & 4 \\
Cocoa powder and cake & 4 & 4 \\
Mustard seed & 4 & 4 \\
Other pulses n.e.c. & 4 & 4 \\
Rapeseed or canola oil, crude & 3 & 4 \\
Must of grape & 4 & 4 \\
Flours and meals of oil seeds or oleaginous fruits, except those of mustard & 4 & 4 \\
Oil of olive residues & 4 & 4 \\
Cereals n.e.c. & 4 & 4 \\
Cow peas, dry & 4 & 4 \\
Balata, gutta-percha, guayule, chicle and similar natural gums in primary forms or in plates, sheets or strip & 4 & 4 \\
Cake of cottonseed & 4 & 4 \\
Flour, meal, powder, flakes, granules and pellets of potatoes & 3 & 4 \\
Groundnuts, excluding shelled & 4 & 4 \\
Vegetable waxes (other than triglycerides), whether or not refined or coloured & 4 & 4 \\
Juice of lemon & 4 & 4 \\
Molasses & 4 & 4 \\
Natural rubber in other forms & 4 & 4 \\
Natural rubber in primary forms & 4 & 4 \\
Palm kernels & 4 & 4 \\
Palm oil & 4 & 4 \\
Cocoa beans & 4 & 4 \\
Cottonseed oil & 4 & 4 \\
Leather used \& waste & 4 & 4 \\
Oil of castor beans & 4 & 4 \\
Bran of rice & 4 & 4 \\
Cotton waste & 4 & 4 \\
Areca nuts & 4 & 4 \\
Raw silk (not thrown) & 4 & 4 \\
Residues of fatty substances & 4 & 4 \\
Safflower seed & 4 & 4 \\
Sugar cane & 4 & 4 \\
Triticale & 3 & 4 \\
Cloves (whole stems), raw & 4 & 4 \\
Tapioca of potatoes & 4 & 4 \\
True hemp, raw or retted & 3 & 4 \\
Bran of maize & 3 & 4 \\
Cocoa husks and shells & 4 & 4 \\
Cocoa paste not defatted & 4 & 4 \\
Melonseed & 4 & 4 \\
Oil of palm kernel & 4 & 4 \\
Starch of cassava & 4 & 4 \\
Coir, raw & 4 & 4 \\
Rice & 4 & 4 \\
Bran of pulses & 3 & 4 \\
Cassava, dry & 4 & 4 \\
Chicory roots & 4 & 4 \\
Other fibre crops, raw, n.e.c. & 4 & 4 \\
Citrus juice, concentrated n.e.c. & 4 & 4 \\
Flour of rye & 3 & 4 \\
Juice of citrus fruit n.e.c. & 4 & 4 \\
Beet pulp & 3 & 4 \\
Mixed grain & 4 & 4 \\
Other tropical fruit, dried & 4 & 4 \\
Pot barley & 3 & 4 \\
Cake of rapeseed & 3 & 4 \\
Germ of maize & 3 & 4 \\
Coffee husks and skins & 4 & 4 \\
Pulp, waste of fruit for feed & 4 & 4 \\
Flour of roots and tubers n.e.c. & 4 & 4 \\
Edible roots and tubers with high starch or inulin content, n.e.c., fresh & 4 & 4 \\
Flour of rice & 4 & 4 \\
Hay for forage, from other crops n.e.c. & 4 & 4 \\
Safflower-seed oil, crude & 4 & 4 \\
Bagasse & 4 & 4 \\
Locust beans (carobs) & 4 & 4 \\
Silk waste & 4 & 4 \\
Cake of  linseed & 3 & 4 \\
Hempseed & 3 & 4 \\
Bambara beans, dry & 4 & 4 \\
Flax, processed but not spun & 4 & 4 \\
Kola nuts & 4 & 4 \\
Olive residues & 4 & 4 \\
Olives & 4 & 4 \\
Isoglucose & 4 & 4 \\
Other birds & 3 & 4 \\
Beets for fodder & 4 & 4 \\
Flax, raw or retted & 4 & 4 \\
Cotton linters & 4 & 4 \\
Pyrethrum, dried flowers & 4 & 4 \\
Feed, compound, cattle & 4 & 4 \\
Feed, food waste preparations & 4 & 4 \\
Tapioca of cassava & 4 & 4 \\
Feed, compound, poultry & 4 & 4 \\
Mules and hinnies & 3 & 4 \\
Bulgur & 4 & 4 \\
Rice, milled (husked) & 4 & 4 \\
Pigeon peas, dry & 4 & 4 \\
Cake of palm kernel & 4 & 4 \\
Castor oil seeds & 4 & 4 \\
Fonio & 4 & 4 \\
Silk-worm cocoons suitable for reeling & 4 & 4 \\
Karite nuts (sheanuts) & 4 & 4 \\
Flour of millet & 4 & 4 \\
Flour of cassava & 4 & 4 \\
Bran of barley & 4 & 4 \\
Hay, non-leguminous & 4 & 4 \\
Other camelids & 4 & 4 \\
Lupins & 4 & 4 \\
Edible roots and tubers with high starch or inulin content, n.e.c., dry & 4 & 4 \\
Other vegetable juices & 4 & 4 \\
Mustard seed oil, crude & 4 & 4 \\
Abaca, manila hemp, raw & 4 & 4 \\
Oil of citronella & 4 & 4 \\
Copra & 4 & 4 \\
Mango pulp & 4 & 4 \\
Juice of mango & 4 & 4 \\
Liquid margarine & 4 & 4 \\
Buttermilk, dry & 4 & 4 \\
Feed, compound, pigs & 4 & 4 \\
Other sugar crops n.e.c. & 4 & 4 \\
Potato offals & 4 & 4 \\
Sisal, raw & 4 & 4 \\
Juice of tangerine & 4 & 4 \\
Cane sugar, non-centrifugal & 4 & 4 \\
String beans & 4 & 4 \\
Cake of mustard seed & 4 & 4 \\
Cake of sesame seed & 4 & 4 \\
Soya paste & 4 & 4 \\
Swedes for fodder & 4 & 4 \\
Flour of sorghum & 4 & 4 \\
Peppermint, spearmint & 4 & 4 \\
Feed supplements & 4 & 4 \\
Bran of sorghum & 4 & 4 \\
Feed, other concentrates nes & 4 & 4 \\
Cake of rice bran & 4 & 4 \\
Oil of rice bran & 4 & 4 \\
Tung nuts & 4 & 4 \\
\end{longtable}



\end{document}